\documentclass{article}

\PassOptionsToPackage{numbers, compress}{natbib}
\usepackage[final]{neurips_2021}

\usepackage{natbib}
\usepackage[utf8]{inputenc} % allow utf-8 input
\usepackage[T1]{fontenc}    % use 8-bit T1 fonts
\usepackage{hyperref}       % hyperlinks
\usepackage{url}            % simple URL typesetting
\usepackage{booktabs}       % professional-quality tables
\usepackage{amsfonts}       % blackboard math symbols
\usepackage{nicefrac}       % compact symbols for 1/2, etc.
\usepackage{microtype}      % microtypography
\usepackage{xcolor}         % colors
\usepackage{amsmath,amsfonts,bm}

\def\eqref#1{equation~\ref{#1}}
\def\1{\bm{1}}

\DeclareMathAlphabet{\mathsfit}{\encodingdefault}{\sfdefault}{m}{sl}
\SetMathAlphabet{\mathsfit}{bold}{\encodingdefault}{\sfdefault}{bx}{n}

\def\gA{{\mathcal{A}}}

\def\gR{{\mathcal{R}}}
\def\gS{{\mathcal{S}}}

\newcommand{\KL}{D_{\mathrm{KL}}}

\usepackage{subcaption}
\usepackage{graphicx}

\usepackage{algorithm}
\usepackage{algorithmic}
\usepackage{amssymb}
\usepackage{wrapfig}

\newcommand{\diag}{\operatorname{diag}}
\newcommand{\bLozenge}{\mathbin{\blacklozenge}}

\title{Iterative Amortized Policy Optimization}

\author{%
  Joseph Marino\thanks{Now at DeepMind, London, UK. Correspondence to \texttt{josephmarino@deepmind.com}.} \\
  California Institute of Technology \\
   \And
   Alexandre Pich\'e \\
   Mila, Universit\'{e} de Montr\'{e}al
   \AND
   Alessandro Davide Ialongo \\
   University of Cambridge
   \And
   Yisong Yue \\
   California Institute of Technology
}

\begin{document}

\maketitle

\begin{abstract}
  Policy networks are a central feature of deep reinforcement learning (RL) algorithms for continuous control, enabling the estimation and sampling of high-value actions. From the variational inference perspective on RL, policy networks, when used with entropy or KL regularization, are a form of \textit{amortized optimization}, optimizing network parameters rather than the policy distributions directly. However, \textit{direct} amortized mappings can yield suboptimal policy estimates and restricted distributions, limiting performance and exploration. Given this perspective, we consider the more flexible class of \textit{iterative} amortized optimizers. We demonstrate that the resulting technique, iterative amortized policy optimization, yields performance improvements over direct amortization on benchmark continuous control tasks. Accompanying code: { \href{https://github.com/joelouismarino/variational_rl}{\texttt{github.com/joelouismarino/variational\_rl}}}.
\end{abstract}

\section{Introduction}
\label{sec: introduction}

Reinforcement learning (RL) algorithms involve policy evaluation and policy optimization \citep{sutton2018reinforcement}. Given a policy, one can estimate the value for each state or state-action pair following that policy, and given a value estimate, one can improve the policy to maximize the value. This latter procedure, policy optimization, can be challenging in continuous control due to instability and poor asymptotic performance. In deep RL, where policies over continuous actions are often parameterized by deep networks, such issues are typically tackled using regularization from previous policies \citep{schulman2015trust, schulman2017proximal} or by maximizing policy entropy \citep{mnih2016asynchronous, fox2016taming}. These techniques can be interpreted as variational inference \citep{levine2018reinforcement}, using optimization to infer a policy that yields high expected return while satisfying prior policy constraints. This smooths the optimization landscape, improving stability and performance \citep{ahmed2019understanding}.

However, one subtlety arises: when used with entropy or KL regularization, policy networks perform \textit{amortized} optimization \citep{gershman2014amortized}. That is, rather than optimizing the action distribution, e.g.,~mean and variance, many deep RL algorithms, such as soft actor-critic (SAC) \citep{haarnoja2018soft, haarnoja2018soft2}, instead optimize a network to output these parameters, \textit{learning} to optimize the policy. Typically, this is implemented as a direct mapping from states to action distribution parameters. While such \textit{direct} amortization schemes have improved the efficiency of variational inference as ``encoder'' networks \citep{kingma2014stochastic, rezende2014stochastic, mnih2014neural}, they also suffer from several drawbacks: \textit{\textbf{1)}} they tend to provide suboptimal estimates \citep{cremer2018inference, kim2018semi, marino2018iterative}, yielding a so-called ``amortization gap'' in performance \citep{cremer2018inference}, \textit{\textbf{2)}} they are restricted to a single estimate \citep{greff2019multi}, thereby limiting exploration, and \textit{\textbf{3)}} they cannot generalize to new objectives, unlike, e.g., gradient-based \citep{henaff2017model} or gradient-free optimizers \citep{rubinstein2013cross}.

Inspired by techniques and improvements from variational inference, we investigate \textit{iterative} amortized policy optimization. Iterative amortization \citep{marino2018iterative} uses gradients or errors to iteratively update the parameters of a distribution. Unlike direct amortization, which receives gradients only \textit{after} outputting the distribution, iterative amortization uses these gradients \textit{online}, thereby learning to iteratively optimize. In generative modeling settings, iterative amortization empirically outperforms direct amortization \citep{marino2018iterative, marino2018general} and can find multiple modes of the optimization landscape \citep{greff2019multi}.

% Using MuJoCo environments \citep{todorov2012mujoco} from OpenAI gym \citep{brockman2016openai}, we demonstrate performance improvements of iterative amortized policy optimization over direct amortization in model-free and model-based settings. We analyze various aspects of policy optimization, including iterative policy refinement, adaptive computation, and generalization. Identifying policy networks as a form of amortization clarifies suboptimal aspects of direct approaches to policy optimization. Iterative amortization, by harnessing gradient-based feedback during policy optimization, offers an effective and principled improvement.

The contributions of this paper are as follows:
\begin{itemize}
    \item We propose iterative amortized policy optimization, exploiting a new, fruitful connection between amortized variational inference and policy optimization.
    % and we are, to the best of our knowledge, the first to identify entropy and KL-regularized policy networks as a form of amortization, connecting these two areas.
    \item Using the suite of MuJoCo environments \citep{todorov2012mujoco, brockman2016openai}, we demonstrate performance improvements over direct amortized policies, as well as more complex flow-based policies.
    \item We demonstrate novel benefits of this amortization technique: improved accuracy, providing multiple policy estimates, and generalizing to new objectives.
\end{itemize}

% We demonstrate novel benefits of this amortization technique, including iterative policy refinement, adaptive computation, and generalization to new objectives
\section{Background}

\subsection{Preliminaries}

We consider Markov decision processes (MDPs), where $\mathbf{s}_t \in \gS$ and $\mathbf{a}_t \in \gA$ are the state and action at time $t$, resulting in reward $r_t = r (\mathbf{s}_t , \mathbf{a}_t)$. Environment state transitions are given by $\mathbf{s}_{t+1} \sim p_\textrm{env} (\mathbf{s}_{t+1} | \mathbf{s}_t, \mathbf{a}_t)$, and the agent is defined by a parametric distribution, $p_\theta (\mathbf{a}_t | \mathbf{s}_t)$, with parameters $\theta$.\footnote{In this paper, we consider the entropy-regularized case, where $p_\theta (\mathbf{a}_t | \mathbf{s}_t) = \mathcal{U} (-1, 1)$, i.e., uniform. However, we present the derivation for the KL-regularized case for full generality.} The discounted sum of rewards is denoted as $\mathcal{R} (\tau) = \sum_t \gamma^t r_t$, where $\gamma \in ( 0, 1 ]$ is the discount factor, and $\tau = (\mathbf{s}_1, \mathbf{a}_1, \dots)$ is a trajectory. The distribution over trajectories is:
\begin{equation}
    p (\tau) = \rho (\mathbf{s}_1) \prod_{t=1}^{T-1} p_\textrm{env} (\mathbf{s}_{t+1} | \mathbf{s}_t, \mathbf{a}_t) p_\theta (\mathbf{a}_t | \mathbf{s}_t),
\end{equation}
where the initial state is drawn from the distribution $\rho (\mathbf{s}_1)$. The standard RL objective consists of maximizing the expected discounted return, $\mathbb{E}_{p (\tau)} \left[ \gR(\tau) \right]$. For convenience of presentation, we use the undiscounted setting ($\gamma = 1$), though the formulation can be applied with any valid $\gamma$.

\subsection{KL-Regularized Reinforcement Learning}

Various works have formulated RL, planning, and control problems in terms of probabilistic inference \citep{dayan1997using, attias2003planning, toussaint2006probabilistic, todorov2008general, botvinick2012planning, levine2018reinforcement}. These approaches consider the agent-environment interaction as a graphical model, then convert reward maximization into maximum marginal likelihood estimation, learning and inferring a policy that results in maximal reward. This conversion is accomplished by introducing one or more binary observed variables \citep{cooper1988method}, denoted as $\mathcal{O}$ for ``optimality'' \citep{levine2018reinforcement}, with
\begin{equation}
    p (\mathcal{O} = 1 | \tau ) \propto \exp \big( \mathcal{R} (\tau) / \alpha \big), \nonumber
\end{equation}
where $\alpha$ is a temperature hyper-parameter. We would like to infer latent variables, $\tau$, and learn parameters, $\theta$, that yield the maximum log-likelihood of optimality, i.e.,\ $\log p(\mathcal{O} = 1)$. Evaluating this likelihood requires marginalizing the joint distribution, $p(\mathcal{O} = 1) = \int p(\tau, \mathcal{O}=1) d \tau \nonumber$. This involves averaging over all trajectories, which is intractable in high-dimensional spaces. Instead, we can use variational inference to lower bound this objective, introducing a structured approximate posterior distribution:
\begin{equation}
    \pi(\tau | \mathcal{O}) = \rho (\mathbf{s}_1) \prod_{t=1}^{T-1} p_\textrm{env} (\mathbf{s}_{t+1} | \mathbf{s}_t , \mathbf{a}_{t}) \pi (\mathbf{a}_t | \mathbf{s}_t , \mathcal{O}).
\end{equation}
This provides the following lower bound on the objective:
\begin{align}
    \log p(\mathcal{O} = 1) & = \log \int p(\mathcal{O}=1|\tau) p(\tau) d \tau \\
    & \geq \int \pi(\tau | \mathcal{O}) \bigg[ \log \frac{p(\mathcal{O}=1|\tau) p(\tau)}{\pi(\tau | \mathcal{O})} \bigg] d \tau \label{eq: iapo exact integration}\\
    &= \mathbb{E}_\pi [\mathcal{R} (\tau) / \alpha] - \KL(\pi(\tau | \mathcal{O})\|p(\tau)) .
\end{align}
Equivalently, we can multiply by $\alpha$, defining the variational RL objective as:
\begin{equation}
    \mathcal{J} (\pi, \theta) \equiv \mathbb{E}_\pi [\mathcal{R} (\tau)] - \alpha \KL(\pi(\tau | \mathcal{O})\|p(\tau)). \label{eqn:obj}
\end{equation}
This objective consists of the expected return (i.e.,\ the standard RL objective) and a KL divergence between $\pi(\tau | \mathcal{O})$ and $p(\tau)$. In terms of states and actions, this objective is:
\begin{equation}
    \mathcal{J} (\pi, \theta) = \mathbb{E}_{\substack{\mathbf{s}_t, r_t \sim p_\textrm{env} \\ \mathbf{a}_t \sim \pi }} \left[ \sum_{t=1}^T r_t - \alpha\log \frac{\pi ( \mathbf{a}_t | \mathbf{s}_t, \mathcal{O})}{p_\theta (\mathbf{a}_t | \mathbf{s}_t)} \right].
\end{equation}
At a given timestep, $t$, one can optimize this objective by estimating the future terms in the sum using a ``soft'' action-value ($Q_\pi$) network \citep{haarnoja2017reinforcement} or model \citep{piche2019probabilistic}. For instance, sampling $\mathbf{s}_t \sim p_\textrm{env}$, slightly abusing notation, we can write the objective at time $t$ as:
\begin{equation}
    \mathcal{J} (\pi, \theta) = \mathbb{E}_{\pi} \left[ Q_\pi (\mathbf{s}_t , \mathbf{a}_t ) \right] - \alpha \KL (\pi (\mathbf{a}_t | \mathbf{s}_t, \mathcal{O}) || p_\theta (\mathbf{a}_t | \mathbf{s}_t)).
    \label{eq: iapo Q and KL obj}
\end{equation}
Policy optimization in the KL-regularized setting corresponds to maximizing $\mathcal{J}$ w.r.t.~$\pi$. We often consider parametric policies, in which $\pi$ is defined by distribution parameters, $\bm{\lambda}$, e.g., Gaussian mean, $\bm{\mu}$, and variance, $\bm{\sigma}^2$. In this case, policy optimization corresponds to maximizing:
\begin{equation}
    \bm{\lambda} \leftarrow \textrm{arg} \max_{\bm{\lambda}} \mathcal{J} (\pi, \theta).
    \label{eq: generic policy opt}
\end{equation}
Optionally, we can then also learn the policy prior parameters, $\theta$ \citep{abdolmaleki2018maximum}.

% The soft action-value is given as
% \begin{equation}
%     Q_\pi (\mathbf{s}_t , \mathbf{a}_t) = r_t +  \mathbb{E}_{\substack{\mathbf{s}_{t^\prime}, r_{t^\prime} \sim p_\textrm{env} \\ \mathbf{a}_{t^\prime} \sim \pi }} \left[\sum_{t^\prime = t+1}^T r_{t^\prime} - \alpha \log \frac{\pi ( \mathbf{a}_{t^\prime} | \mathbf{s}_{t^\prime}, \mathcal{O})}{p_\theta (\mathbf{a}_{t^\prime} | \mathbf{s}_{t^\prime})} \right],
% \end{equation}
% which can be estimated using temporal difference learning or a model.

% Thus, we see that optimizing the policy, $\pi$, is a structured variational inference problem, requiring stochastic optimization.

\begin{figure*}[t!]
    \centering
    \begin{subfigure}[t]{0.39\textwidth}
        \centering
        \includegraphics[width=0.95\textwidth]{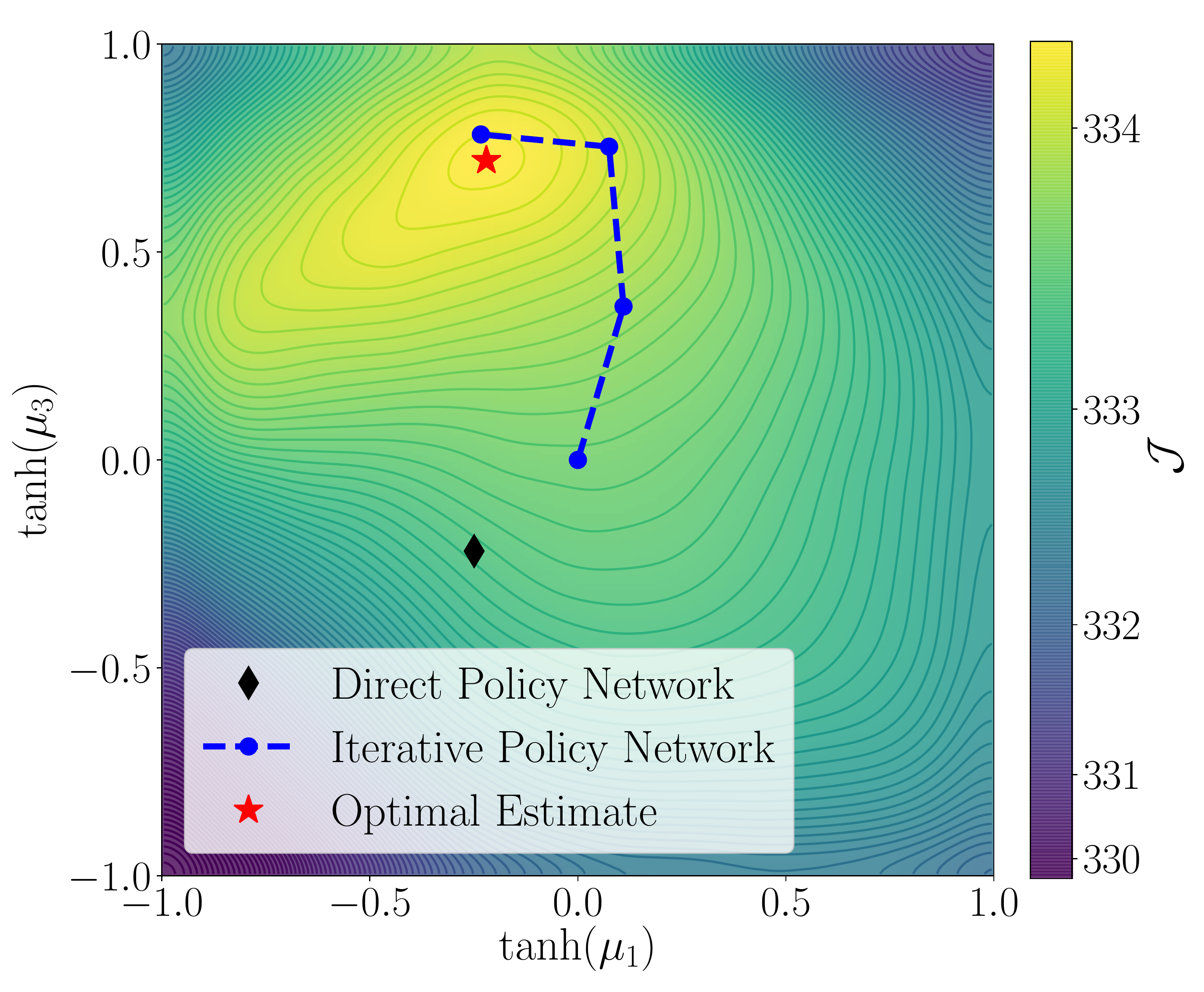}
        % \caption{}
        \label{fig: iapo 2d opt}
    \end{subfigure}%
    ~
    \begin{subfigure}[t]{0.6\textwidth}
        \centering
        \includegraphics[width=\textwidth]{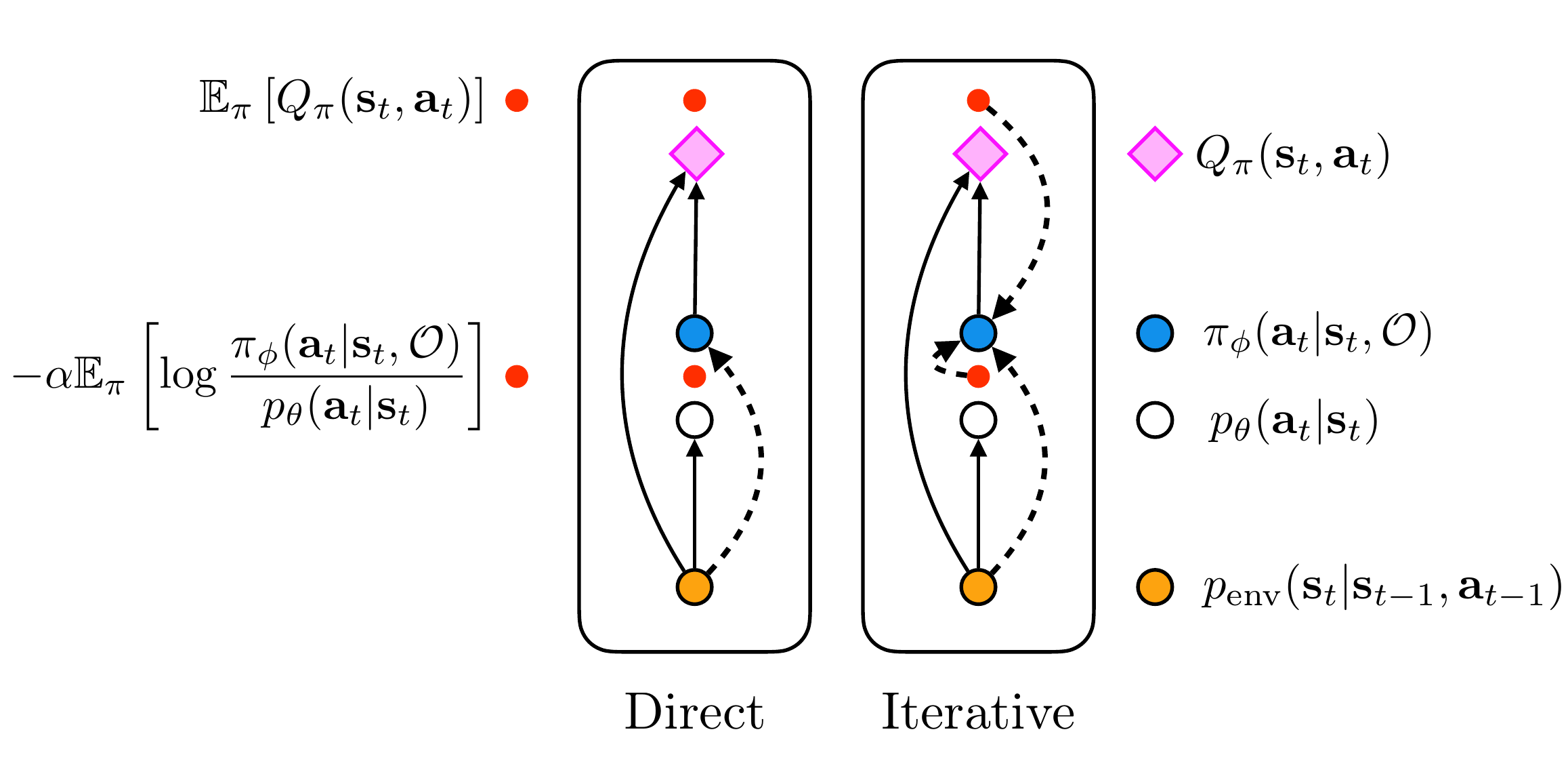}
        % \caption{}
    \end{subfigure}
    \caption{\textbf{Amortization}. \textbf{Left}: Optimization over two dimension of the policy mean, $\mu_1$ and $\mu_3$, for a particular state. A direct amortized policy network outputs a suboptimal estimate, yielding an \textit{amortization gap} in performance. An iterative amortized policy network finds an improved estimate. \textbf{Right}: Diagrams of direct and iterative amortization. Larger circles denote distributions, and smaller red circles denote terms in the objective, $\mathcal{J}$ (Eq.~\ref{eq: iapo Q and KL obj}). Dashed arrows denote amortization. Iterative amortization uses gradient feedback during optimization, while direct amortization does not.}
    \label{fig: iapo amortization}
\end{figure*}

% $\pi_\phi (\mathbf{a} | \mathbf{s})$

\subsection{Entropy \& KL Regularized Policy Networks Perform Direct Amortization}

Policy-based approaches to RL typically do not directly optimize the action distribution parameters, e.g., through gradient-based optimization. Instead, the action distribution parameters are output by a function approximator (deep network), $f_\phi$, which is trained using deterministic \citep{silver2014deterministic, lillicrap2016continuous} or stochastic gradients \citep{williams1992simple, heess2015learning}. When combined with entropy or KL regularization, this policy network is a form of \textit{amortized} optimization \citep{gershman2014amortized}, learning to estimate policies. Again, denoting the action distribution parameters, e.g., mean and variance, as $\bm{\lambda}$, for a given state, $\mathbf{s}$, we can express this direct mapping as
\begin{align}
    \bm{\lambda} \leftarrow f_\phi (\mathbf{s}), && \text{(direct amortization)}
    \label{eq: iapo direct amortization}
\end{align}
denoting the corresponding policy as $\pi_\phi (\mathbf{a} | \mathbf{s}, \mathcal{O}; \bm{\lambda})$. Thus, $f_\phi$ attempts to \textit{learn} to optimize Eq.~\ref{eq: generic policy opt}. This setup is shown in Figure~\ref{fig: iapo amortization} (Right). Without entropy or KL regularization, i.e.~$\pi_\phi (\mathbf{a} | \mathbf{s}) = p_\theta (\mathbf{a} | \mathbf{s})$, we can instead interpret the network as directly integrating the LHS of Eq.~\ref{eq: iapo exact integration}, which is less efficient and more challenging. Regularization smooths the optimization landscape, yielding more stable improvement and higher asymptotic performance \citep{ahmed2019understanding}.

Viewing policy networks as a form of direct amortized variational optimizer (Eq.~\ref{eq: iapo direct amortization}) allows us to see that they are similar to ``encoder'' networks in variational autoencoders (VAEs) \citep{kingma2014stochastic, rezende2014stochastic}. However, there are several drawbacks to direct amortization.

% This raises the following question: \textit{are policy networks providing fully-optimized policy objectives?} In VAEs, it is empirically observed that

\paragraph{Amortization Gap.} Direct amortization results in suboptimal approximate posterior estimates, with the resulting gap in the variational bound referred to as the \textit{amortization gap} \citep{cremer2018inference}. Thus, in the RL setting, an amortized policy, $\pi_\phi$, results in worse performance than the optimal policy within the parametric policy class, denoted as $\widehat{\pi}$. The amortization gap is the gap in following inequality:
\begin{equation}
    \mathcal{J} (\pi_\phi, \theta) \leq \mathcal{J} (\widehat{\pi}, \theta). \nonumber
\end{equation}
Because $\mathcal{J}$ is a variational bound on the RL objective,~i.e., expected return, a looser bound, due to amortization, prevents one from more completely optimizing this objective.

\begin{figure}[!t]
% \begin{wrapfigure}{l}{0.48\textwidth}

\begin{minipage}{0.46\textwidth}
\centering
\begin{algorithm}[H]
\caption{Direct Amortization}
\label{alg: iapo direct amortization}
\begin{algorithmic}
\STATE Initialize $\phi$
\FOR{each environment step}
\STATE $\bm{\lambda} \leftarrow f_\phi (\mathbf{s}_t)$
\STATE $\mathbf{a}_t \sim \pi_\phi (\mathbf{a}_t | \mathbf{s}_t, \mathcal{O}; \bm{\lambda})$
\STATE $\mathbf{s}_{t+1} \sim p_{\textrm{env}} (\mathbf{s}_{t+1} | \mathbf{s}_t, \mathbf{a}_t)$
\ENDFOR
\FOR{each training step}
\STATE $\phi \leftarrow \phi + \eta \nabla_\phi \mathcal{J}$
\ENDFOR
\end{algorithmic}
\end{algorithm}

\end{minipage}
\hfill
\begin{minipage}{0.46\textwidth}
\centering
\begin{algorithm}[H]
\caption{Iterative Amortization}
\label{alg: iapo iterative amortization}
\begin{algorithmic}
\STATE Initialize $\phi$
\FOR{each environment step}
\STATE Initialize $\bm{\lambda}$
\FOR{each policy optimization iteration}
\STATE $\bm{\lambda} \leftarrow f_\phi (\mathbf{s}_t, \bm{\lambda}, \nabla_{\bm{\lambda}} \mathcal{J})$
\ENDFOR
\STATE $\mathbf{a}_t \sim \pi_\phi (\mathbf{a}_t | \mathbf{s}_t, \mathcal{O}; \bm{\lambda})$
\STATE $\mathbf{s}_{t+1} \sim p_{\textrm{env}} (\mathbf{s}_{t+1} | \mathbf{s}_t, \mathbf{a}_t)$
\ENDFOR
\FOR{each training step}
\STATE $\phi \leftarrow \phi + \eta \nabla_\phi \mathcal{J}$
\ENDFOR
\end{algorithmic}
\end{algorithm}

\end{minipage}
% \end{wrapfigure}
\end{figure}

This is shown in Figure~\ref{fig: iapo amortization} (Left),\footnote{Additional 2D plots are shown in Figure~\ref{fig: iapo additional 2d plots} in the Appendix.} where $\mathcal{J}$ is plotted over two dimensions of the policy mean at a particular state in the MuJoCo environment \texttt{Hopper-v2}. The estimate of a direct amortized policy ($\bLozenge$) is suboptimal, far from the optimal estimate (\textcolor{red}{$\bigstar$}). While the relative difference in the objective is relatively small, suboptimal estimates prevent sampling and exploring high-value regions of the action-space. That is, suboptimal estimates have only a \textit{minor} impact on evaluation performance (see Appendix \ref{sec: iapo additional iters}) but hinder effective data collection. 

\paragraph{Single Estimate.} Direct amortization is limited to a single, static estimate. In other words, if there are multiple high-value regions of the action-space, a uni-modal (e.g., Gaussian) direct amortized policy is restricted to only one region, thereby limiting exploration. Note that this is an additional restriction beyond simply considering uni-modal distributions, as a generic optimization procedure may arrive at multiple uni-modal estimates depending on initialization and stochastic sampling (see Section~\ref{sec: iapo benefits}). While multi-modal distributions reduce the severity of this restriction \citep{tang2018boosting, haarnoja2018latent}, the other limitations of direct amortization still persist.

\paragraph{Inability to Generalize Across Objectives.} Direct amortization is a feedforward procedure, receiving gradients from the objective only \textit{after} estimation. This is contrast to other forms of optimization, which receive gradients (feedback) \textit{during} estimation. Thus, unlike other optimizers, direct amortization is incapable of generalizing to new objectives, e.g.,~if $Q_\pi (\mathbf{s}, \mathbf{a})$ or $p_\theta (\mathbf{a} | \mathbf{s})$ change, which is a desirable capability for adapting to new tasks or environments.

To improve upon this scheme and overcome these drawbacks, in Section~\ref{sec: iapo iapo}, we turn to a technique developed in generative modeling, \textit{iterative amortization} \citep{marino2018iterative}, retaining the efficiency of amortization while employing a more flexible iterative estimation procedure.

\subsection{Related Work}

Previous works have investigated methods for improving policy optimization. QT-Opt \citep{kalashnikov2018qt} uses the cross-entropy method (CEM) \citep{rubinstein2013cross}, an iterative derivative-free optimizer, to optimize a $Q$-value estimator for robotic grasping. CEM and related methods are also used in model-based RL for performing model-predictive control \citep{nagabandi2018neural, chua2018deep, piche2019probabilistic, hafner2019learning}. Gradient-based policy optimization \citep{henaff2017model, srinivas2018universal, bharadhwaj2020model}, in contrast, is less common, however, gradient-based optimization can also be combined with CEM \citep{amos2020differentiable}.  Most policy-based methods use direct amortization, either using a feedforward \citep{haarnoja2018soft} or recurrent \citep{guez2019investigation} network. Similar approaches have also been applied to model-based value estimates \citep{byravan2020imagined, clavera2020model, amos2020model}, as well as combining direct amortization with model predictive control \citep{lee2019safe} and planning \citep{riviere2020glas}. A separate line of work has explored improving the policy distribution, using normalizing flows \citep{haarnoja2018latent, tang2018boosting} and latent variables \citep{tirumala2019exploiting}. In principle, iterative amortization can perform policy optimization in each of these settings.

Iterative amortized policy optimization is conceptually similar to negative feedback control \citep{astrom2008feedback}, using errors to update policy estimates. However, while conventional feedback control methods are often restricted in their applicability, e.g., linear systems and quadratic cost, iterative amortization is generally applicable to any differentiable control objective. This is analogous to the generalization of Kalman filtering \citep{kalman1960new} to amortized filtering \citep{marino2018general} for state estimation.

\section{Iterative Amortized Policy Optimization}
\label{sec: iapo iapo}

% We now describe our proposed policy optimization scheme, iterative amortized policy optimization. We first describe the formulation of iterative amortization in Section~\ref{sec: iapo formulation}, then describe considerations that distinguish iterative amortization from direct methods in Section~\ref{sec: iapo considerations}.

\begin{figure*}[t!]
    \centering
    \begin{subfigure}[t]{0.28\textwidth}
        \centering
        \includegraphics[width=\textwidth]{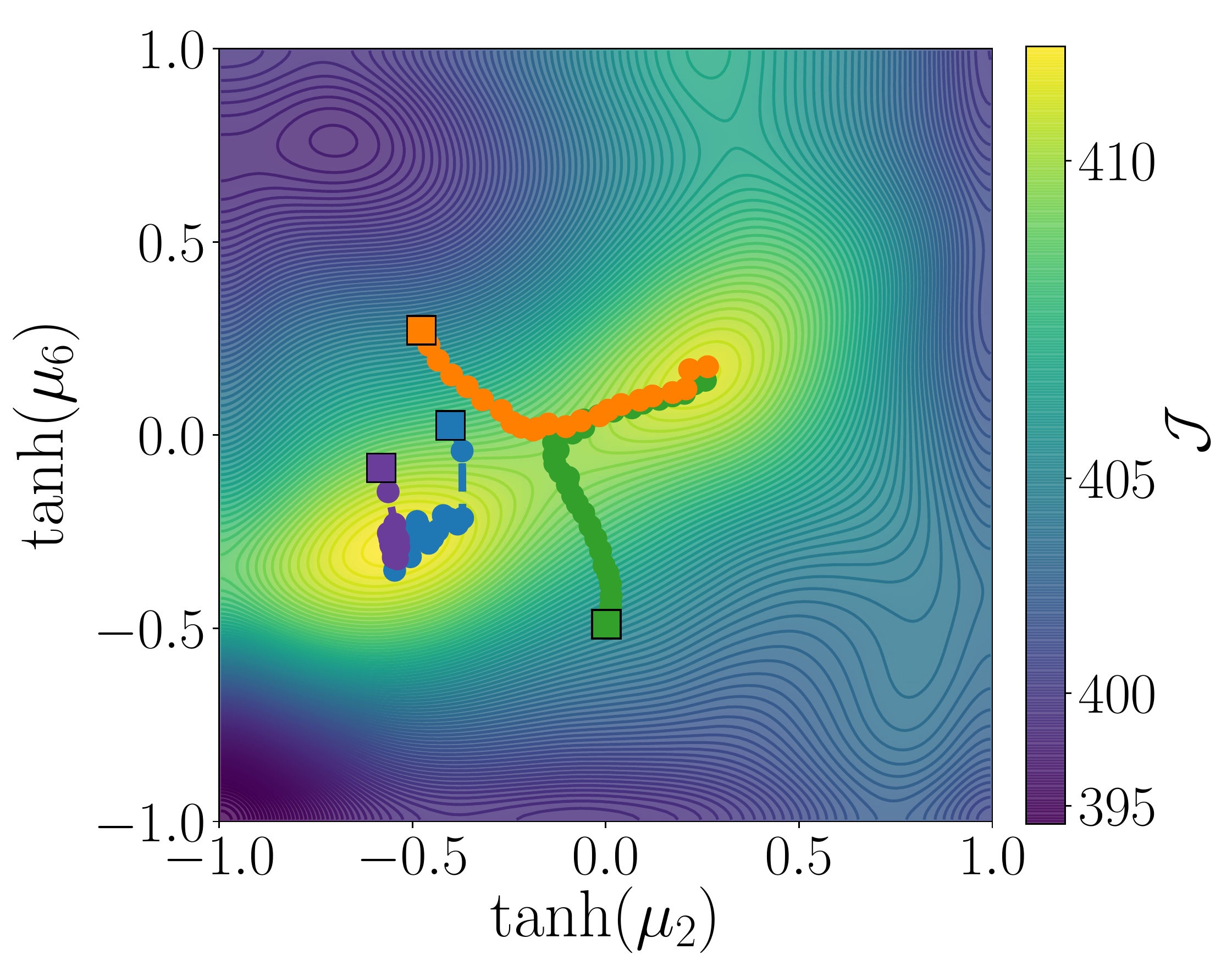}
        % \caption{}
        \label{fig: iapo 2d opt 2}
    \end{subfigure}%
    ~
    \begin{subfigure}[t]{0.48\textwidth}
        \centering
        \includegraphics[width=\textwidth]{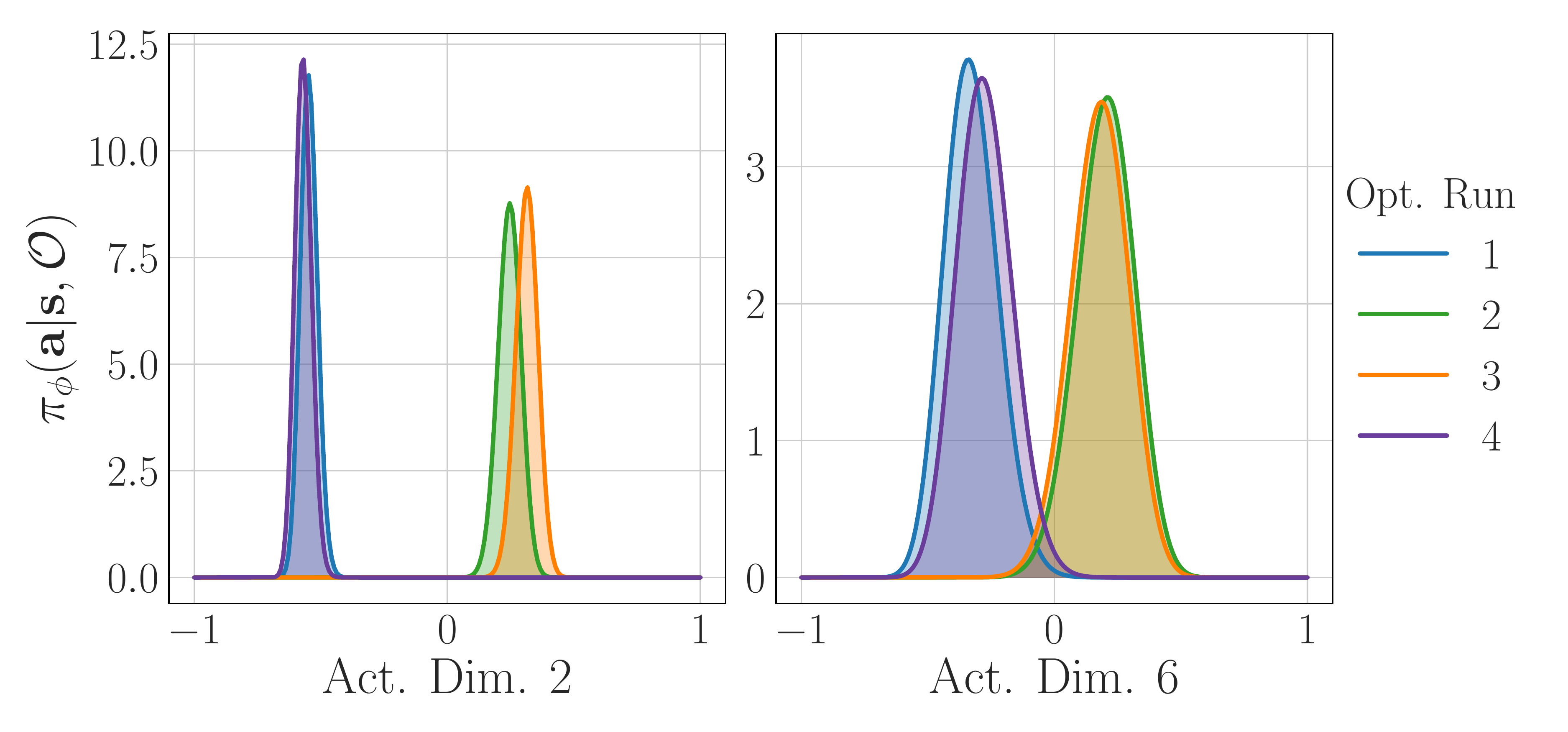}
        % \caption{}
    \end{subfigure}
    \caption{\textbf{Estimating Multiple Policy Modes}. Unlike direct amortization, which is restricted to a single estimate, iterative amortization can effectively sample from multiple high-value action modes. This is shown for a particular state in \texttt{Ant-v2}, showing multiple optimization runs across two action dimensions (\textbf{Left}). Each square denotes an initialization. The optimizer finds both modes, with the densities plotted on the \textbf{Right}. This capability provides increased flexibility in action exploration.}
    \label{fig: iapo multi-modal sampling}
\end{figure*}

\subsection{Formulation}
\label{sec: iapo formulation}

Iterative amortization \citep{marino2018iterative} utilizes errors or gradients to update the approximate posterior distribution parameters. While various forms exist, we consider gradient-encoding models \citep{andrychowicz2016learning} due to their generality. Compared with direct amortization (Eq.~\ref{eq: iapo direct amortization}), we use iterative amortized optimizers of the general form
\begin{align}
    \bm{\lambda} \leftarrow f_\phi (\mathbf{s}, \bm{\lambda}, \nabla_{\bm{\lambda}} \mathcal{J}), && \text{(iterative amortization)}
\end{align}
also shown in Figure~\ref{fig: iapo amortization} (Right), where $f_\phi$ is a deep network and $\bm{\lambda}$ are the action distribution parameters. For example, if $\pi = \mathcal{N} (\mathbf{a}; \bm{\mu}, \diag (\bm{\sigma}^2))$, then $\bm{\lambda} \equiv \left[\bm{\mu}, \bm{\sigma}  \right]$. Technically, $\mathbf{s}$ is redundant, as the state dependence is already captured in $\mathcal{J}$, but this can empirically improve performance \citep{marino2018iterative}. In practice, the update is carried out using a ``highway'' gating operation \citep{hochreiter1997long, srivastava2015training}. Denoting $\bm{\omega}_\phi \in [0, 1]$ as the gate and $\bm{\delta}_\phi$ as the update, both of which are output by $f_\phi$, the gating operation is expressed as
\begin{equation}
    \bm{\lambda} \leftarrow \bm{\omega}_\phi \odot \bm{\lambda} + (\mathbf{1} - \bm{\omega}_\phi) \odot \bm{\delta}_\phi,
    \label{eq: iapo gated update}
\end{equation}
where $\odot$ denotes element-wise multiplication. This update is typically run for a fixed number of steps, and, as with a direct policy, the iterative optimizer is trained using stochastic gradient estimates of $\nabla_\phi \mathcal{J}$, obtained through the path-wise derivative estimator \citep{kingma2014stochastic, rezende2014stochastic, heess2015learning}. Because the gradients $\nabla_{\bm{\lambda}} \mathcal{J}$ must be estimated online, i.e., during policy optimization, this scheme requires some way of estimating $\mathcal{J}$ online through a parameterized $Q$-value network \citep{mnih2013playing} or a differentiable model \citep{heess2015learning}.

% \subsection{Considerations}
% \label{sec: iapo considerations}

\subsection{Benefits of Iterative Amortization}
\label{sec: iapo benefits}

% \subsubsection{Added Flexibility}

\paragraph{Reduced Amortization Gap.}
Iterative amortized optimizers are more flexible than their direct counterparts, incorporating feedback from the objective \textit{during} policy optimization (Algorithm~\ref{alg: iapo iterative amortization}), rather than only \textit{after} optimization (Algorithm~\ref{alg: iapo direct amortization}). Increased flexibility improves the accuracy of optimization, thereby tightening the variational bound \citep{marino2018iterative, marino2018general}. We see this flexibility in Figure~\ref{fig: iapo amortization} (Left), where an iterative amortized policy network iteratively refines the policy estimate (\textcolor{blue}{$\bullet$}), quickly arriving near the optimal estimate.

\paragraph{Multiple Estimates.}
Iterative amortization, by using stochastic gradients and random initialization, can traverse the optimization landscape. As with any iterative optimization scheme, this allows iterative amortization to obtain multiple valid estimates, referred to as ``multi-stability'' in the generative modeling literature \citep{greff2019multi}. We illustrate this capability across two action dimensions in Figure~\ref{fig: iapo multi-modal sampling} for a state in the \texttt{Ant-v2} MuJoCo environment. Over multiple policy optimization runs, iterative amortization finds multiple modes, sampling from two high-value regions of the action space. This provides increased flexibility in action exploration, despite only using a uni-modal policy distribution.

\paragraph{Generalization Across Objectives.}
Iterative amortization uses the gradients of the objective \textit{during} optimization, i.e., feedback, allowing it to potentially generalize to new or updated objectives. We see this in Figure~\ref{fig: iapo amortization} (Left), where iterative amortization, despite being trained with a \textit{different} value estimator, is capable of generalizing to this new objective. We demonstrate this capability further in Section~\ref{sec: iapo experiments}. This opens the possibility of accurately and efficiently performing policy optimization in new settings, e.g., a rapidly changing model or new tasks.

\subsection{Consideration: Mitigating Value Overestimation}
\label{sec: iapo mitigating value overest}

Why are more powerful policy optimizers typically not used in practice? Part of the issue stems from value overestimation. Model-free approaches generally estimate $Q_\pi$ using function approximation and temporal difference learning. However, this has the pitfall of value overestimation, i.e., positive bias in the estimate, $\widehat{Q}_\pi$ \citep{thrun1993issues}. This issue is tied to uncertainty in the value estimate, though it is distinct from optimism under uncertainty. If the policy can exploit regions of high uncertainty, the resulting target values will introduce positive bias into the estimate. More flexible policy optimizers exacerbate the problem, exploiting this uncertainty to a greater degree. Further, a rapidly changing policy increases the difficulty of value estimation \citep{rajeswaran2020game}.

Various techniques have been proposed for mitigating value overestimation in deep RL. The most prominent technique, double deep $Q$-network \citep{van2016deep} maintains two $Q$-value estimates \citep{van2010double}, attempting to decouple policy optimization from value estimation. Fujimoto et al. \cite{fujimoto2018addressing} apply and improve upon this technique for actor-critic settings, estimating the target $Q$-value as the minimum of two $Q$-networks, $Q_{\psi_1}$ and $Q_{\psi_2}$:
\begin{equation}
    \widehat{Q}_\pi (\mathbf{s}, \mathbf{a}) = \min_{i=1,2} Q_{\psi_i^\prime} (\mathbf{s}, \mathbf{a}), \nonumber
\end{equation}
where $\psi_i^\prime$ denotes the target network parameters. As noted by Fujimoto et al. \cite{fujimoto2018addressing}, this not only counteracts value overestimation, but also penalizes high-variance value estimates, because the minimum decreases with the variance of the estimate. Ciosek et al. \cite{ciosek2019better} noted that, for a bootstrapped ensemble of two $Q$-networks, the minimum operation can be interpreted as estimating
\begin{equation}
    \widehat{Q}_\pi (\mathbf{s}, \mathbf{a}) = \mu_Q (\mathbf{s}, \mathbf{a}) - \beta \sigma_Q (\mathbf{s}, \mathbf{a}),
    \label{eq: iapo pessimistic value est}
\end{equation}
with mean $\mu_Q (\mathbf{s}, \mathbf{a}) \equiv \frac{1}{2} \sum_{i=1,2} Q_{\psi_i^\prime} (\mathbf{s}, \mathbf{a})$, standard deviation $\sigma_Q (\mathbf{s}, \mathbf{a}) \equiv (\frac{1}{2} \sum_{i=1,2} (Q_{\psi_i^\prime} (\mathbf{s}, \mathbf{a}) -  \mu_Q (\mathbf{s}, \mathbf{a}))^2 )^{1/2}$, and $\beta = 1$. Thus, to further penalize high-variance value estimates, preventing value overestimation, we can increase $\beta$. For large $\beta$, however, value estimates become overly pessimistic, negatively impacting training. Thus, $\beta$ reduces target value variance at the cost of increased bias.

\begin{wrapfigure}{r}{0.5\textwidth}
  \begin{center}
    \begin{subfigure}[t]{0.23\textwidth}
        \centering
        \includegraphics[width=0.95\textwidth]{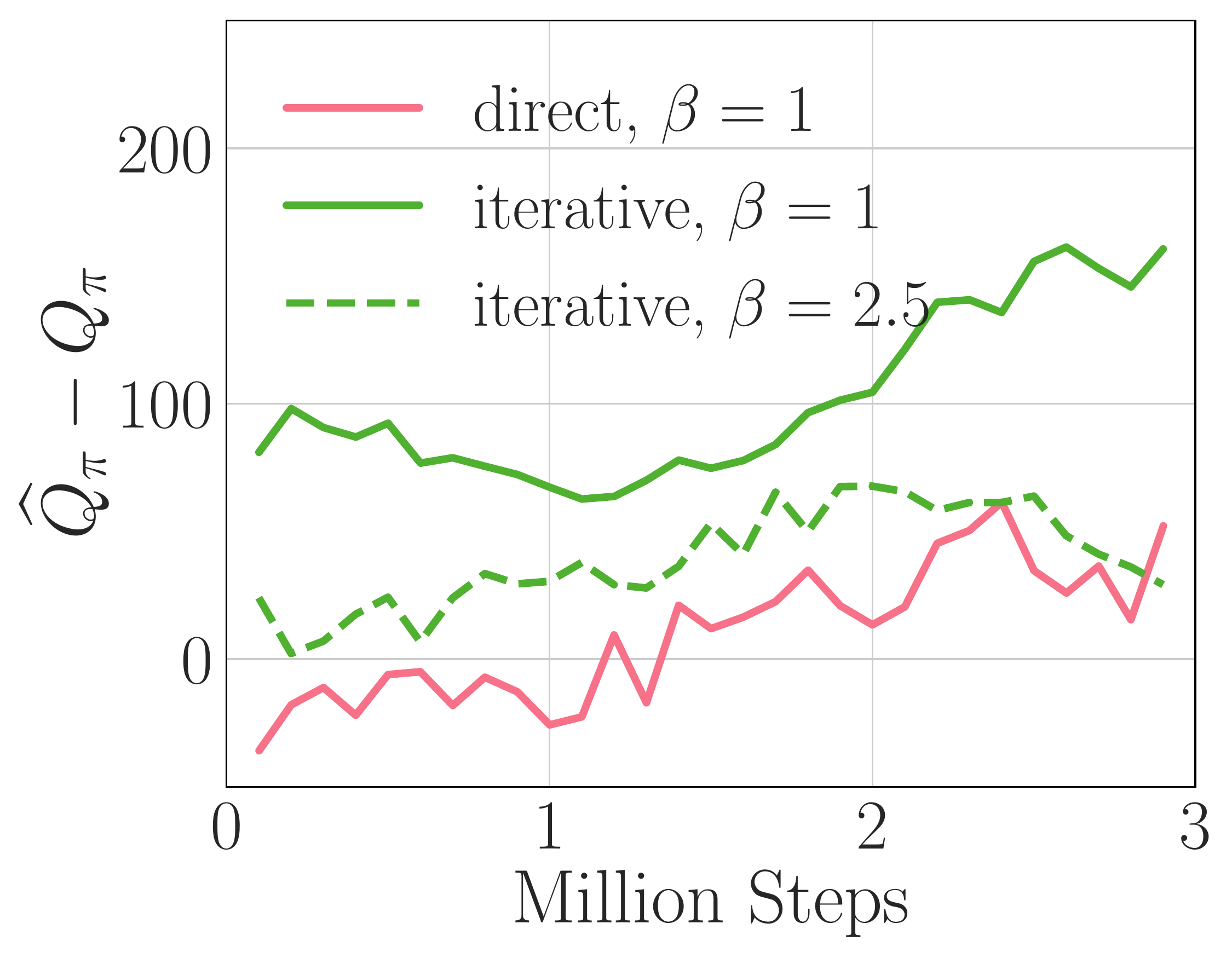}
        \caption{}
        \label{fig: value bias}
    \end{subfigure}% 
    ~
    \begin{subfigure}[t]{0.23\textwidth}
        \centering
        \includegraphics[width=0.94\textwidth]{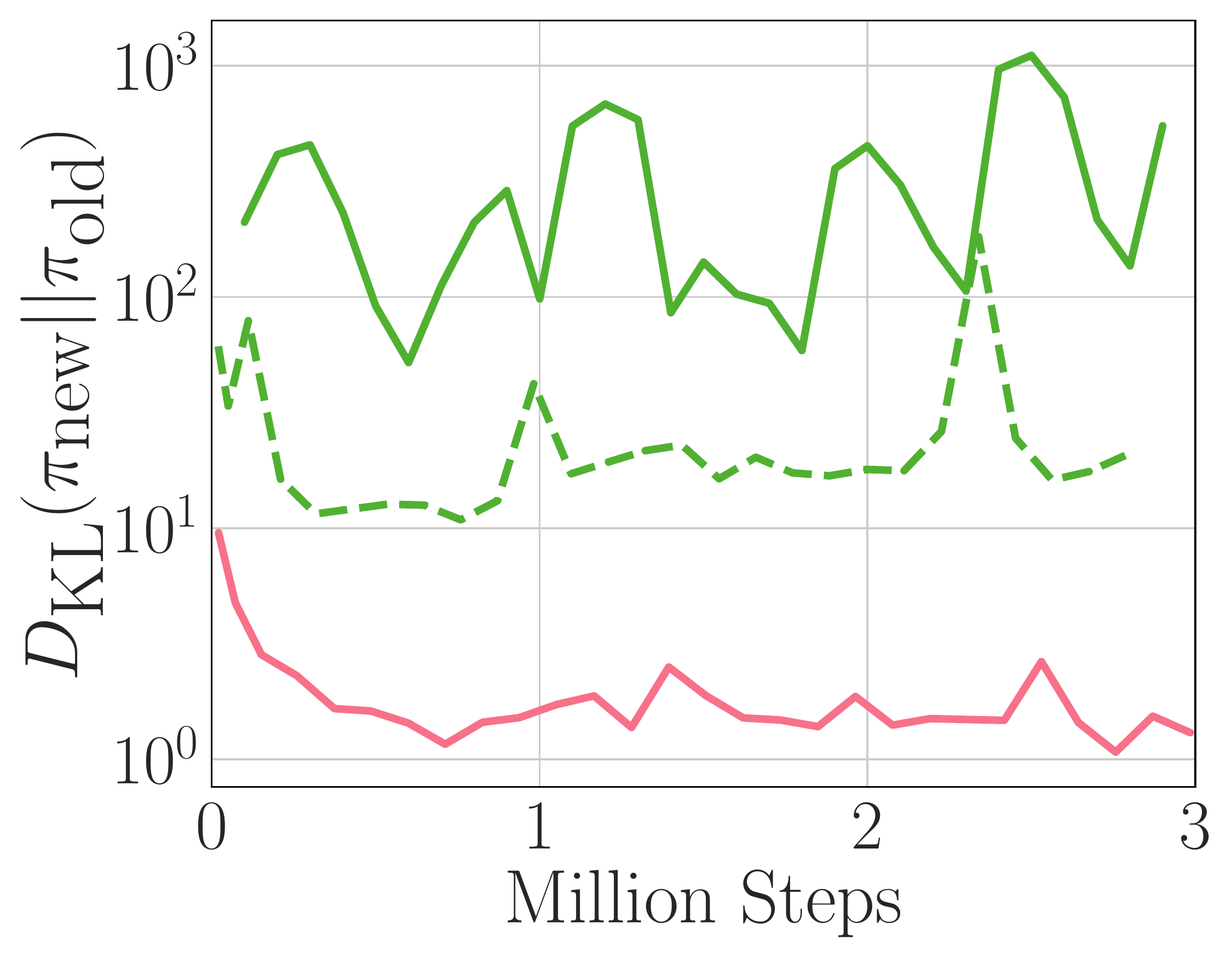}
        \caption{}
        \label{fig: agent kl}
    \end{subfigure}
  \end{center}
  \caption{\textbf{Mitigating Value Overestimation}. With $\beta=1$, iterative amortization results in (\textbf{a}) higher value overestimation and (\textbf{b}) a more rapidly changing policy as compared with direct amortization. Increasing $\beta$ helps to mitigate these issues.}
    \label{fig: value overestimation}
\end{wrapfigure}

Due to the flexibility of iterative amortization, the default $\beta = 1$ results in increased value bias and a more rapidly changing policy as compared with direct amortization (Figure~\ref{fig: value overestimation}). Further penalizing high-variance target values ($\beta = 2.5$) reduces value overestimation and improves stability. For details, see Appendix \ref{appendix: value bias est}. Recent techniques for mitigating overestimation have been proposed, such as adjusting $\alpha$ \citep{fox2019toward}. In offline RL, this issue has been tackled through the action prior \citep{fujimoto2019off, kumar2019stabilizing, wu2019behavior} or by altering $Q$-network training \citep{agarwal2019optimistic, kumar2020conservative}. While such techniques could be used here, increasing $\beta$ provides a simple solution with no additional computational overhead. This is a meaningful insight toward applying more powerful policy optimizers.

\begin{figure*}[t!]
    \centering
    \begin{subfigure}[t]{0.37\textwidth}
        \centering
        \includegraphics[width=\textwidth]{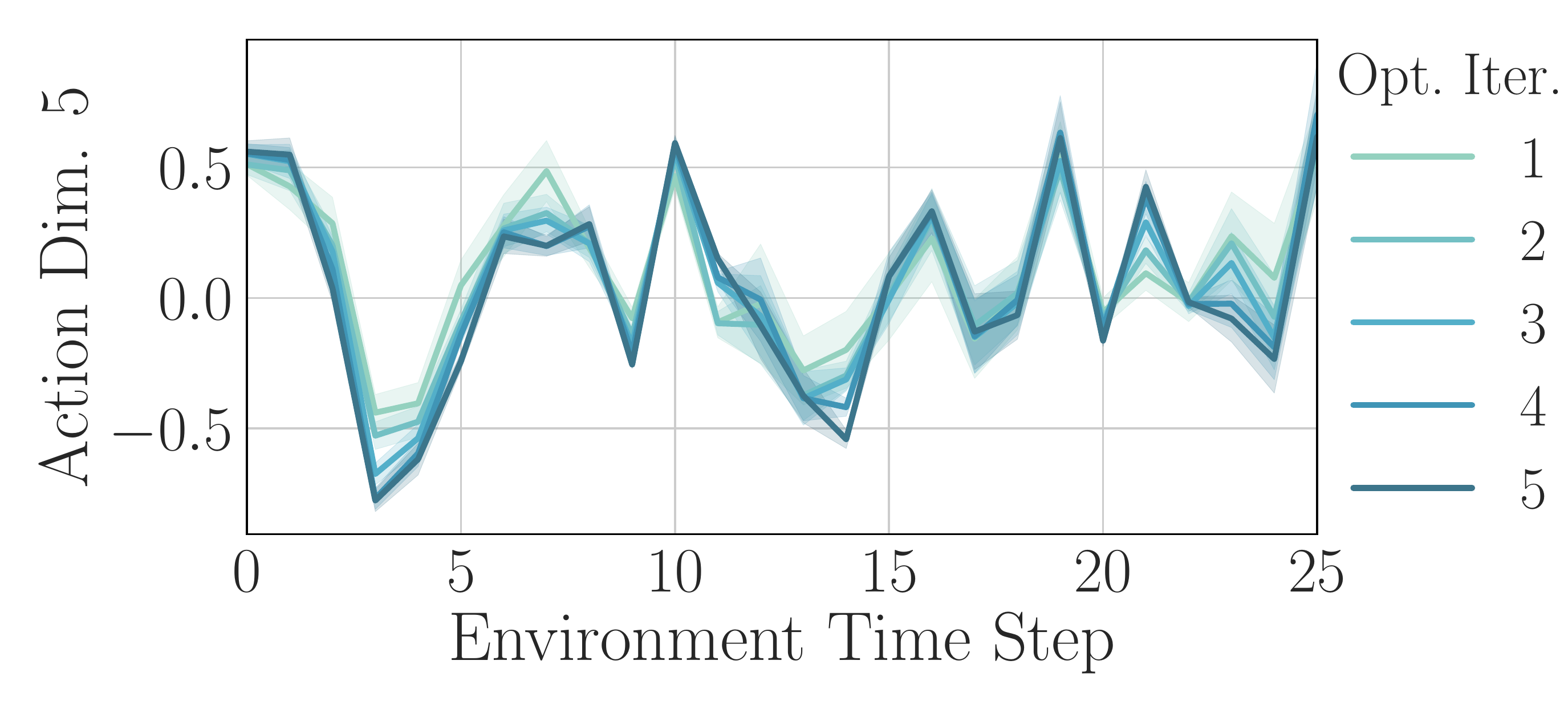}
        \caption{Policy}
        \label{fig: iapo policy opt single dim ant}
    \end{subfigure}% 
    ~
    \begin{subfigure}[t]{0.37\textwidth}
        \centering
        \includegraphics[width=\textwidth]{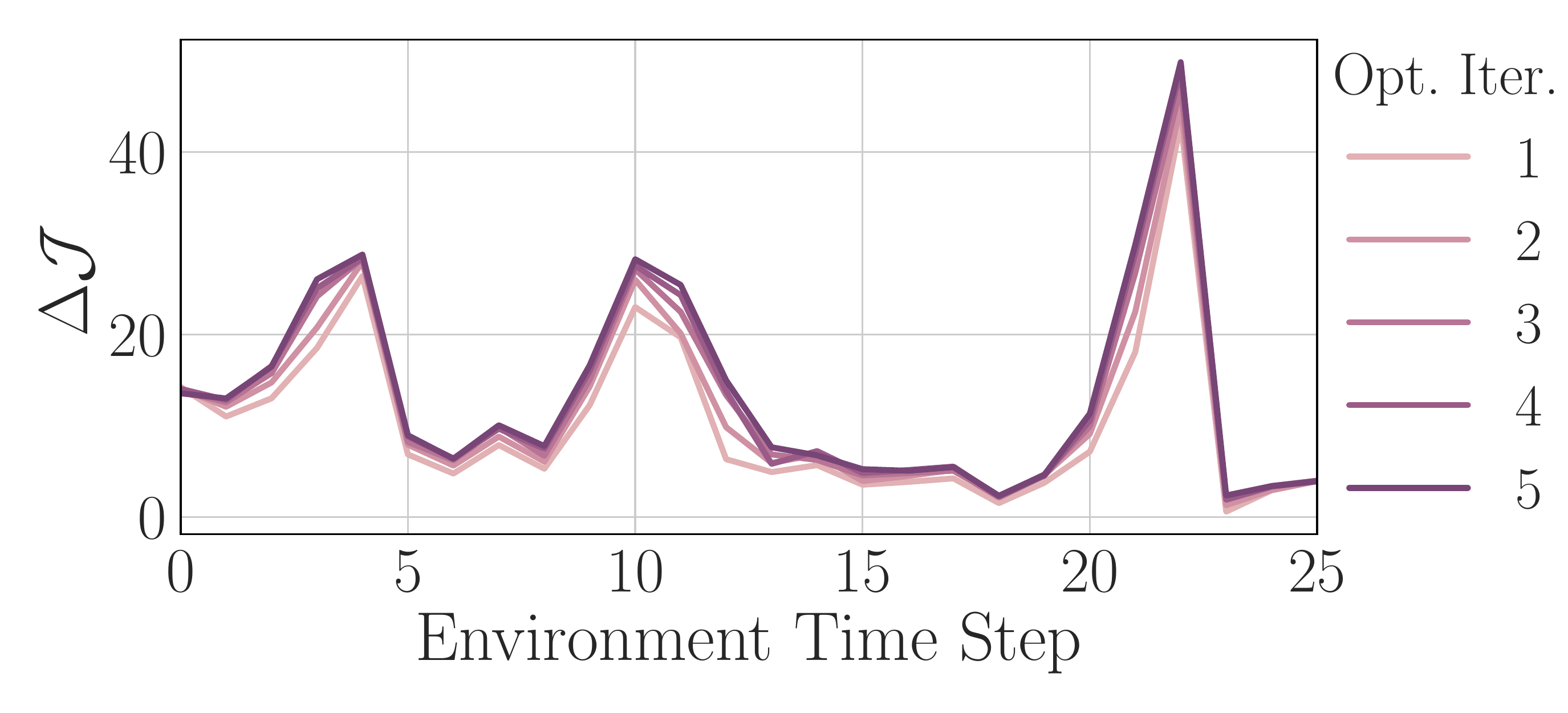}
        \caption{Improvement}
        \label{fig: iapo policy opt obj ant}
    \end{subfigure}% 
    ~
    \begin{subfigure}[t]{0.24\textwidth}
        \centering
        \includegraphics[width=\textwidth]{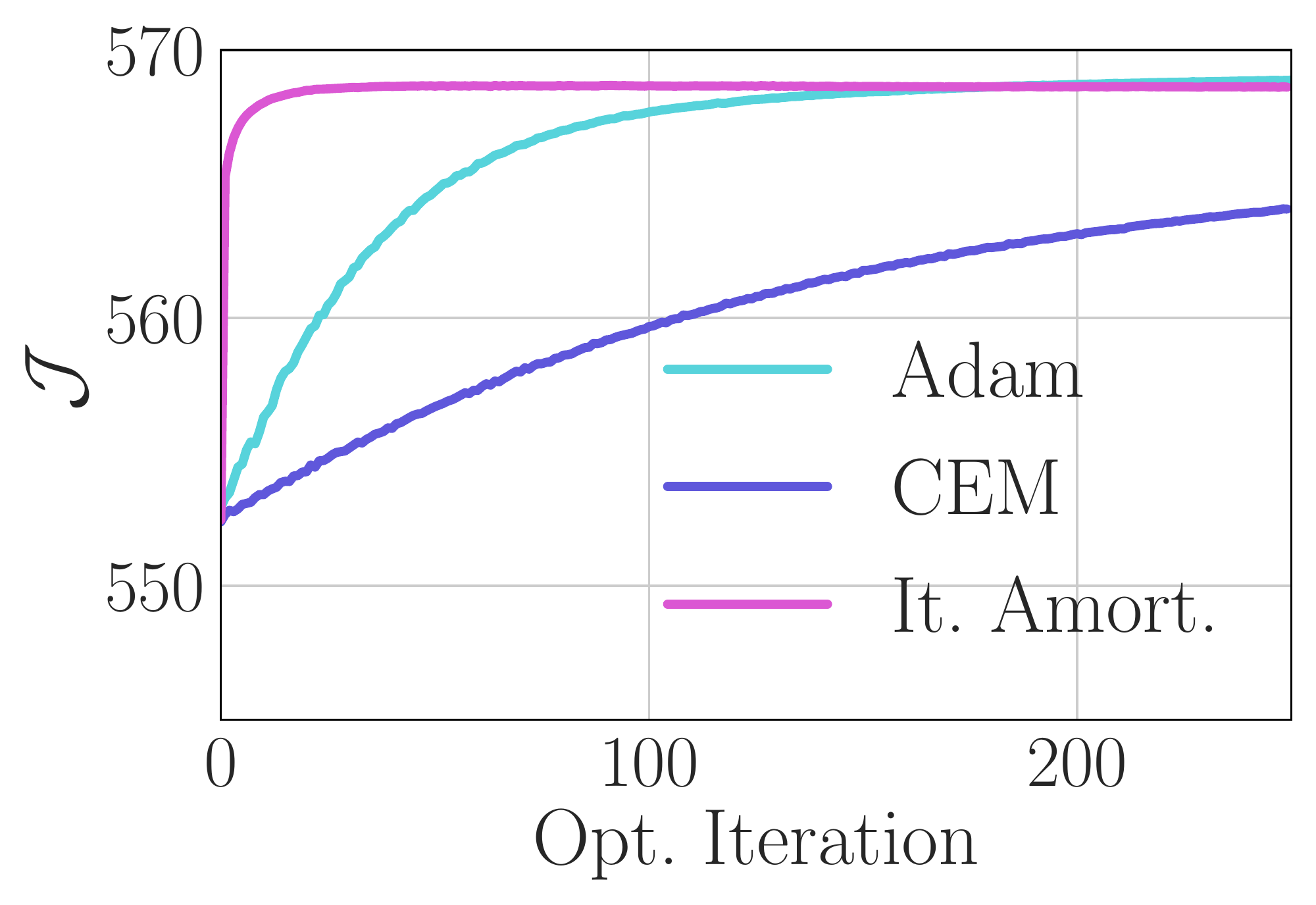}
        \caption{Comparison}
        \label{fig: iapo comp opt ant 2}
    \end{subfigure}
    \caption{\textbf{Policy Optimization}. Visualization over time steps of \textbf{(a)} one dimension of the policy distribution and \textbf{(b)} the improvement in the objective, $\Delta \mathcal{J}$, across policy optimization iterations. \textbf{(c)} Comparison of iterative amortization with Adam \citep{kingma2014adam} (gradient-based) and CEM \citep{rubinstein2013cross} (gradient-free). Iterative amortization is an order of magnitude more efficient.}
    \label{fig: iapo policy optimization}
\end{figure*}

% Appendix~\ref{appendix: value bias est}

% Discuss value overestimation issue. Discuss possible ways of mitigating. Describe pessimism solution. Show value estimate curves? Show resulting KL.

% Papers related to (mitigating) value overestimation:
% \begin{itemize}
%     \item Value overestimation due to value maximization \citep{thrun1993issues}
%     \item Conservative policy updates \citep{kakade2002approximately}
%     \item Double Q-network \citep{van2010double, van2016deep}
%     \item Min of Double Q-network \citep{fujimoto2018addressing}
%     \item Optimistic actor-critic \citep{ciosek2019better}
%     \item Temperature adjustment based on advantage variance \citep{fox2019toward}
%     \item Weighted Q-Learning \citep{cini2020deep}
%     \item Game-theoretic formulation \citep{rajeswaran2020game}
% \end{itemize}

% Papers related specifically to offline RL:
% \begin{itemize}
%     \item BCQ \citep{fujimoto2019off}
%     \item BEAR \citep{kumar2019stabilizing}
%     \item BRAC \citep{wu2019behavior}
%     \item Random Ensemble Mixture \citep{agarwal2019optimistic}
%     \item Conservative Q-Learning \citep{kumar2020conservative}
% \end{itemize}
\section{Experiments}
\label{sec: iapo experiments}

\subsection{Setup}

To focus on policy optimization, we implement iterative amortized policy optimization using the soft actor-critic (SAC) setup described by Haarnoja et al. \cite{haarnoja2018soft2}. This uses two $Q$-networks, uniform action prior, $p_\theta (\mathbf{a} | \mathbf{s}) = \mathcal{U}(-1, 1)$, and a tuning scheme for the temperature, $\alpha$. In our experiments, ``direct'' refers to direct amortization employed in SAC, i.e., a direct policy network, and ``iterative'' refers to iterative amortization. Both approaches use the \textit{same} network architecture, adjusting only the number of inputs and outputs to accommodate gradients, current policy estimates, and gated updates (Sec.~\ref{sec: iapo formulation}). Unless otherwise stated, we use $5$ iterations per time step for iterative amortization, following \cite{marino2018iterative}. For details, refer to Appendix \ref{appendix: experiment details} and Haarnoja et al. \cite{haarnoja2018soft, haarnoja2018soft2}.

% Appendix~\ref{appendix: experiment details}
% Link to code...and results?

\subsection{Analysis}

\subsubsection{Visualizing Policy Optimization}

We provide 2D visualizations of iterative amortized policy optimization in Figures \ref{fig: iapo amortization} \& \ref{fig: iapo multi-modal sampling}, with additional 2D plots in Appendix \ref{sec: iapo additional 2d opt plots}. In Figure~\ref{fig: iapo policy optimization}, we visualize iterative refinement using a single action dimension from \texttt{Ant-v2} across time steps. The refinements in Figure~\ref{fig: iapo policy opt single dim ant} give rise to the objective improvements in Figure~\ref{fig: iapo policy opt obj ant}. We compare with Adam \citep{kingma2014adam} (gradient-based) and CEM \citep{rubinstein2013cross} (gradient-free) in Figure~\ref{fig: iapo comp opt ant 2}, where iterative amortization is \textit{an order of magnitude} more efficient. This trend is consistent across environments, as shown in Appendix \ref{sec: iapo comparison with iterative optimizers}.

% Appendix~\ref{sec: iapo comparison with iterative optimizers}

\subsubsection{Performance Comparison}
\label{sec: iapo mf perf comp}

\begin{figure}[t!]
    \centering
    \begin{subfigure}[t]{0.22\textwidth}
        \centering
        \includegraphics[width=\textwidth]{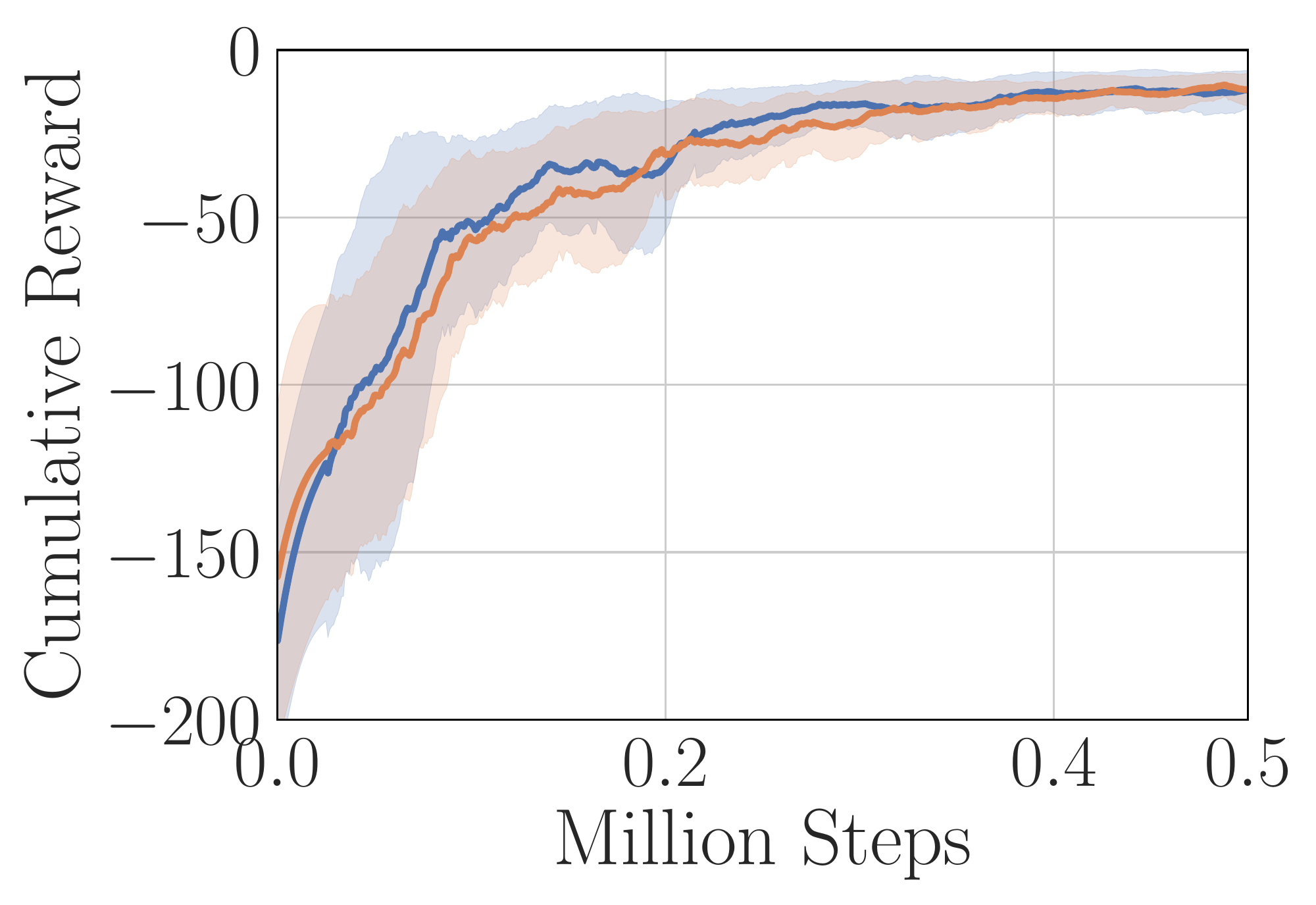}
        \caption{\scriptsize \texttt{Reacher-v2}}
    \end{subfigure}%
    ~
    \begin{subfigure}[t]{0.22\textwidth}
        \centering
        \includegraphics[width=\textwidth]{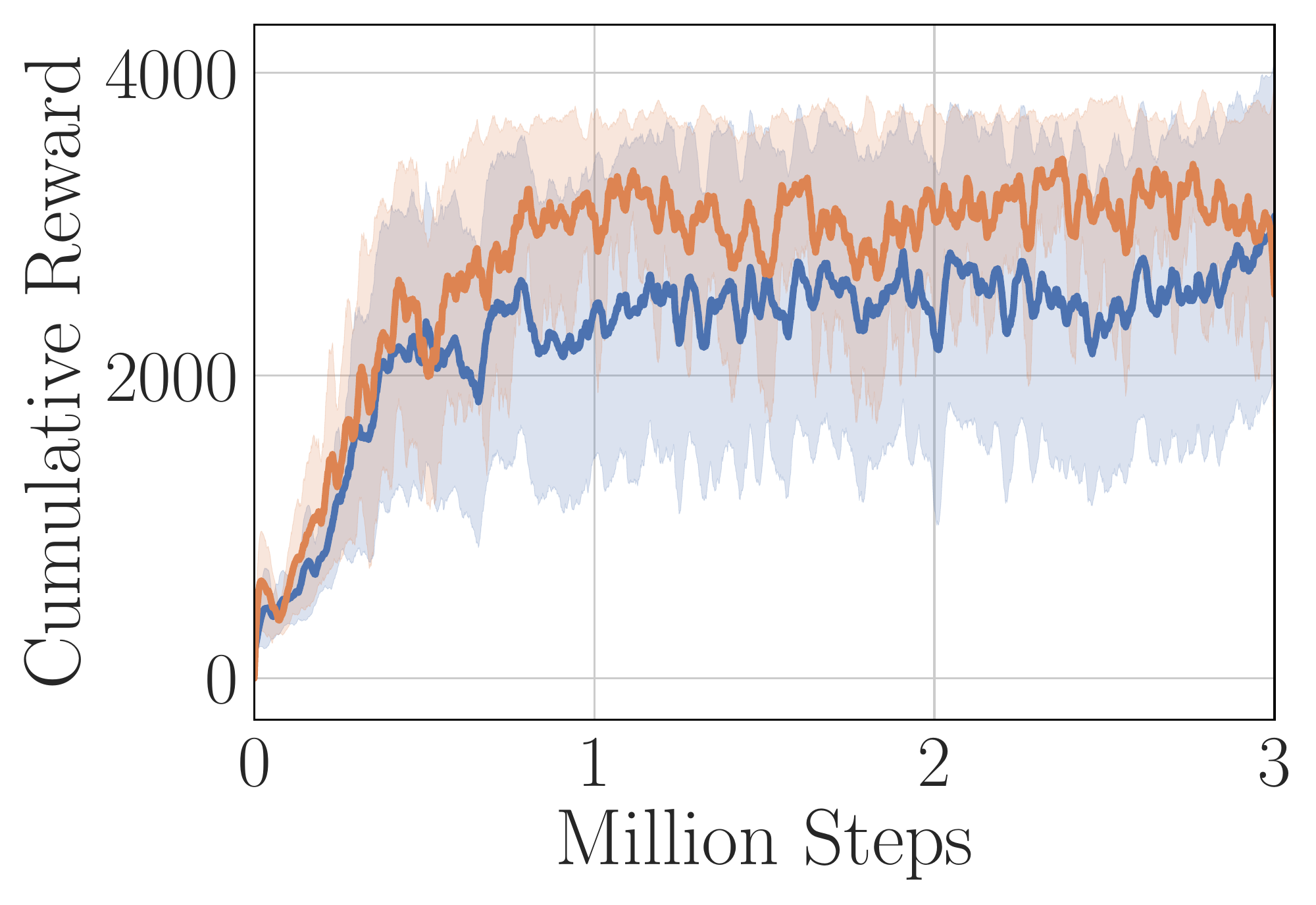}
        \caption{\scriptsize \texttt{Hopper-v2}}
    \end{subfigure}%
    ~ 
    \begin{subfigure}[t]{0.22\textwidth}
        \centering
        \includegraphics[width=\textwidth]{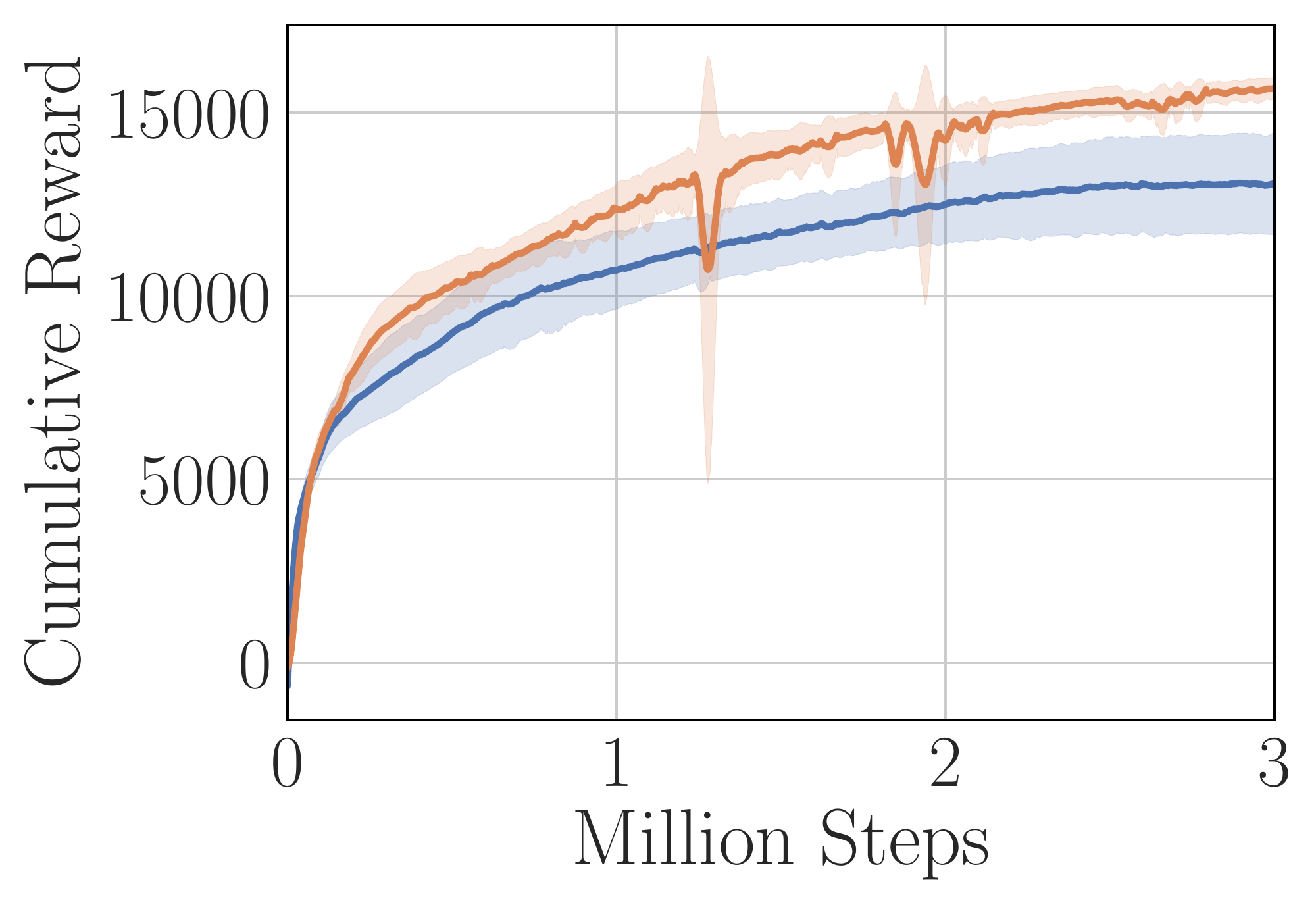}
        \caption{\scriptsize \texttt{HalfCheetah-v2}}
        \label{fig: iapo mf perf halfcheetah}
    \end{subfigure}%
    ~ 
    \begin{subfigure}[t]{0.22\textwidth}
        \centering
        \includegraphics[width=\textwidth]{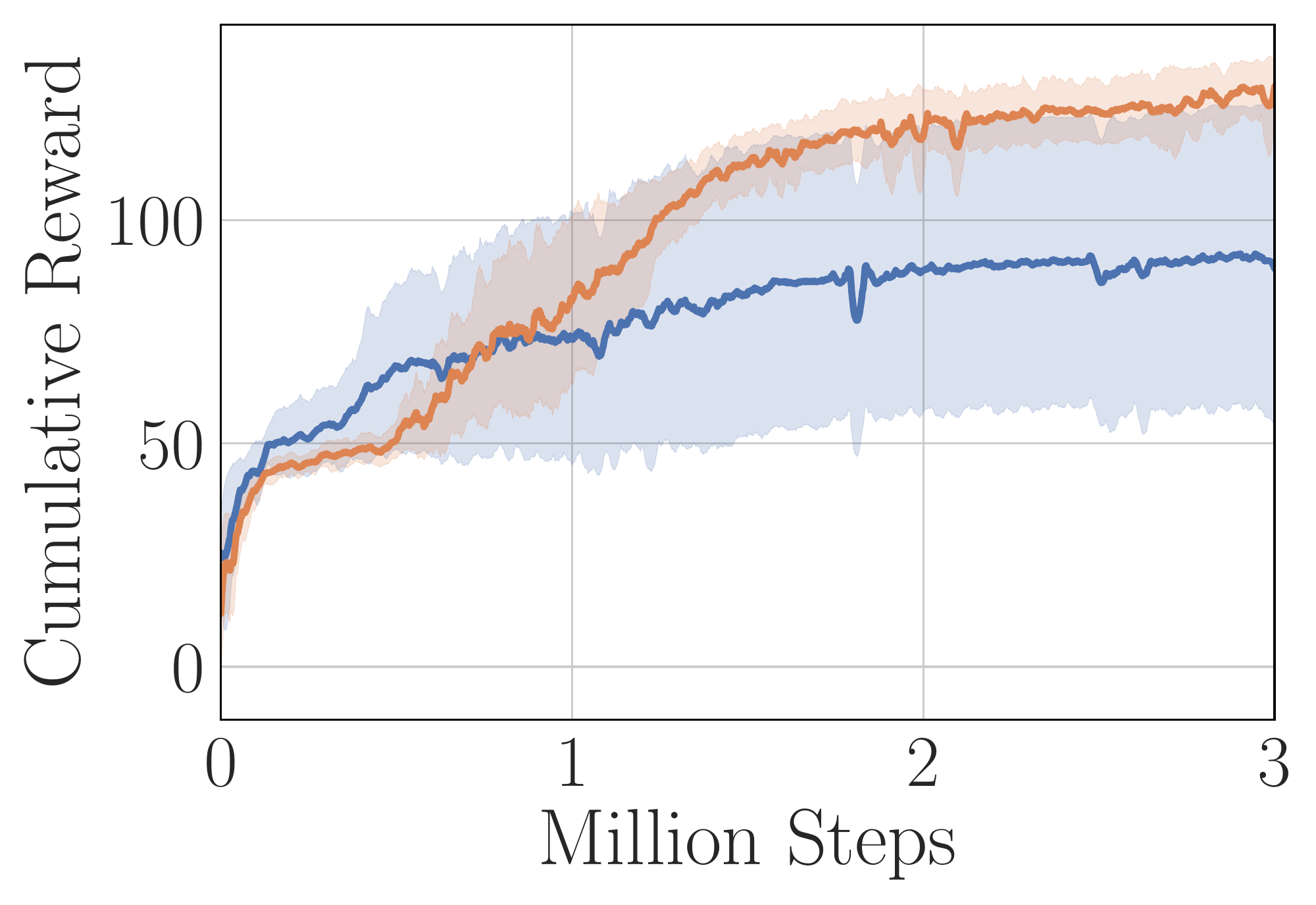}
        \caption{\scriptsize \texttt{Swimmer-v2}}
    \end{subfigure}
    
    \begin{subfigure}[t]{0.22\textwidth}
        \centering
        \includegraphics[width=\textwidth]{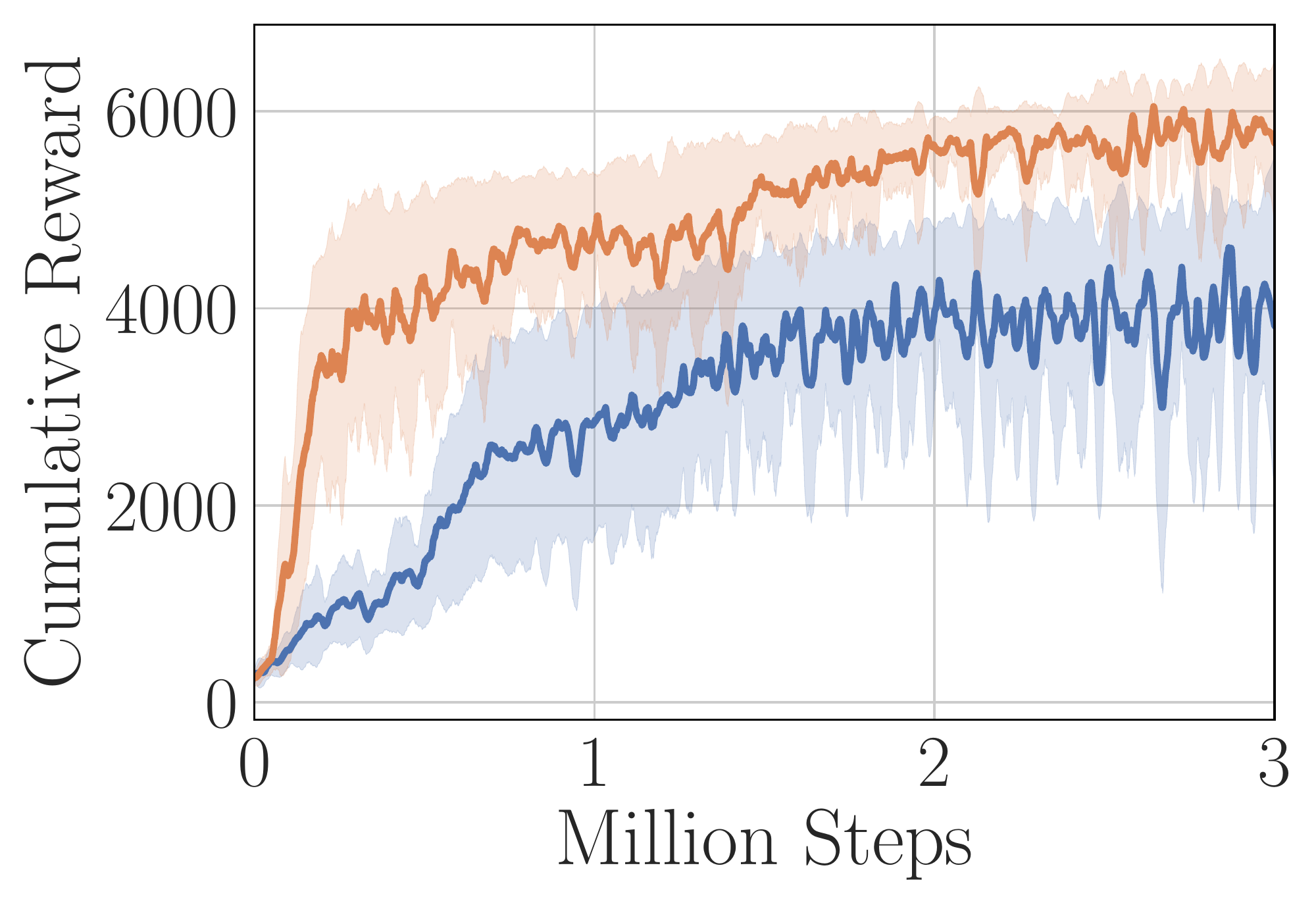}
        \caption{\scriptsize \texttt{Walker2d-v2}}
    \end{subfigure}%
    ~
    \begin{subfigure}[t]{0.22\textwidth}
        \centering
        \includegraphics[width=\textwidth]{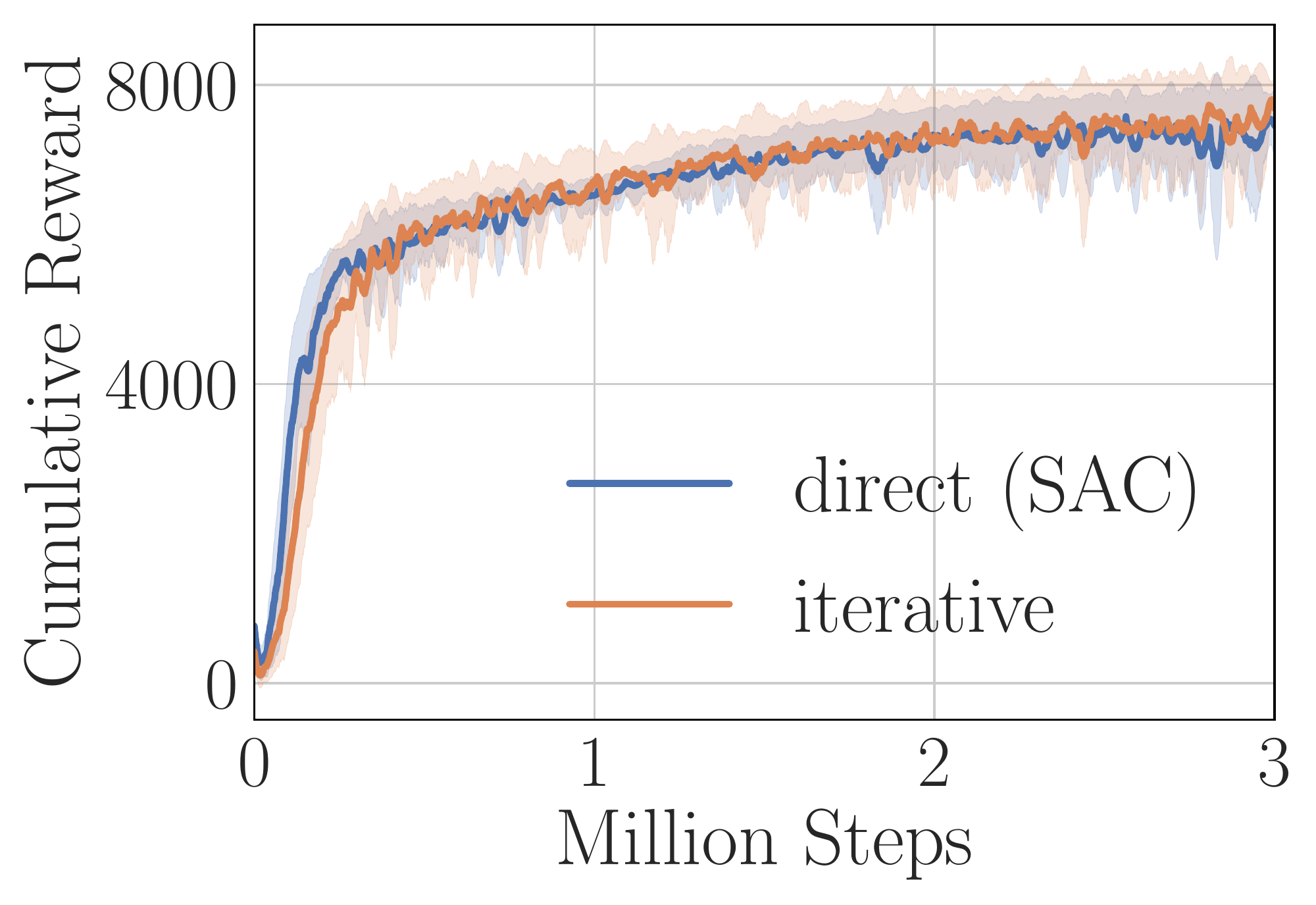}
        \caption{\scriptsize \texttt{Ant-v2}}
    \end{subfigure}%
    ~ 
    \begin{subfigure}[t]{0.22\textwidth}
        \centering
        \includegraphics[width=\textwidth]{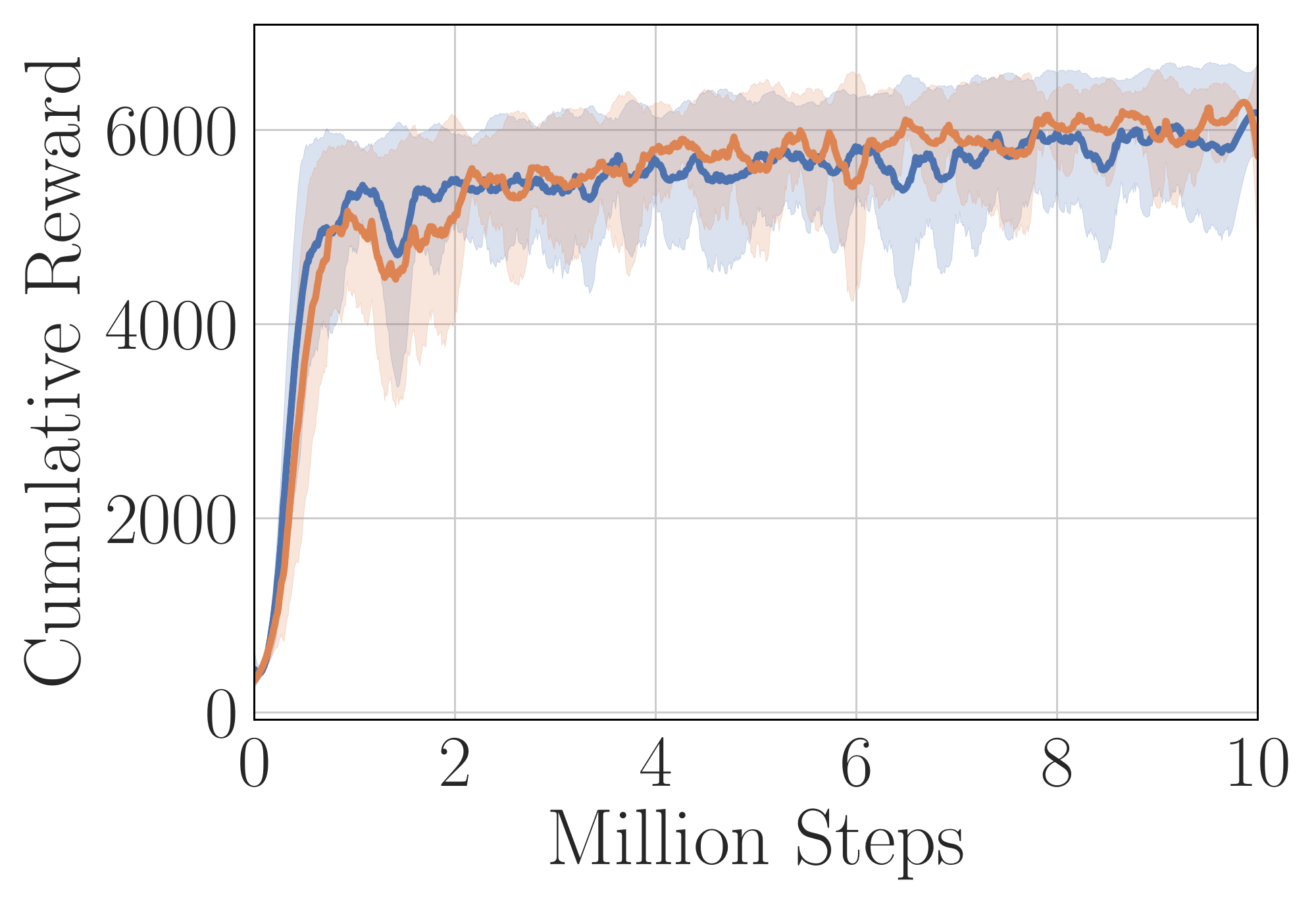}
        \caption{\scriptsize \texttt{Humanoid-v2}}
    \end{subfigure}%
    ~ 
    \begin{subfigure}[t]{0.22\textwidth}
        \centering
        \includegraphics[width=\textwidth]{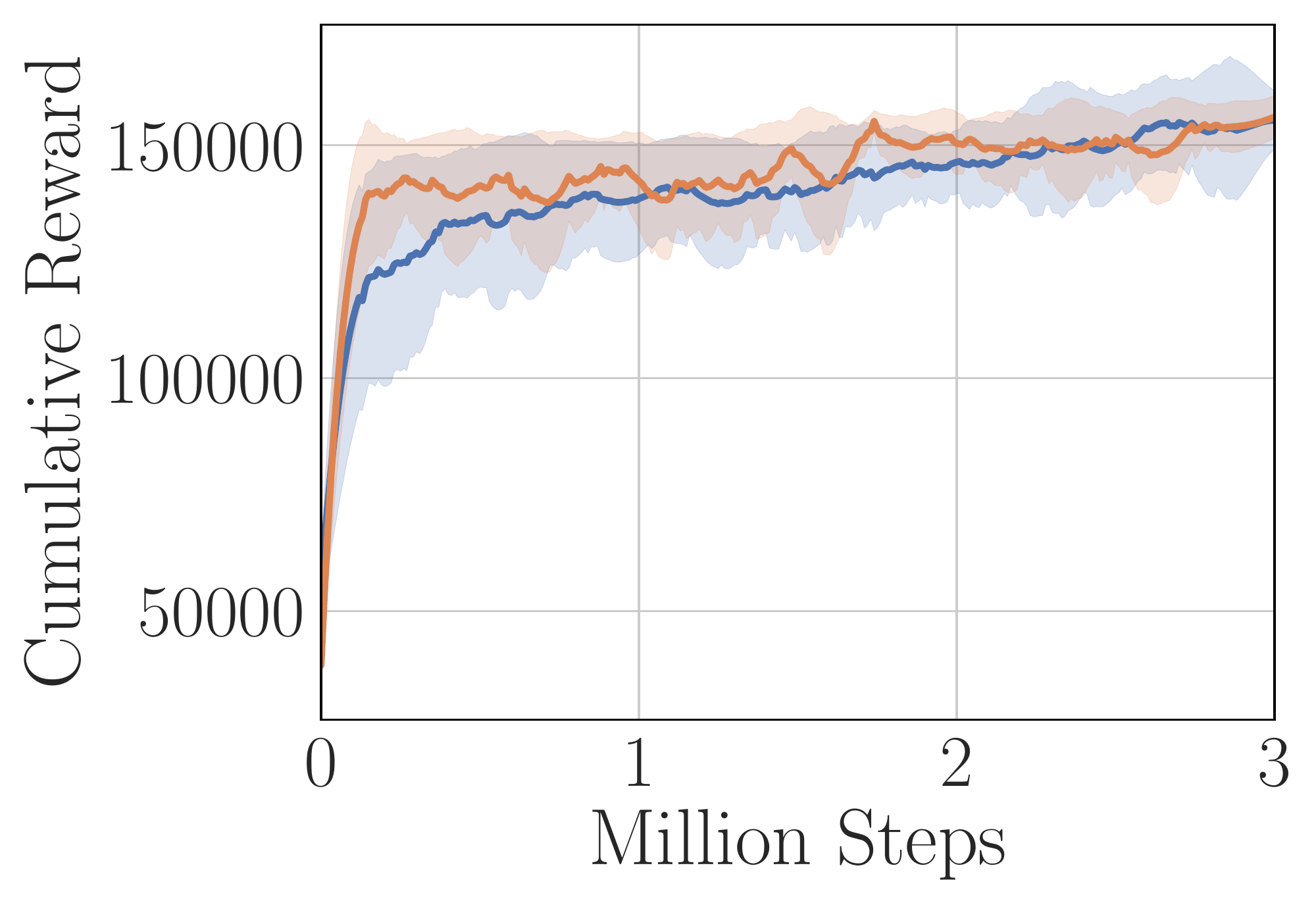}
        \caption{\scriptsize \texttt{HumanoidStandup-v2}}
    \end{subfigure}
    \caption{\textbf{Performance Comparison}. Iterative amortized policy optimization performs comparably with or better than direct amortization across MuJoCo environments. Curves show the mean $\pm$ std.~dev.~of performance over $5$ random seeds.}
    \label{fig: model-free performance}
\end{figure}

We evaluate iterative amortized policy optimization on the suite of MuJoCo \citep{todorov2012mujoco} continuous control tasks from OpenAI gym \citep{brockman2016openai}. In Figure~\ref{fig: model-free performance}, we compare the cumulative reward of direct and iterative amortized policy optimization across environments. Each curve shows the mean and $\pm$ standard deviation of $5$ random seeds. In all cases, iterative amortized policy optimization matches or outperforms the baseline direct amortized method, both in sample efficiency and final performance. Iterative amortization also yields more consistent, lower variance performance. 

\subsubsection{Improved Exploration: Multiple Policy Modes}

As described in Section \ref{sec: iapo benefits}, iterative amortization is capable of obtaining multiple estimates, i.e., multiple modes of the optimization objective. To confirm that iterative amortization has captured multiple modes, at the end of training, we take an iterative agent trained on \texttt{Walker2d-v2} and histogram the distances between policy means across separate runs of policy optimization per state (Fig.~\ref{fig: iapo mean distance walker2d}). For the state with the largest distance, we plot 2D projections of the optimization objective, $\mathcal{J}$, across action dimensions in Figure \ref{fig: iapo multi dim opt walker2d}, as well as the policy density across $10$ optimization runs (Fig.~\ref{fig: iapo multi dim density walker2d}). The multi-modal policy optimization surface shown in Figure \ref{fig: iapo multi dim opt walker2d} results in the multi-modal policy in Figure \ref{fig: iapo multi dim density walker2d}. Additional results on other environments are presented in Appendix \ref{appendix: multiple policy estimates}.

To better understand whether the performance benefits of iterative amortization are coming purely from improved exploration via multiple modes, we also compare with direct amortization with a multi-modal policy distribution. This is formed using inverse autoregressive flows \citep{kingma2016improved}, a type of normalizing flow (NF). Results are presented in Appendix \ref{appendix: comp with nf policies}. Using a multi-modal policy reduces the performance deficiencies on \texttt{Hopper-v2} and \texttt{Walker2d-v2}, indicating that much of the benefit of iterative amortization is due to lifting direct amortization's restriction to a single, uni-modal policy estimate. Yet, direct + NF still struggles on \texttt{HalfCheetah-v2} compared with iterative amortization, suggesting that more complex, multi-modal distributions are not the \textit{only} consideration.

\begin{figure}[t!]
    \centering
    \begin{subfigure}[t]{0.22\textwidth}
        \centering
        \includegraphics[width=\textwidth]{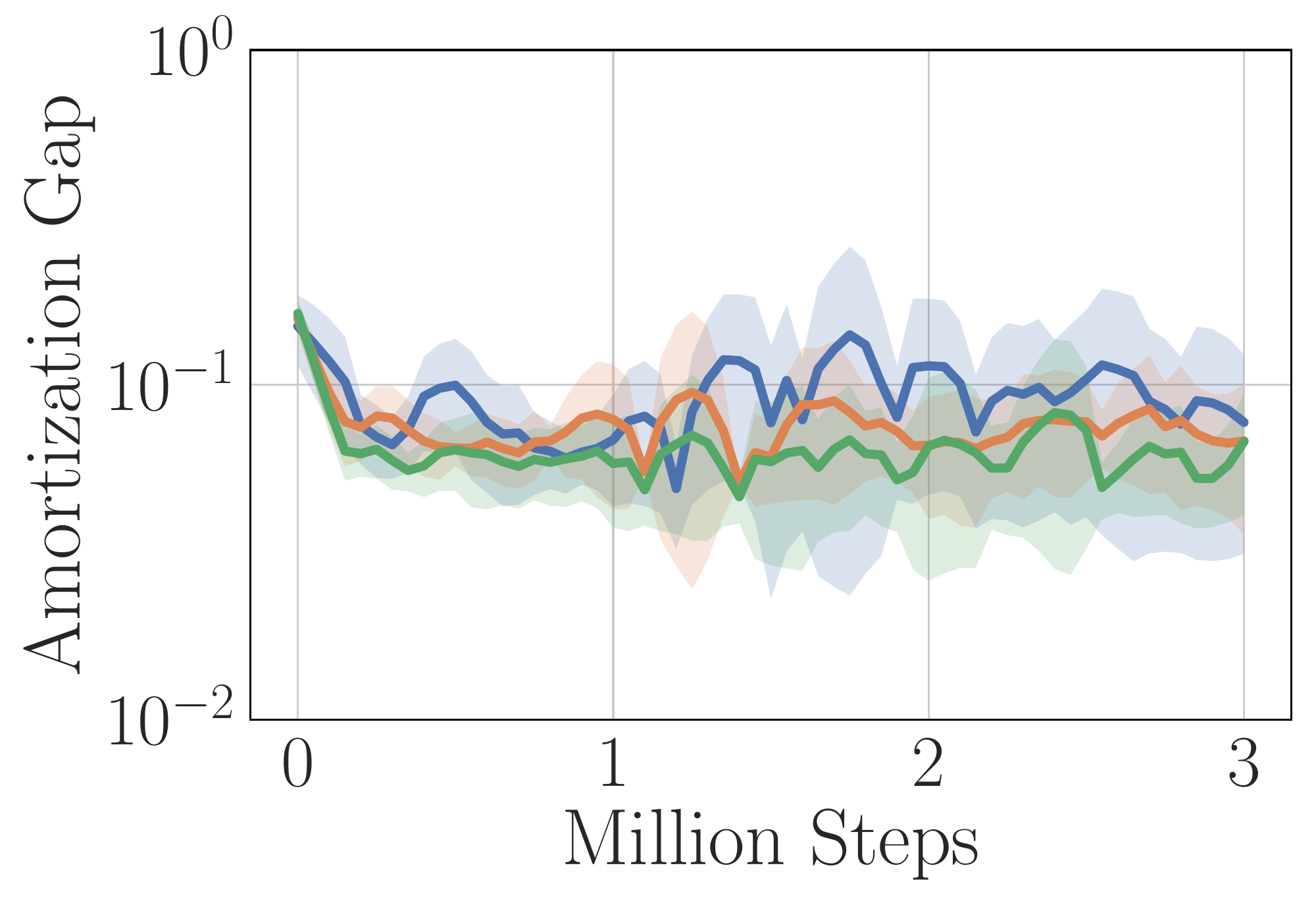}
        \caption{\scriptsize \texttt{Hopper-v2}}
    \end{subfigure}%
    ~ 
    \begin{subfigure}[t]{0.22\textwidth}
        \centering
        \includegraphics[width=\textwidth]{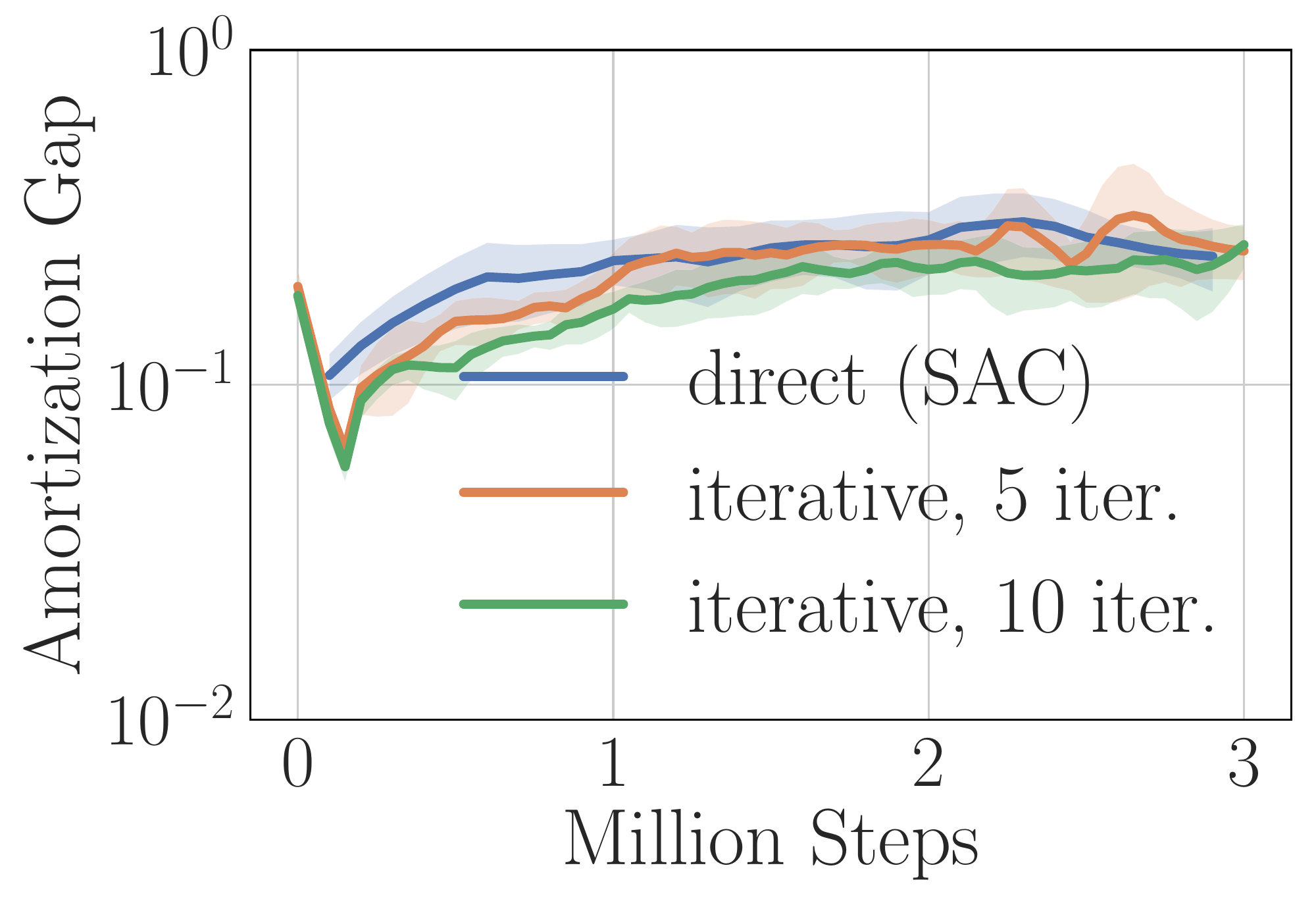}
        \caption{\scriptsize \texttt{HalfCheetah-v2}}
    \end{subfigure}%
    ~ 
    \begin{subfigure}[t]{0.22\textwidth}
        \centering
        \includegraphics[width=\textwidth]{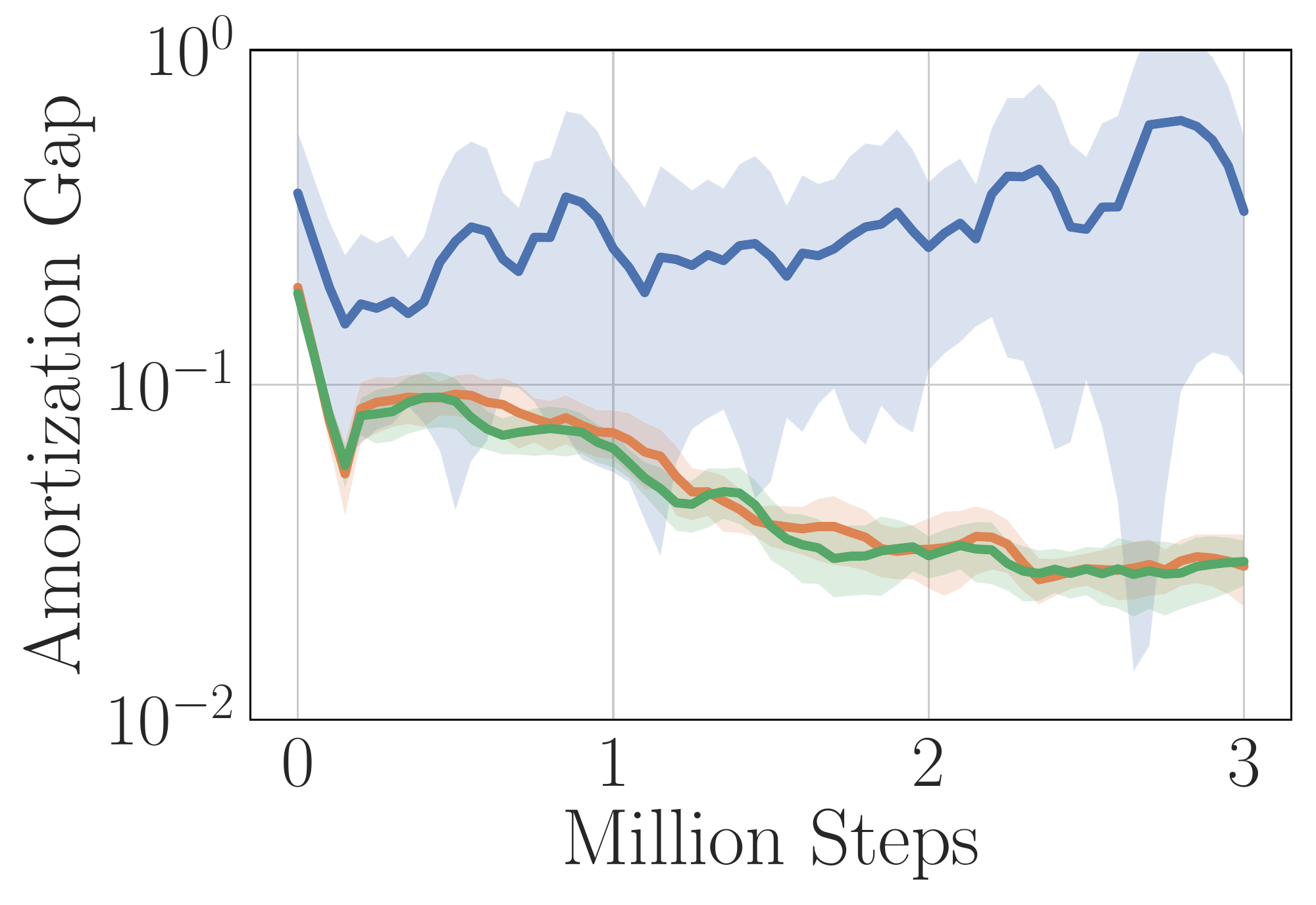}
        \caption{\scriptsize \texttt{Walker2d-v2}}
    \end{subfigure}%
    ~ 
    \begin{subfigure}[t]{0.22\textwidth}
        \centering
        \includegraphics[width=\textwidth]{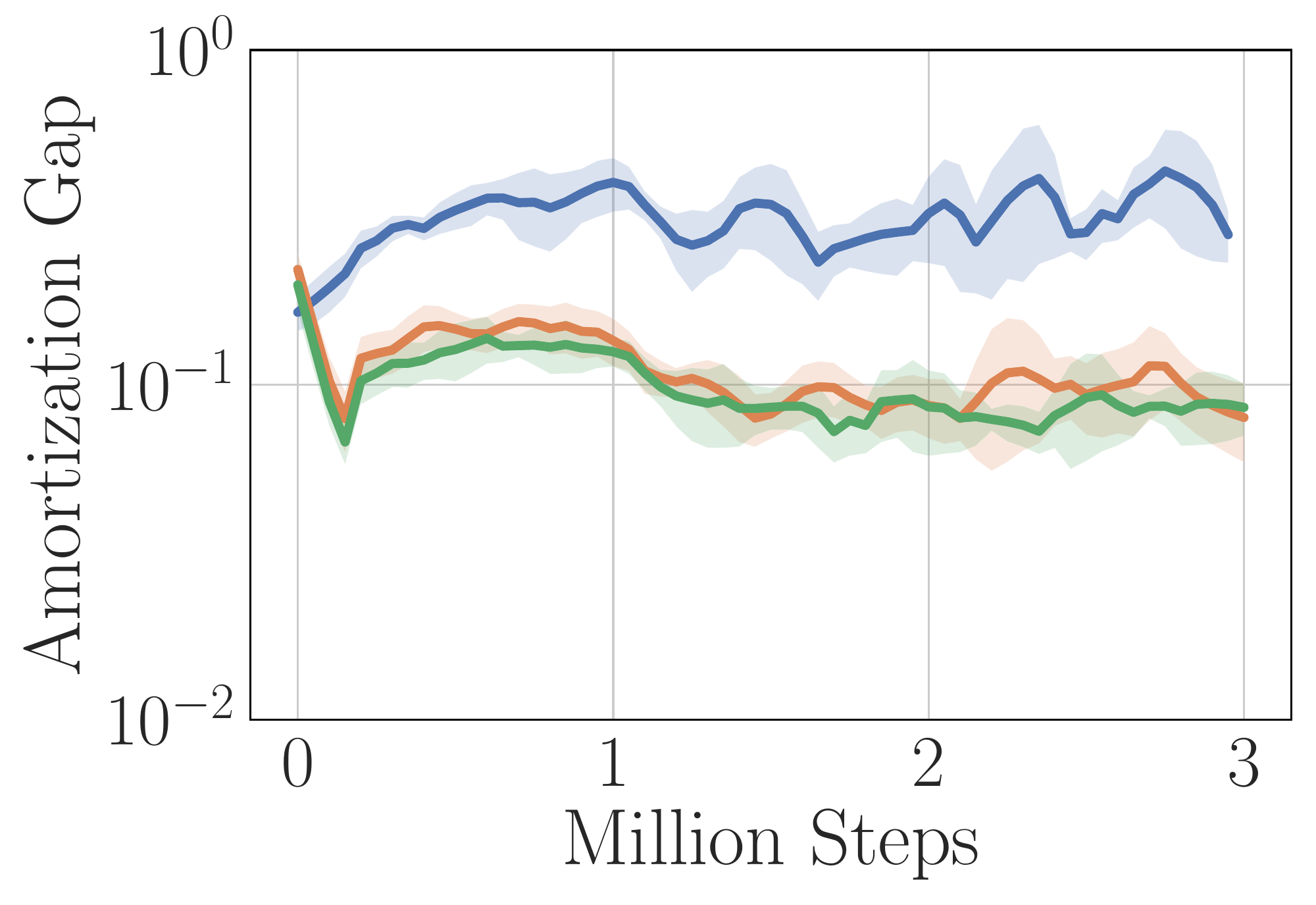}
        \caption{\scriptsize \texttt{Ant-v2}}
    \end{subfigure}
    \caption{\textbf{Amortization Gap}. Iterative amortization achieves similar or lower amortization gaps than direct amortization. Gaps are estimated using stochastic gradient-based optimization over $100$ random states. Curves show the mean and $\pm$ std.~dev.~over $5$ random seeds.}
    \label{fig: iapo amortization gap}
\end{figure}

\subsubsection{Improved Optimization: Amortization Gap}
\label{sec: iapo decreased am gap}

To evaluate policy optimization accuracy, we estimate per-step amortization gaps, performing additional iterations of gradient ascent on $\mathcal{J}$ w.r.t.~the policy parameters, $\bm{\lambda} \equiv \left[ \bm{\mu}, \bm{\sigma} \right]$ (see Appendix \ref{appendix: am gap calc}). To analyze generalization, we also evaluate the iterative agents trained with $5$ iterations for an additional $5$ amortized iterations. Results are shown in Figure~\ref{fig: iapo amortization gap}. We emphasize that it is challenging to \textit{directly} compare amortization gaps across optimization schemes, as these involve different value functions, and therefore different objectives. Likewise, we estimate the amortization gap using the learned $Q$-networks, which may be biased (Figure~\ref{fig: value overestimation}). Nevertheless, we find that iterative amortized policy optimization achieves, on average, lower amortization gaps than direct amortization across all environments. Additional amortized iterations at evaluation yield further estimated improvement, demonstrating generalization beyond the optimization horizon used during training.

The amortization gaps are small relative to the objective, playing a negligible role in \textit{evaluation} performance (see Appendix \ref{sec: iapo additional iters}). Rather, improved policy optimization is helpful for \textit{training}, allowing the agent to explore states where value estimates are highest. To probe this further, we train iterative amortized policy optimization while varying the number of iterations per step in $\{ 1, 2, 5 \}$, yielding optimizers with varying degrees of accuracy. Note that each optimizer is, in theory, capable of finding multiple modes. In Figure \ref{fig: iapo train iters perf am gap halfcheetah}, we see that training with additional iterations improves performance and optimization accuracy. We stress that the exact form of this relationship depends on the $Q$-value estimator and other factors. We present additional results in Appendix \ref{sec: iapo additional iters}.

\begin{figure}[t!]
    \centering
    \begin{subfigure}[t]{0.22\textwidth}
    \centering
    \includegraphics[width=\textwidth]{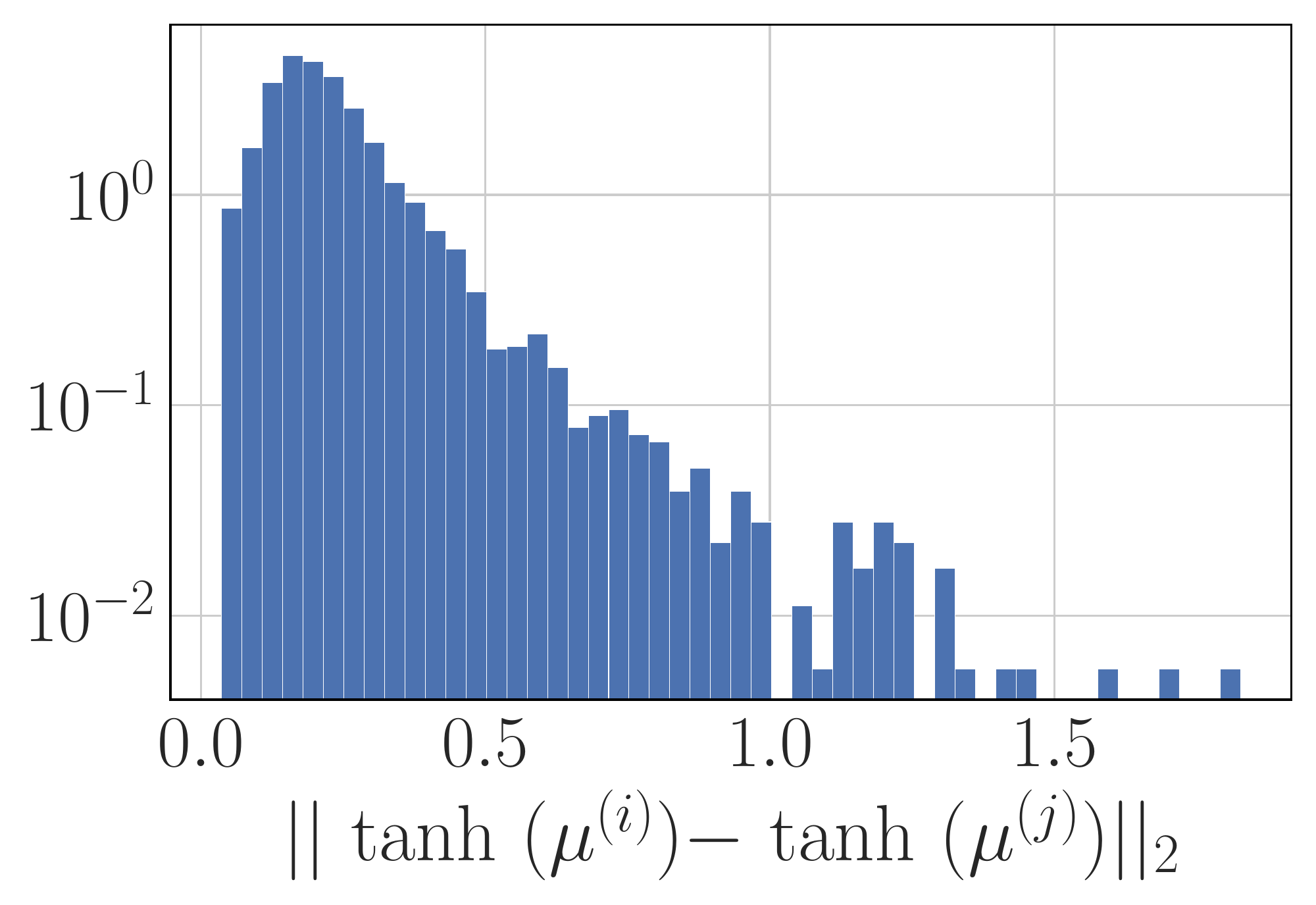}
    \caption{}
    \label{fig: iapo mean distance walker2d}
    \end{subfigure} 
    \begin{subfigure}[t]{0.20\textwidth}
    \centering
    \includegraphics[width=\textwidth]{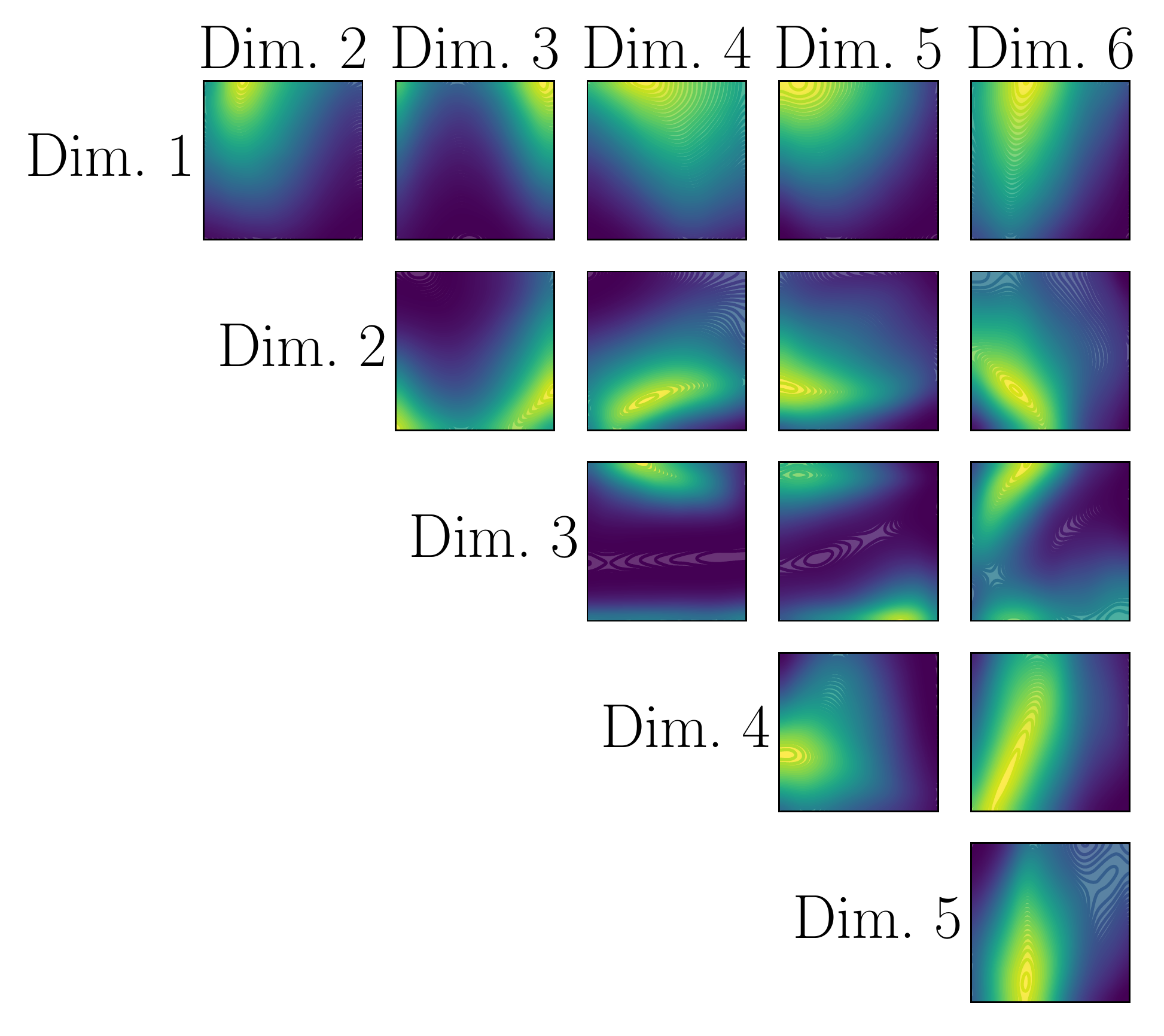}
    \caption{}
    \label{fig: iapo multi dim opt walker2d}
    \end{subfigure}
    \begin{subfigure}[t]{0.52\textwidth}
    \centering
    \includegraphics[width=\textwidth]{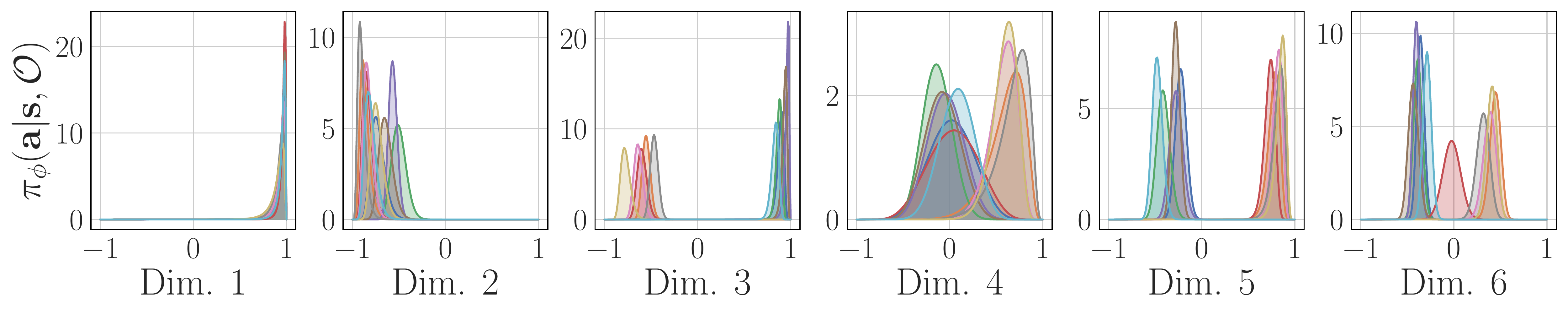}
    \caption{}
    \label{fig: iapo multi dim density walker2d}
    \end{subfigure}
    \caption{\textbf{Multiple Policy Modes}. \textbf{(a)} Histogram of distances between policy means ($\bm{\mu}$) across optimization runs ($i$ and $j$) over seeds and states on \texttt{Walker2d-v2} at $3$ million environment steps. For the state with the largest distance, \textbf{(b)} shows the projected optimization surface on each pair of action dimensions, and  \textbf{(c)} shows the policy density for $10$ optimization runs.}
    \label{fig: iapo multiples modes experiment}
\end{figure}

\begin{figure}
\centering
\begin{minipage}{.65\textwidth}
  \centering
  \begin{subfigure}[t]{0.37\textwidth}
        \centering
        \includegraphics[width=\textwidth]{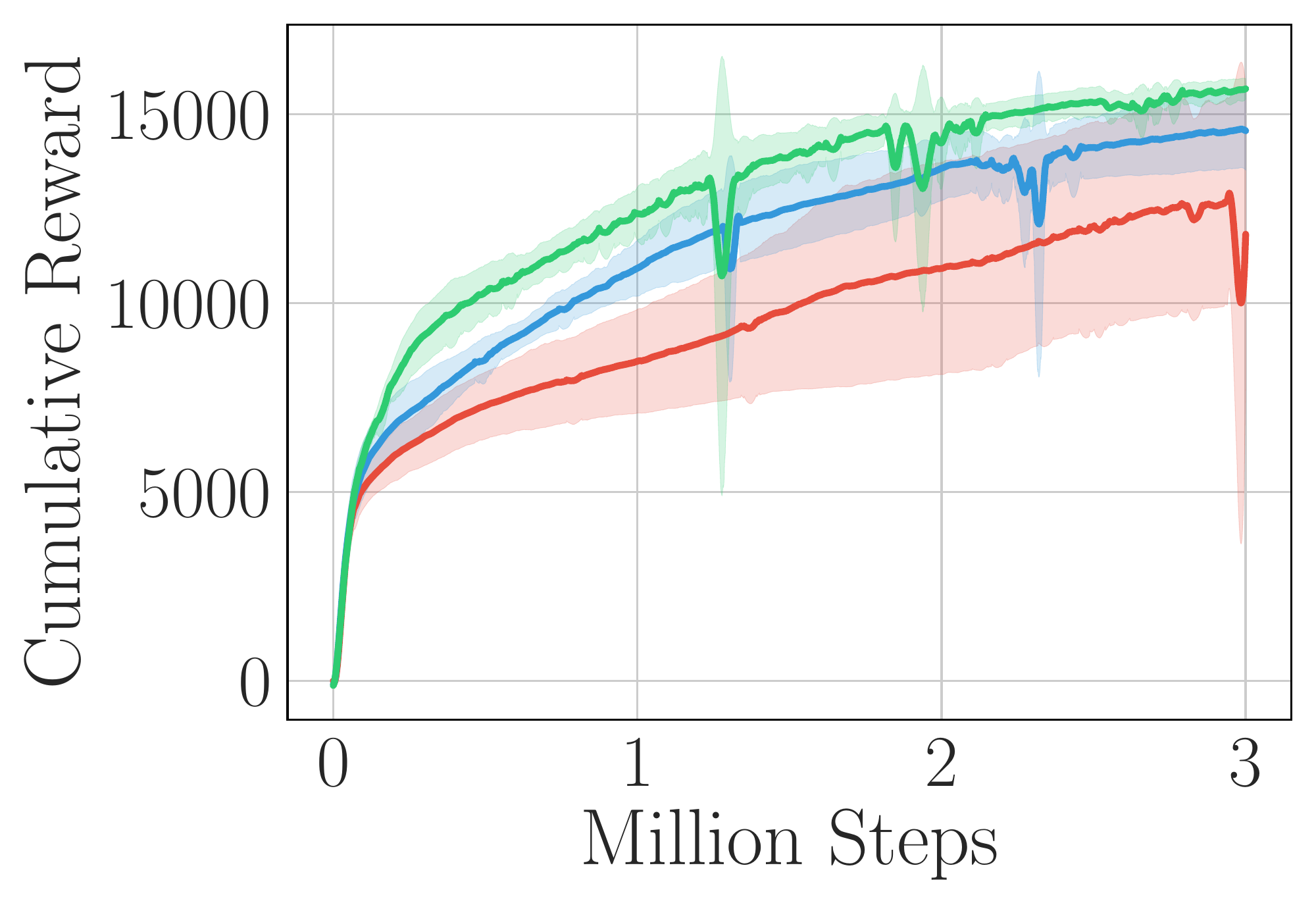}
        % \caption{}
    \end{subfigure}
    ~
    \begin{subfigure}[t]{0.37\textwidth}
        \centering
        \includegraphics[width=\textwidth]{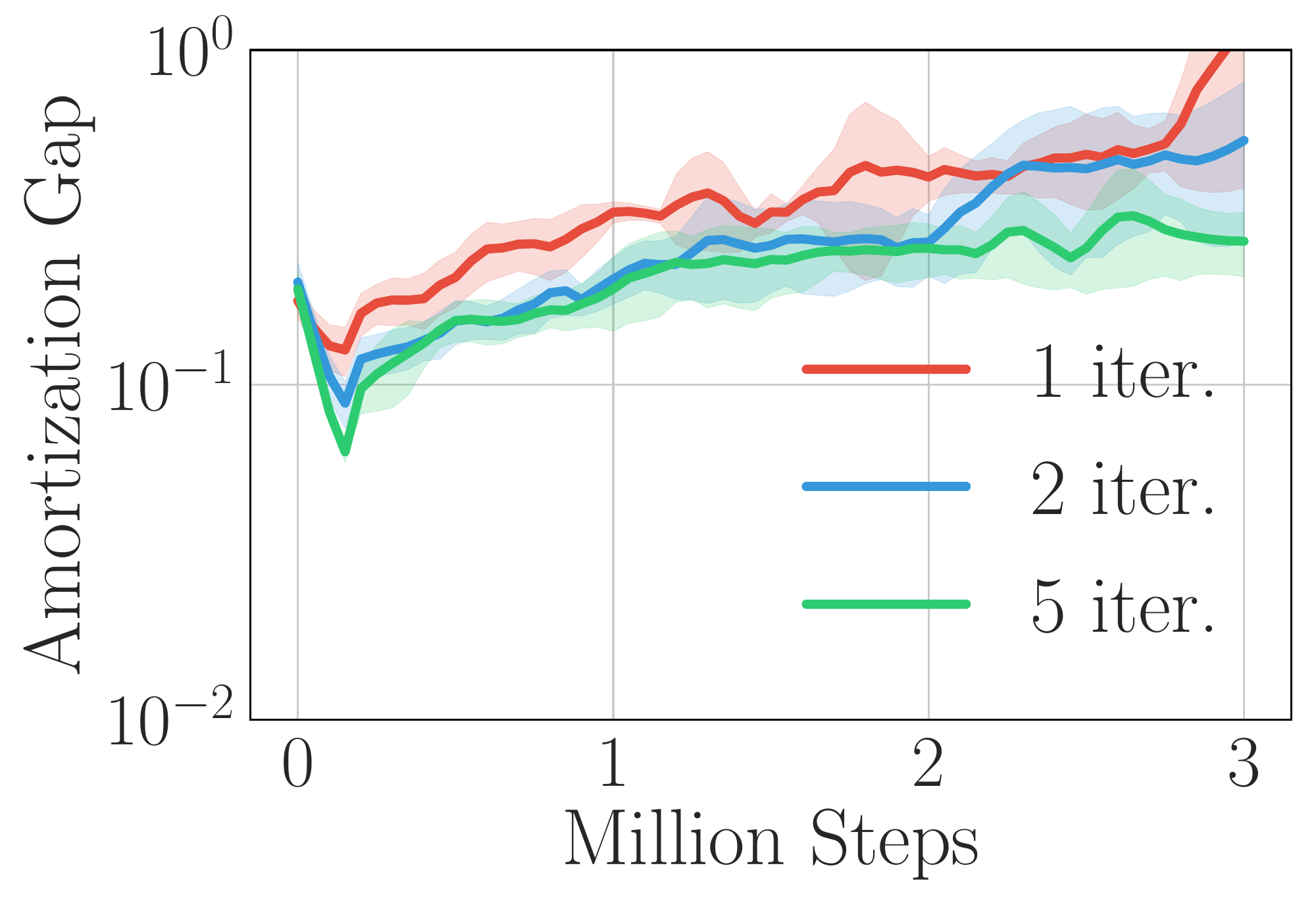}
        % \caption{}
    \end{subfigure}
    \caption{\textbf{Varying Iterations During Training}. Performance (\textbf{Left}) and estimated amortization gap (\textbf{Right}) for varying numbers of policy optimization iterations per step during training on \texttt{HalfCheetah-v2}. Increasing the iterations generally improves performance and decreases the estimated amortization gap.}
    \label{fig: iapo train iters perf am gap halfcheetah}
\end{minipage}% 
~ \thickspace \thickspace \thickspace
\begin{minipage}{.3\textwidth}
  \centering
  \includegraphics[width=0.83\textwidth]{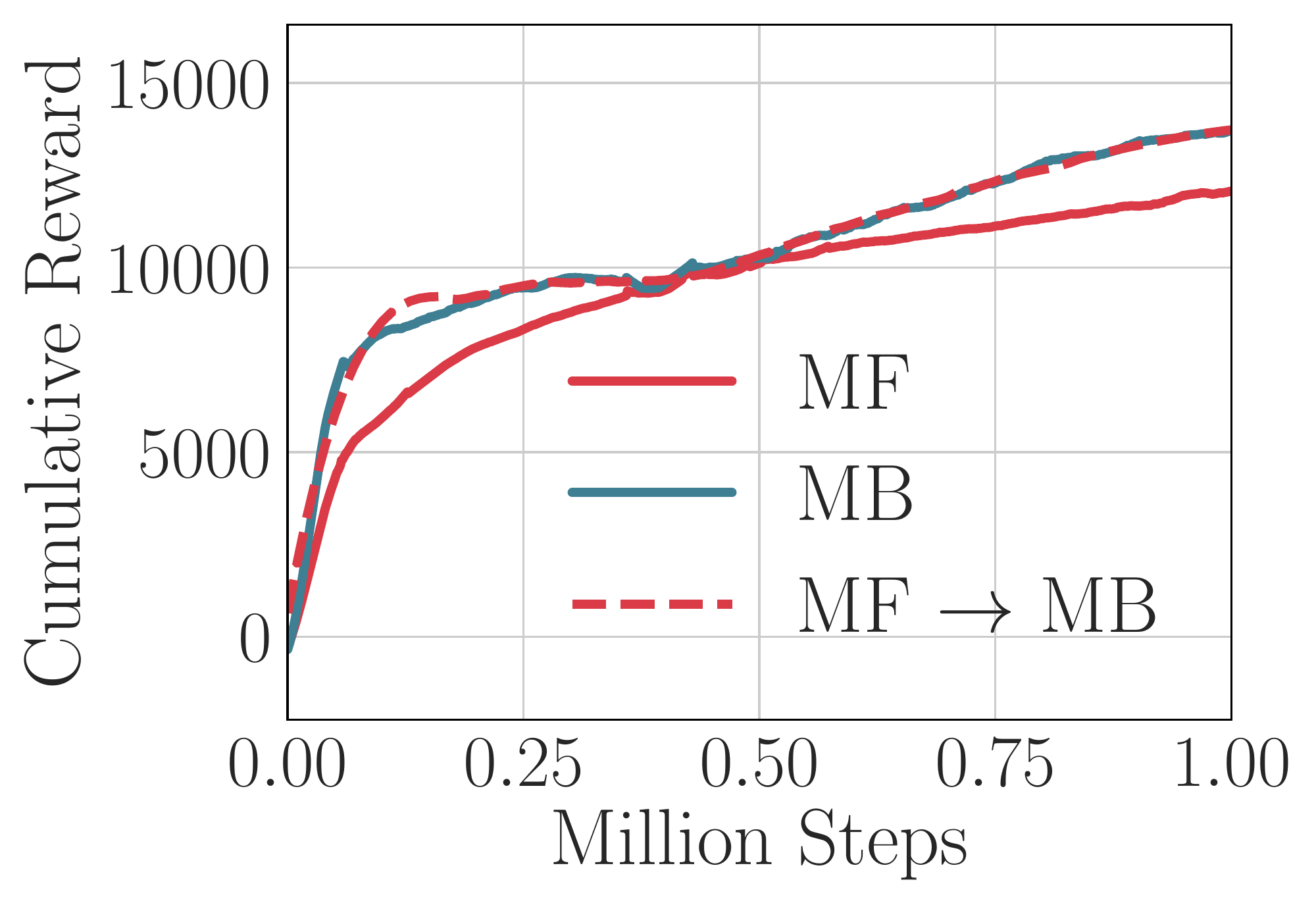}
    \caption{\textbf{Zero-shot generalization} of iterative amortization from model-free (MF) to model-based (MB) value estimates.}
    \label{fig: iapo mb transfer}
\end{minipage}
\end{figure}

\subsubsection{Generalizing to Model-Based Value Estimates}

Direct amortization is a purely feedforward process and is therefore incapable of generalizing to new objectives. In contrast, because iterative amortization is formulated through gradient-based feedback, such optimizers may be capable of generalizing to new objective estimators, as shown in Figure~\ref{fig: iapo amortization}. To demonstrate this capability further, we apply iterative amortization with model-based value estimators, using a learned deterministic model on \texttt{HalfCheetah-v2} (see Appendix \ref{appendix: model-based value estimation}). We evaluate the generalizing capabilities in Figure~\ref{fig: iapo mb transfer} by transferring the policy optimizer from a model-free agent to a model-based agent. Iterative amortization generalizes to these new value estimates, \textit{instantly} recovering the performance of the model-based agent. This highlights the opportunity for instantly incorporating new tasks, goals, or model estimates into policy optimization.

% While our analysis has centered on the model-free setting, iterative amortized policy optimization can also be applied to model-based value estimates. As model-based RL remains an active research area \citep{janner2019trust}, we provide a proof-of-concept in this setting, using a learned deterministic model on \texttt{HalfCheetah-v2} (see Appendix A.5). As shown in Figure~\ref{fig: iapo mb perf}, iterative amortization outperforms direct amortization in this setting. Iterative amortization refines planned trajectories, shown for a single state dimension in Figure~\ref{fig: iapo mb planning}, yielding corresponding improvements (Figure~\ref{fig: iapo mb improvement}). Further, because we are learning an iterative policy optimizer, we can zero-shot transfer a policy optimizer trained with a model-free value estimator to a model-based value estimator (Figure~\ref{fig: iapo mb transfer}). This is not possible with a direct amortized optimizer, which does not use value estimates online during policy optimization. Iterative amortization is capable of generalizing to new value estimates, \textit{instantly} incorporating updated value estimates in policy optimization. This demonstrates and highlights the opportunity for improving model-based planning through iterative amortization.

% Appendix~\ref{appendix: model-based value estimation}
\section{Discussion}
\label{sec: iapo discussion}

We have introduced iterative amortized policy optimization, a flexible and powerful policy optimization technique. In so doing, we have identified KL-regularized policy networks as a form of direct amortization, highlighting several limitations: 1) limited accuracy, as quantified by the amortization gap, 2) restriction to a single estimate, limiting exploration, and 3) inability to generalize to new objectives, limiting the transfer of these policy optimizers. As shown through our empirical analysis, iterative amortization provides a step toward improving each of these restrictions, with accompanying improvements in performance over direct amortization. Thus, iterative amortization can serve as a drop-in replacement and improvement over direct policy networks in deep RL.

This improvement, however, is accompanied by added challenges. As highlighted in Section \ref{sec: iapo mitigating value overest}, improved policy optimization can exacerbate issues in $Q$-value estimation stemming from positive bias. Note that this is not unique to iterative amortization, but applies broadly to any improved optimizer. We have provided a simple solution that involves adjusting a factor, $\beta$, to counteract this bias. Yet, we see this as an area for further investigation, perhaps drawing on insights from the offline RL community \cite{kumar2020conservative}. In addition to value estimation issues, iterative amortized policy optimization incurs computational costs that scale linearly with the number of iterations. This is in comparison with direct amortization, which has constant computational cost. Fortunately, unlike standard optimizers, iterative amortization adaptively tunes step sizes. Thus, relative improvements rapidly diminish with each additional iteration, enabling accurate optimization with exceedingly few iterations. In practice, even a single iteration per time step can work surprisingly well.

Although we have discussed three separate limitations of direct amortization, these factors are highly interconnected. By broadening policy optimization to an iterative procedure, we automatically obtain a potentially more accurate and general policy optimizer, with the capability of obtaining multiple modes. While our analysis suggests that improved exploration resulting from multiple modes is the primary factor affecting performance, future work could tease out these effects further and assess the relative contributions of these improvements in additional environments. We are hopeful that iterative amortized policy optimization, by providing a more powerful, exploratory, and general optimizer, will enable a range of improved RL algorithms.

% Although iterative amortization requires additional computation, this could be combined with some form of adaptive computation time \citep{graves2016adaptive, figurnov2018probabilistic}, gauging the required iterations. Likewise, efficiency depends, in part, on the policy initialization, which could be improved by learning the action prior, $p_\theta (\mathbf{a} | \mathbf{s})$ \citep{abdolmaleki2018maximum}. The power of iterative amortization is in using the value estimate during policy optimization to iteratively improve the policy \textit{online}. This is reminiscent of negative feedback control \citep{astrom2008feedback}, using errors to guide policy optimization, and exploring these connections further could prove useful. We are hopeful that iterative amortized policy optimization, by providing a more powerful, exploratory, and general optimizer, will enable a range of improved RL algorithms. 

\begin{ack}
JM acknowledges Scott Fujimoto for helpful discussions. This work was funded in part by NSF \#1918839 and Beyond Limits. JM is currently employed by Google DeepMind. The authors declare no other competing interests related to this work.
\end{ack}

\bibliography{main}
\bibliographystyle{plain}

%%%%%%%%%%%%%%%%%%%%%%%%%%%%%%%%%%%%%%%%%%%%%%%%%%%%%%%%%%%%
\section*{Checklist}

% %%% BEGIN INSTRUCTIONS %%%
% The checklist follows the references.  Please
% read the checklist guidelines carefully for information on how to answer these
% questions.  For each question, change the default \answerTODO{} to \answerYes{},
% \answerNo{}, or \answerNA{}.  You are strongly encouraged to include a {\bf
% justification to your answer}, either by referencing the appropriate section of
% your paper or providing a brief inline description.  For example:
% \begin{itemize}
%   \item Did you include the license to the code and datasets? \answerYes{See Section~\ref{gen_inst}.}
%   \item Did you include the license to the code and datasets? \answerNo{The code and the data are proprietary.}
%   \item Did you include the license to the code and datasets? \answerNA{}
% \end{itemize}
% Please do not modify the questions and only use the provided macros for your
% answers.  Note that the Checklist section does not count towards the page
% limit.  In your paper, please delete this instructions block and only keep the
% Checklist section heading above along with the questions/answers below.
% %%% END INSTRUCTIONS %%%

\begin{enumerate}

\item For all authors...
\begin{enumerate}
  \item Do the main claims made in the abstract and introduction accurately reflect the paper's contributions and scope?
    \answerYes{We demonstrate performance improvements and novel benefits of iterative amortization in Section~\ref{sec: iapo experiments}. Each claim is supported by empirical evidence.}
  \item Did you describe the limitations of your work?
    \answerYes{We discuss the additional computation requirements and the necessity of mitigating value overestimation.}
  \item Did you discuss any potential negative societal impacts of your work?
    \answerNo{We do not see any immediate societal impacts of this work beyond the general impacts of improvements in machine learning.}
  \item Have you read the ethics review guidelines and ensured that your paper conforms to them?
    \answerYes{}
\end{enumerate}

\item If you are including theoretical results...
\begin{enumerate}
  \item Did you state the full set of assumptions of all theoretical results?
    \answerNA{}
	\item Did you include complete proofs of all theoretical results?
    \answerNA{}
\end{enumerate}

\item If you ran experiments...
\begin{enumerate}
  \item Did you include the code, data, and instructions needed to reproduce the main experimental results (either in the supplemental material or as a URL)?
    \answerYes{We have included code in the supplementary material with an accompanying README file.}
  \item Did you specify all the training details (e.g., data splits, hyperparameters, how they were chosen)?
    \answerYes{All details and hyperparameters are provided in the Appendix.}
	\item Did you report error bars (e.g., with respect to the random seed after running experiments multiple times)?
    \answerYes{We report error bars on each of our main quantitative results comparing performance.}
	\item Did you include the total amount of compute and the type of resources used (e.g., type of GPUs, internal cluster, or cloud provider)?
    \answerYes{We provide these details in the supplementary material.}
\end{enumerate}

\item If you are using existing assets (e.g., code, data, models) or curating/releasing new assets...
\begin{enumerate}
  \item If your work uses existing assets, did you cite the creators?
    \answerYes{We cite the creators of the environments and software packages used in this paper.}
  \item Did you mention the license of the assets?
    \answerYes{We mention or cite the license for each asset.}
  \item Did you include any new assets either in the supplemental material or as a URL?
    \answerNA{}
  \item Did you discuss whether and how consent was obtained from people whose data you're using/curating?
    \answerNA{}
  \item Did you discuss whether the data you are using/curating contains personally identifiable information or offensive content?
    \answerNA{}
\end{enumerate}

\item If you used crowdsourcing or conducted research with human subjects...
\begin{enumerate}
  \item Did you include the full text of instructions given to participants and screenshots, if applicable?
    \answerNA{}
  \item Did you describe any potential participant risks, with links to Institutional Review Board (IRB) approvals, if applicable?
    \answerNA{}
  \item Did you include the estimated hourly wage paid to participants and the total amount spent on participant compensation?
    \answerNA{}
\end{enumerate}

\end{enumerate}

%%%%%%%%%%%%%%%%%%%%%%%%%%%%%%%%%%%%%%%%%%%%%%%%%%%%%%%%%%%%
\newpage
\appendix
\section{Experiment Details}
\label{appendix: experiment details}

Accompanying code is available at \href{https://github.com/joelouismarino/variational_rl}{\texttt{github.com/joelouismarino/variational\_rl}}. Experiments were implemented in \texttt{PyTorch}\footnote{\href{https://github.com/pytorch/pytorch/blob/master/LICENSE}{\texttt{https://github.com/pytorch/pytorch/blob/master/LICENSE}}} \cite{paszke2019pytorch}, analyzed with \texttt{NumPy}\footnote{\href{https://numpy.org/doc/stable/license.html}{\texttt{https://numpy.org/doc/stable/license.html}}} \cite{harris2020array}, and visualized with \texttt{Matplotlib}\footnote{\href{https://matplotlib.org/stable/users/license.html}{\texttt{https://matplotlib.org/stable/users/license.html}}} \cite{Hunter2007} and \texttt{seaborn}\footnote{\href{https://github.com/mwaskom/seaborn/blob/master/LICENSE}{\texttt{https://github.com/mwaskom/seaborn/blob/master/LICENSE}}} \cite{Waskom2021}. Logging and experiment management was handled through \texttt{Comet} \cite{CometML}. Experiments were performed on NVIDIA 1080Ti GPUs with Intel i7 8-core processors (@$4.20$GHz) on local machines, with each experiment requiring on the order of $2$ days to $1$ week to complete. We obtained the MuJoCo \cite{todorov2012mujoco} software library through an Academic Lab license.

\subsection{2D Plots}
\label{appendix: iapo 2d plots}

% In Figures~\ref{fig: iapo amortization} and \ref{fig: iapo multi-modal sampling}, we plot the estimated variational objective, $\mathcal{J}$, as a function of two dimensions of the policy mean, $\bm{\mu}$. To create these plots, we first perform policy optimization (direct amortization in Figure~\ref{fig: iapo amortization} and iterative amortization in Figure~\ref{fig: iapo multi-modal sampling}) estimating the policy mean and variance. While holding all other dimensions of the policy constant, we then estimate the variational objective while varying two dimensions of the mean ($1$ \& $3$ in Figure~\ref{fig: iapo amortization} and $2$ \& $6$ in Figure~\ref{fig: iapo multi-modal sampling}). Iterative amortization is additionally performed while preventing any updates to the constant dimensions. Even in this restricted setting, iterative amortization is capable of optimizing the policy.

In Figures 1 and 2, we plot the estimated variational objective, $\mathcal{J}$, as a function of two dimensions of the policy mean, $\bm{\mu}$. To create these plots, we first perform policy optimization (direct amortization in Figure 1 and iterative amortization in Figure 2), estimating the policy mean and variance. This is performed using on-policy trajectories from evaluation episodes (for a direct agent in Figure 1 and an iterative agent in Figure 2). While holding all other dimensions of the policy constant, we then estimate the variational objective while varying two dimensions of the mean ($1$ \& $3$ in Figure 1 and $2$ \& $6$ in Figure 2). Iterative amortization is additionally performed while preventing any updates to the constant dimensions. Even in this restricted setting, iterative amortization is capable of optimizing the policy. Additional 2D plots comparing direct vs.~iterative amortization on other environments are shown in Figure~\ref{fig: iapo additional 2d plots}, where we see similar trends.

\subsection{Value Bias Estimation}
\label{appendix: value bias est}

% We estimate the bias in the $Q$-value estimator using a similar procedure as \cite{fujimoto2018addressing}, comparing the estimate of the $Q$-networks ($\widehat{Q}_\pi$) with a Monte Carlo estimate of the future objective in the actual environment, $Q_\pi$, using a set of state-action pairs. To enable comparison across setups, we collect $100$ state-action pairs using a uniform random policy, then evaluate the estimator's bias, $\mathbb{E}_{\mathbf{s}, \mathbf{a}} \left[ \widehat{Q}_\pi - Q_\pi \right]$, throughout training. To obtain the Monte Carlo estimate of $Q_\pi$, we use $100$ action samples, which are propagated through all future time steps. The result is discounted using the same discounting factor as used during training, $\gamma = 0.99$, as well as the same Lagrange multiplier, $\alpha$. Figure~\ref{fig: value bias} shows the mean and $\pm$ standard deviation across the $100$ state-action pairs.

We estimate the bias in the $Q$-value estimator using a similar procedure as \cite{fujimoto2018addressing}, comparing the estimate of the $Q$-networks ($\widehat{Q}_\pi$) with a Monte Carlo estimate of the future objective in the actual environment, $Q_\pi$, using a set of state-action pairs. To enable comparison across setups, we collect $100$ state-action pairs using a uniform random policy, then evaluate the estimator's bias, $\mathbb{E}_{\mathbf{s}, \mathbf{a}} \left[ \widehat{Q}_\pi - Q_\pi \right]$, throughout training. To obtain the Monte Carlo estimate of $Q_\pi$, we use $100$ action samples, which are propagated through all future time steps. The result is discounted using the same discounting factor as used during training, $\gamma = 0.99$, as well as the same Lagrange multiplier, $\alpha$. Figure 3 shows the mean and $\pm$ standard deviation across the $100$ state-action pairs.

\subsection{Amortization Gap Estimation}
\label{appendix: am gap calc}

Calculating the amortization gap in the RL setting is challenging, as properly evaluating the variational objective, $\mathcal{J}$, involves unrolling the environment. During training, the objective is estimated using a set of $Q$-networks and/or a learned model. However, finding the optimal policy distribution, $\widehat{\pi}$, under these learned value estimates may not accurately reflect the amortization gap, as the value estimator likely contains positive bias (Figure 3). Because the value estimator is typically locally accurate near the current policy, we estimate the amortization gap by performing gradient ascent on $\mathcal{J}$ w.r.t.~the policy distribution parameters, $\bm{\lambda}$, initializing from the amortized estimate (from $\pi_\phi$). This is a form \textit{semi-amortized} variational inference \citep{hjelm2016iterative, krishnan2017challenges, kim2018semi}. We use the Adam optimizer \citep{kingma2014adam} with a learning rate of $5 \times 10^{-3}$ for $100$ gradient steps, which we found consistently converged. This results in the estimated optimized $\widehat{\pi}$. We estimate the gap using $100$ on-policy states, calculating $\mathcal{J} (\theta, \widehat{\pi}) - \mathcal{J} (\theta, \pi)$, i.e.~the improvement in the objective after gradient-based optimization. Figure 6 shows the resulting mean and $\pm$ standard deviation. We also run iterative amortized policy optimization for an additional $5$ iterations during this evaluation, empirically yielding an additional decrease in the estimated amortization gap.

\subsection{Hyperparameters}
\label{appendix: iapo hyperparameters}

Our setup follows that of soft actor-critic (SAC) \citep{haarnoja2018soft, haarnoja2018soft2}, using a uniform action prior, i.e.~entropy regularization, and two $Q$-networks \citep{fujimoto2018addressing}. Off-policy training is performed using a replay buffer \citep{lin1992self, mnih2013playing}. Training hyperparameters are given in Table~\ref{table: iapo training hyperparameters}. 

\paragraph{Temperature} Following \cite{haarnoja2018soft2}, we adjust the temperature, $\alpha$, to maintain a specified entropy constraint, $\epsilon_\alpha = |\mathcal{A}|$, where $|\mathcal{A}|$ is the size of the action space, i.e.~the dimensionality.

% \begin{table}[h!]
% \centering
% \caption{\textbf{Policy Inputs \& Outputs}.}
% \label{table:policy inputs and outputs}
% \begin{tabular}{lrr}
% \toprule
%  & Inputs & Outputs \\
% \midrule
% Direct & $\mathbf{s}$ & $\bm{\lambda}$ \\
% Iterative & $\mathbf{s}$, $\bm{\lambda}$, $\nabla_{\bm{\lambda}} \mathcal{J}$ & $\bm{\delta}$, $\bm{\omega}$ \\
% \bottomrule
% \end{tabular}
% \end{table}

% \begin{table}[h!]
% \centering
% \caption{\textbf{Policy Networks}.}
% \label{table:policy hyperparameters}
% \begin{tabular}{lr}
% \toprule
% Hyperparameter & Value \\
% \midrule
% Number of Layers & 2 \\
% Number of Units / Layer & 256 \\
% Non-linearity & \texttt{ReLU} \\
% \bottomrule
% \end{tabular}
% \end{table}

\begin{table}[h!]
\parbox{.45\linewidth}{
\centering
\caption{\textbf{Policy Inputs \& Outputs}.}
\label{table:policy inputs and outputs}
\begin{tabular}{lrr}
\toprule
 & Inputs & Outputs \\
\midrule
Direct & $\mathbf{s}$ & $\bm{\lambda}$ \\
Iterative & $\mathbf{s}$, $\bm{\lambda}$, $\nabla_{\bm{\lambda}} \mathcal{J}$ & $\bm{\delta}$, $\bm{\omega}$ \\
\bottomrule
\end{tabular}
}
\hfill
\parbox{.45\linewidth}{
\centering
\caption{\textbf{Policy Networks}.}
\label{table:policy hyperparameters}
\begin{tabular}{lr}
\toprule
Hyperparameter & Value \\
\midrule
Number of Layers & 2 \\
Number of Units / Layer & 256 \\
Non-linearity & \texttt{ReLU} \\
\bottomrule
\end{tabular}
}
\end{table}

% \paragraph{Policy} We use the same network architecture (number of layers, units/layer, non-linearity) for both direct and iterative amortized policy optimizers (Table~\ref{table:policy hyperparameters}). Each policy network results in Gaussian distribution parameters, and we apply a $\texttt{tanh}$ transform to ensure $\mathbf{a} \in [-1, 1]$ \citep{haarnoja2018soft}. In the case of a Gaussian, the distribution parameters are $\bm{\lambda} = \left[ \bm{\mu}, \bm{\sigma} \right]$. The inputs and outputs of each optimizer form are given in Table~\ref{table:policy inputs and outputs}. Again, $\bm{\delta}$ and $\bm{\omega}$ are respectively the update and gate of the iterative amortized optimizer (Eq.~\ref{eq: iapo gated update}), each of which are defined for both $\bm{\mu}$ and $\bm{\sigma}$. Following \cite{marino2018iterative}, we apply layer normalization \citep{ba2016layer} individually to each of the inputs to iterative amortized optimizers.

\begin{figure*}[t!]
    \centering
    \begin{subfigure}[t]{0.48\textwidth}
        \centering
        \includegraphics[width=\textwidth]{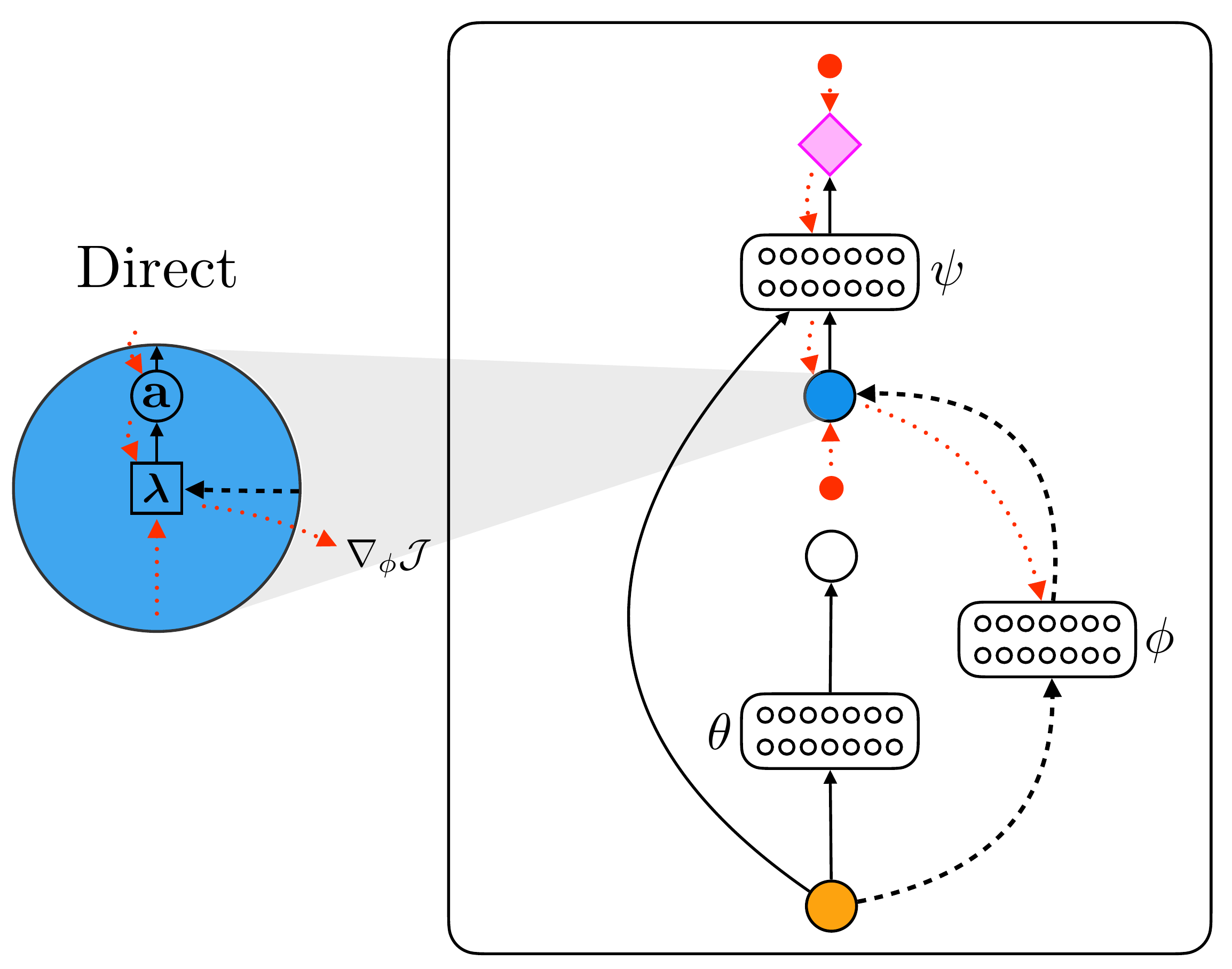}
        \caption{Direct Amortization}
    \end{subfigure}%
    ~ 
    \begin{subfigure}[t]{0.48\textwidth}
        \centering
        \includegraphics[width=\textwidth]{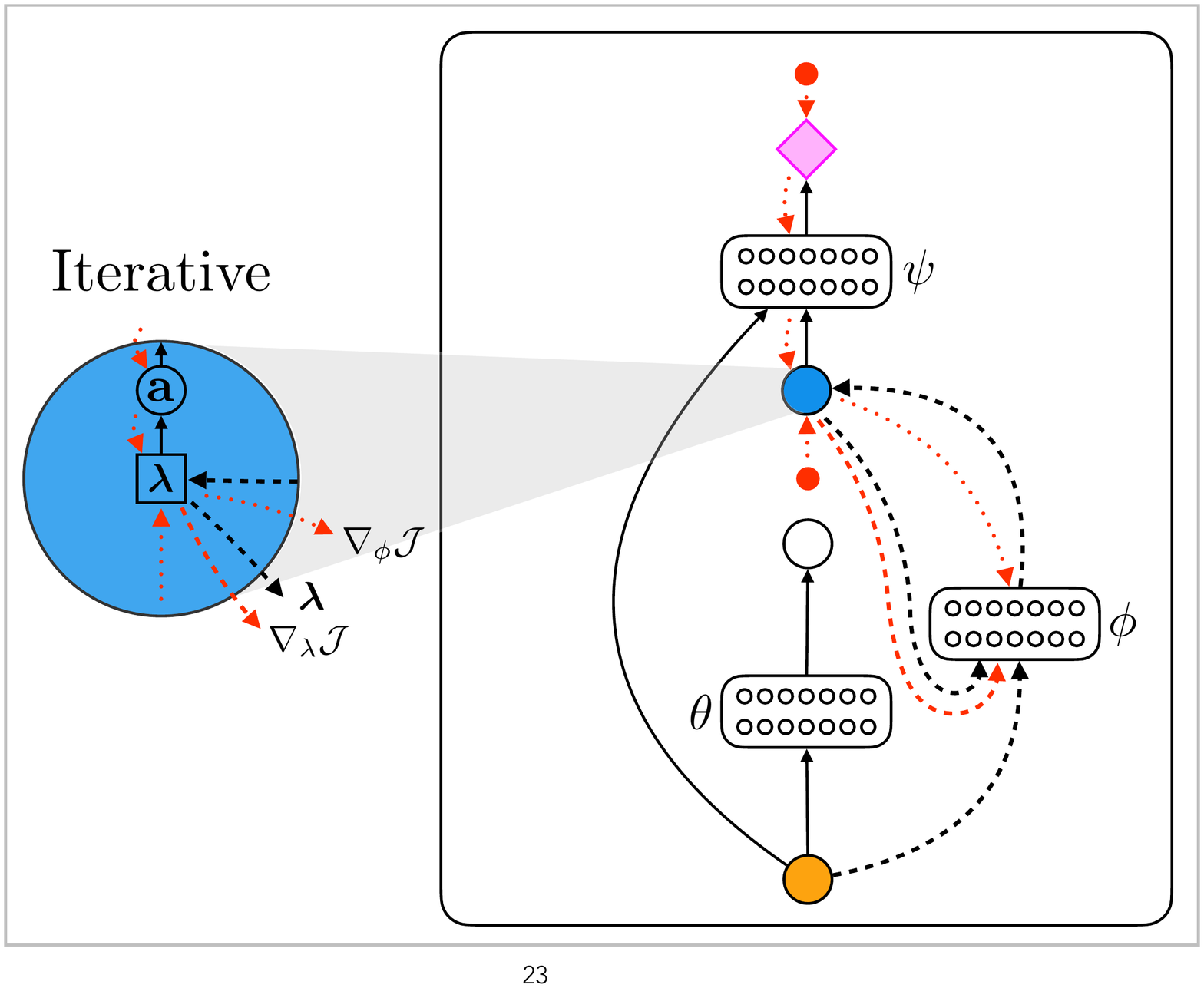}
        \caption{Iterative Amortization}
    \end{subfigure}
    \caption{\textbf{Amortized Optimizers}. Diagrams of \textbf{(a)} direct and \textbf{(b)} iterative amortized policy optimization. As in Figure 1, larger circles represent probability distributions, and smaller red circles represent terms in the objective. Red dotted arrows represent gradients. In addition to the state, $\mathbf{s}_t$, iterative amortization uses the current policy distribution estimate, $\bm{\lambda}$, and the policy optimization gradient, $\nabla_{\bm{\lambda}} \mathcal{J}$, to iteratively optimize $\mathcal{J}$. Like direct amortization, the optimizer network parameters, $\phi$, are updated using $\nabla_\phi \mathcal{J}$. This generally requires some form of stochastic gradient estimation to differentiate through $\mathbf{a}_t \sim \pi (\mathbf{a}_t | \mathbf{s}_t, \mathcal{O}; \bm{\lambda})$.}
    \label{fig: iapo amortized opt diagrams}
\end{figure*}

\paragraph{Policy} We use the same network architecture (number of layers, units/layer, non-linearity) for both direct and iterative amortized policy optimizers (Table~\ref{table:policy hyperparameters}). Each policy network results in Gaussian distribution parameters, and we apply a $\texttt{tanh}$ transform to ensure $\mathbf{a} \in [-1, 1]$ \citep{haarnoja2018soft}. In the case of a Gaussian, the distribution parameters are $\bm{\lambda} = \left[ \bm{\mu}, \bm{\sigma} \right]$. The inputs and outputs of each optimizer form are given in Table~\ref{table:policy inputs and outputs}. Again, $\bm{\delta}$ and $\bm{\omega}$ are respectively the update and gate of the iterative amortized optimizer (Eq.~12 in the main text), each of which are defined for both $\bm{\mu}$ and $\bm{\sigma}$. Following \cite{marino2018iterative}, we apply layer normalization \citep{ba2016layer} individually to each of the inputs to iterative amortized optimizers. We initialize iterative amortization with $\bm{\mu} = \mathbf{0}$ and $\bm{\sigma} = \mathbf{1}$, however, these could be initialized from a learned action prior \citep{marino2018general}.

% \begin{table}[h!]
% \centering
% \caption{\textbf{$Q$-value Network Architecture A}.}
% \label{table:iapo q value arch a}
% \begin{tabular}{lr}
% \toprule
% Hyperparameter & Value \\
% \midrule
% Number of Layers & 2 \\
% Number of Units / Layer & 256 \\
% Non-linearity & \texttt{ReLU} \\
% Layer Normalization & False \\
% Connectivity & Sequential \\
% \bottomrule
% \end{tabular}
% \end{table}

% \begin{table}[h!]
% \centering
% \caption{\textbf{$Q$-value Network Architecture B}.}
% \label{table:iapo q value arch b}
% \begin{tabular}{lr}
% \toprule
% Hyperparameter & Value \\
% \midrule
% Number of Layers & 3 \\
% Number of Units / Layer & 512 \\
% Non-linearity & \texttt{ELU} \\
% Layer Normalization & True \\
% Connectivity & Highway \\
% \bottomrule
% \end{tabular}
% \end{table}

\begin{table}[h!]
\parbox{.45\linewidth}{
\centering
\caption{\textbf{$Q$-value Network Architecture A}.}
\label{table:iapo q value arch a}
\begin{tabular}{lr}
\toprule
Hyperparameter & Value \\
\midrule
Number of Layers & 2 \\
Number of Units / Layer & 256 \\
Non-linearity & \texttt{ReLU} \\
Layer Normalization & False \\
Connectivity & Sequential \\
\bottomrule
\end{tabular}
}
\hfill
\parbox{.45\linewidth}{
\centering
\caption{\textbf{$Q$-value Network Architecture B}.}
\label{table:iapo q value arch b}
\begin{tabular}{lr}
\toprule
Hyperparameter & Value \\
\midrule
Number of Layers & 3 \\
Number of Units / Layer & 512 \\
Non-linearity & \texttt{ELU} \\
Layer Normalization & True \\
Connectivity & Highway \\
\bottomrule
\end{tabular}
}
\end{table}

% \paragraph{$Q$-value} We investigated two $Q$-value network architectures. Architecture A (Table~\ref{table:iapo q value arch a}) is the same as that from \cite{haarnoja2018soft}. Architecture B (Table~\ref{table:iapo q value arch b}) is a wider, deeper network with highway connectivity \citep{srivastava2015training}, layer normalization \citep{ba2016layer}, and ELU nonlinearities \citep{clevert2015fast}. We initially compared each $Q$-value network architecture using each policy optimizer on each environment. The results in Figure~\ref{fig: model-free performance} were obtained using the better performing architecture in each case, given in Table~\ref{table:iapo q net per env}. As in \cite{fujimoto2018addressing}, we use an ensemble of $2$ separate $Q$-networks in each experiment.

\begin{figure*}[t!]
    \centering
    \begin{subfigure}[t]{0.24\textwidth}
        \centering
        \includegraphics[width=\textwidth]{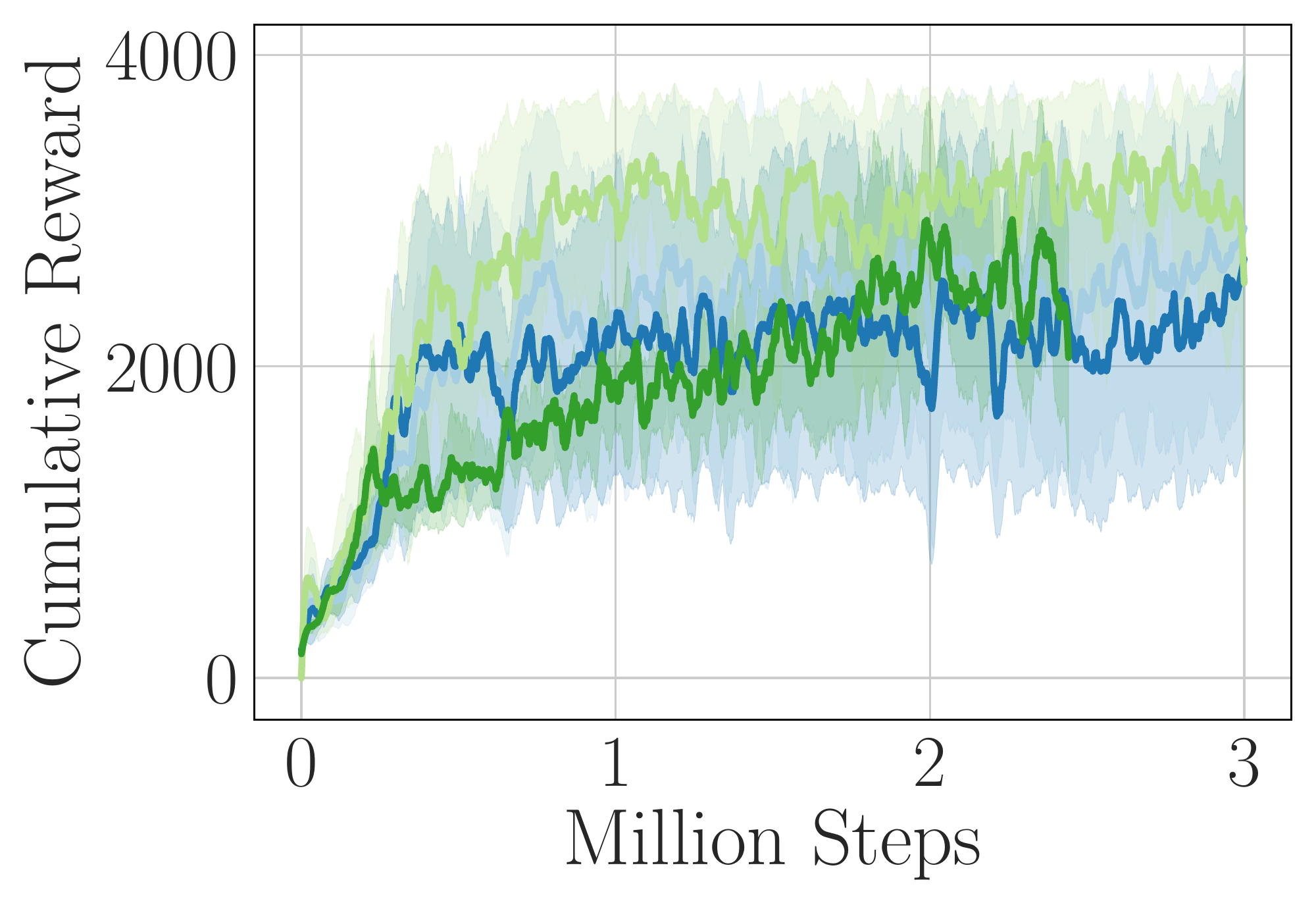}
        \caption{\texttt{Hopper-v2}}
    \end{subfigure}%
    ~ 
    \begin{subfigure}[t]{0.24\textwidth}
        \centering
        \includegraphics[width=\textwidth]{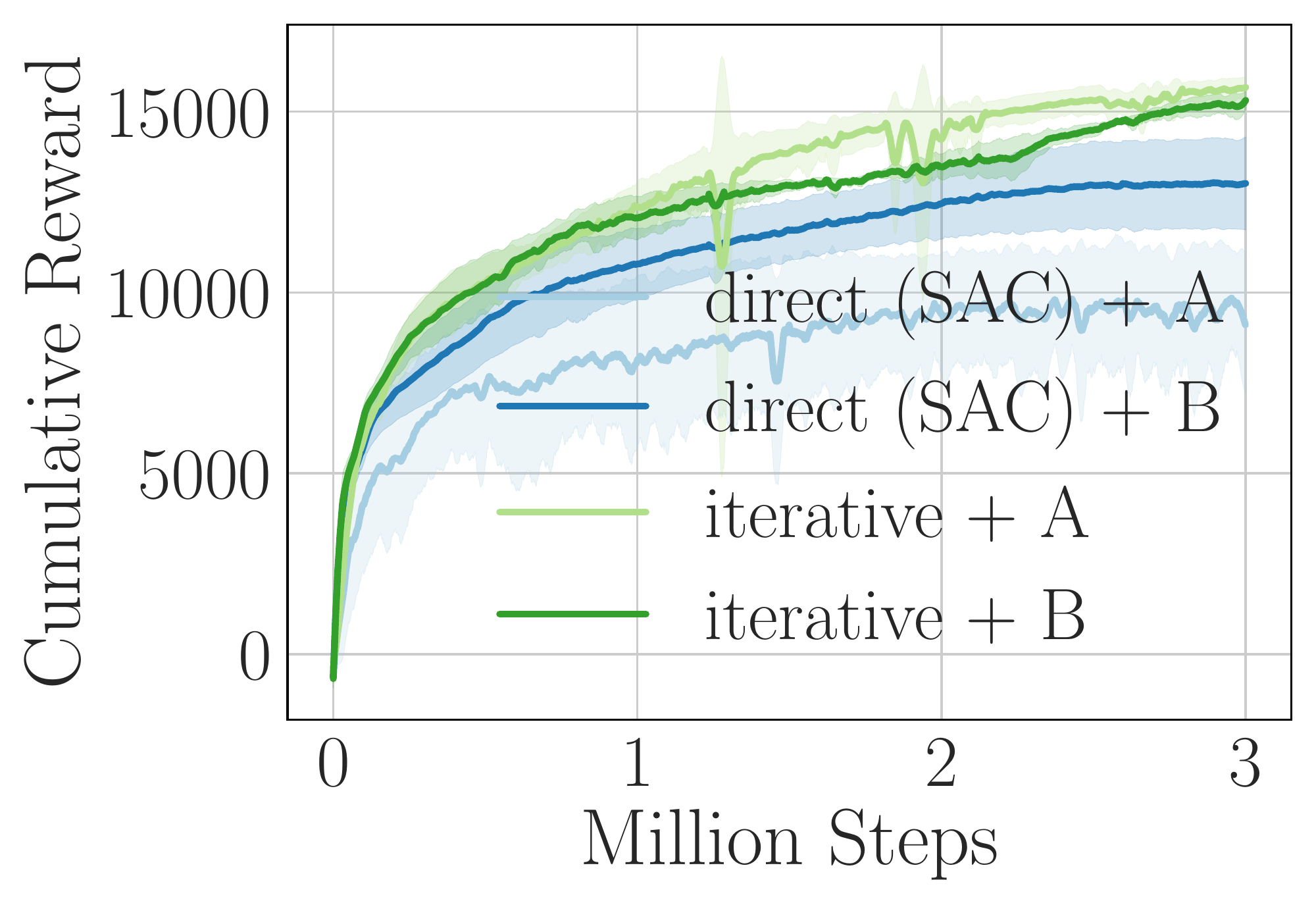}
        \caption{\texttt{HalfCheetah-v2}}
    \end{subfigure}%
    ~ 
    \begin{subfigure}[t]{0.24\textwidth}
        \centering
        \includegraphics[width=\textwidth]{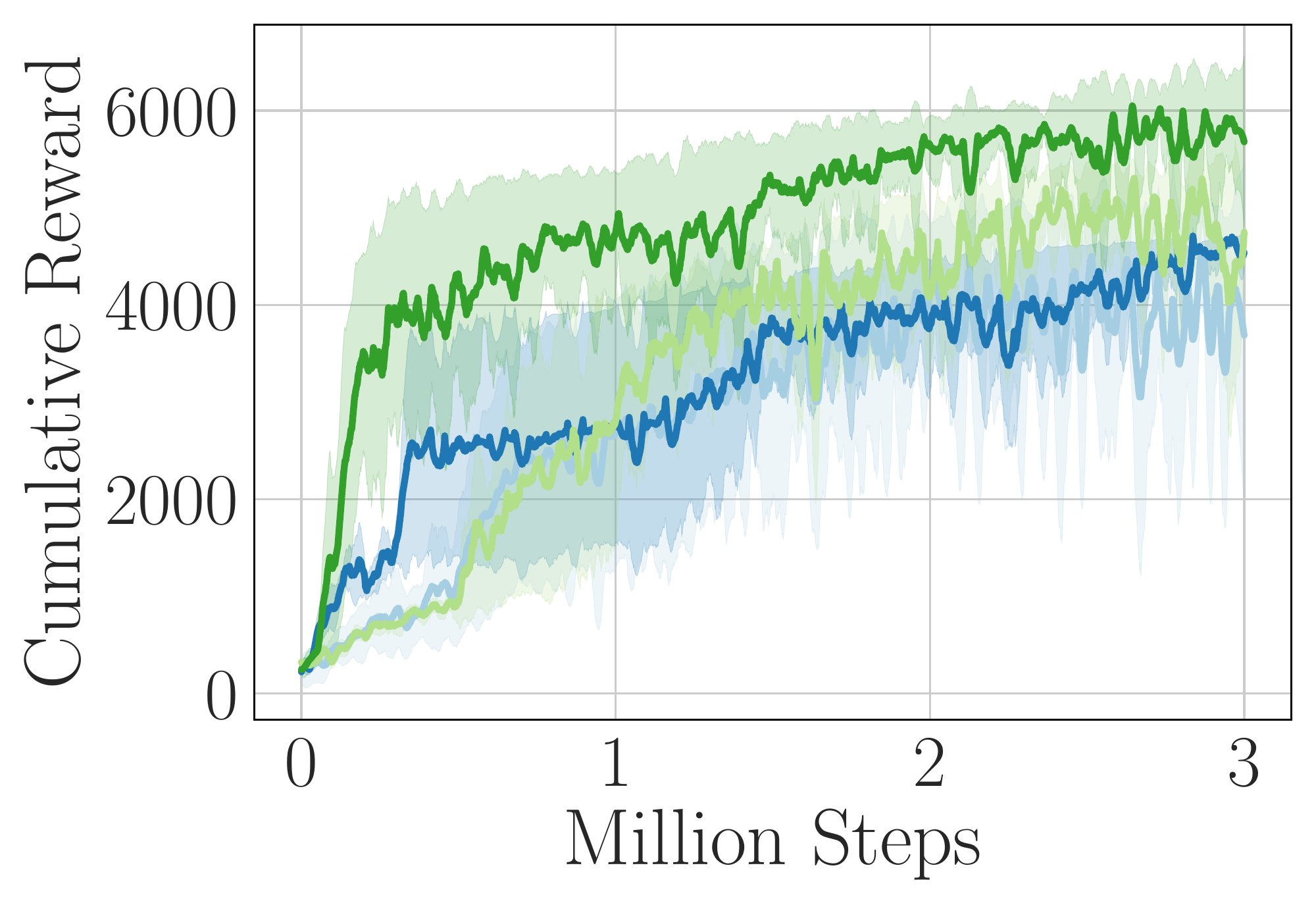}
        \caption{\texttt{Walker2d-v2}}
    \end{subfigure}%
    ~ 
    \begin{subfigure}[t]{0.24\textwidth}
        \centering
        \includegraphics[width=\textwidth]{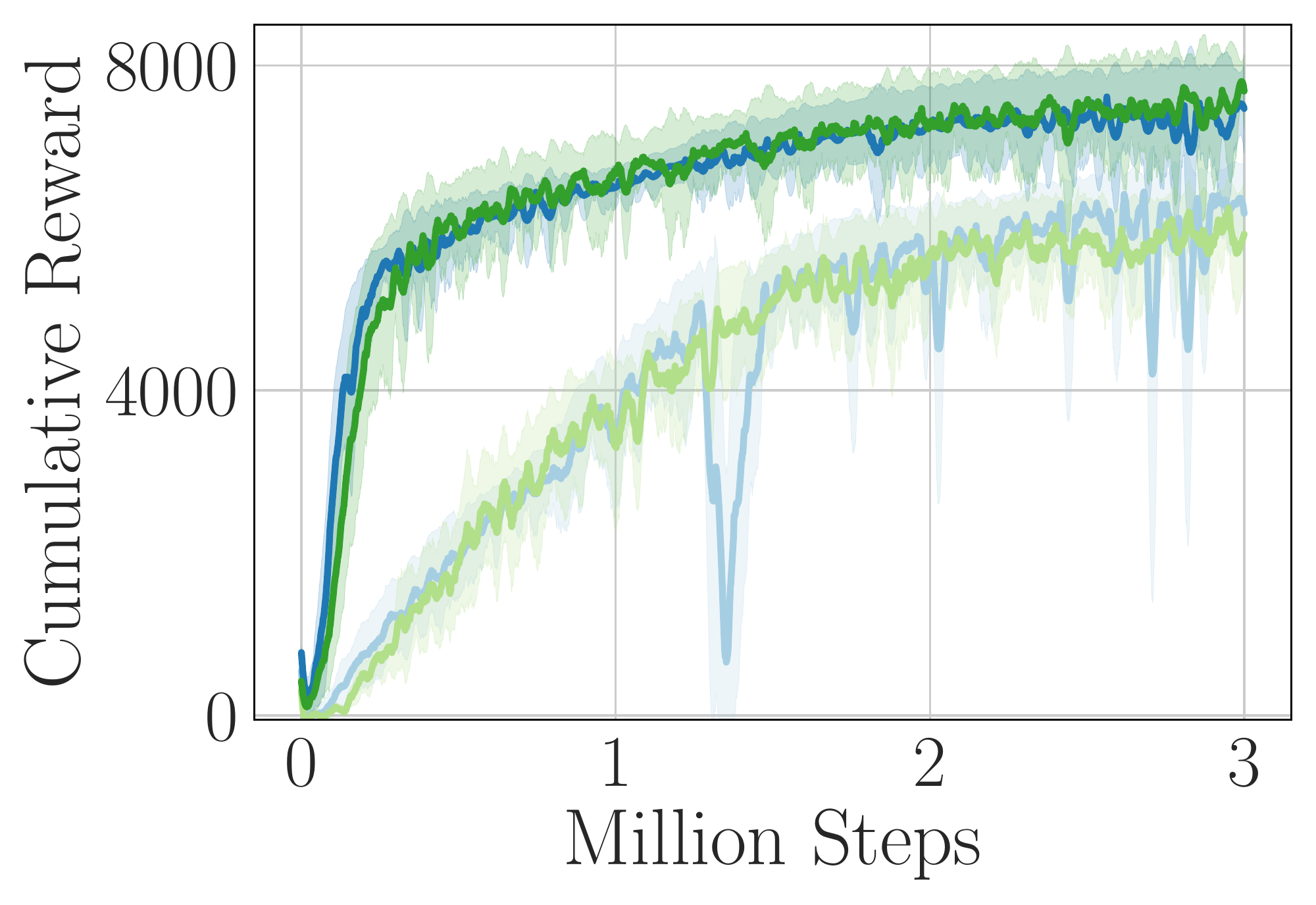}
        \caption{\texttt{Ant-v2}}
    \end{subfigure}
    \caption{\textbf{Value Architecture Comparison}. Plots show performance for $\geq 3$ seeds for each value architecture (A or B) for each policy optimization technique (direct or iterative). Note: results for iterative $+$ B on \texttt{Hopper-v2} were obtained with an overly pessimistic value estimate ($\beta = 2.5$ rather than $\beta=1.5$) and are consequently worse.}
    \label{fig: iapo value arch plots}
\end{figure*}

\paragraph{$Q$-value} We investigated two $Q$-value network architectures. Architecture A (Table~\ref{table:iapo q value arch a}) is the same as that from \cite{haarnoja2018soft}. Architecture B (Table~\ref{table:iapo q value arch b}) is a wider, deeper network with highway connectivity \citep{srivastava2015training}, layer normalization \citep{ba2016layer}, and ELU nonlinearities \citep{clevert2015fast}. We initially compared each $Q$-value network architecture using each policy optimizer on each environment, as shown in Figure~\ref{fig: iapo value arch plots}. The results in Figure 5 were obtained using the better performing architecture in each case, given in Tables~\ref{table:iapo q net per env} \& \ref{table:iapo q net per env2}. As in \cite{fujimoto2018addressing}, we use an ensemble of $2$ separate $Q$-networks in each experiment.

\begin{table*}[h]
\centering
\caption{\textbf{$Q$-value Network Architecture by Environment}.}
\label{table:iapo q net per env}
\begin{tabular}{lrrrrrr}
\toprule
 & \tiny \texttt{InvertedPendulum-v2} & \tiny \texttt{InvertedDoublePendulum-v2} & \tiny \texttt{Hopper-v2} & \tiny \texttt{HalfCheetah-v2} & \tiny \texttt{Walker2d-v2} & \tiny \texttt{Ant-v2} \\
\midrule
Direct & A & A & A & B & A & B \\
Iterative & A & A & A & A & B & B \\
\bottomrule
\end{tabular}
\end{table*}

\begin{table*}[h]
\centering
\caption{\textbf{$Q$-value Network Architecture by Environment (Continued)}.}
\label{table:iapo q net per env2}
\begin{tabular}{lrrrr}
\toprule
 & \tiny \texttt{Reacher-v2} & \tiny \texttt{Swimmer-v2} & \tiny \texttt{Humanoid-v2} & \tiny \texttt{HumanoidStandup-v2} \\
\midrule
Direct & A & A & B & B \\
Iterative & A & A & B & B \\
\bottomrule
\end{tabular}
\end{table*}

% \paragraph{Value Pessimism ($\beta$)} As discussed in Section~\ref{sec: iapo mitigating value overest}, the increased flexibility of iterative amortization allows it to potentially exploit inaccurate value estimates. We increased the pessimism hyperparameter, $\beta$, to further penalize variance in the value estimate. Experiments with direct amortization use the default $\beta = 1$ in all environments, as we did not find that increasing $\beta$ helped in this setup. For iterative amortization, we use $\beta = 1.5$ on \texttt{Hopper-v2} and $\beta=2.5$ on all other environments. This is only applied during training; while collecting data in the environment, we use $\beta = 1$ to not overly penalize exploration.

\paragraph{Value Pessimism ($\beta$)} As discussed in Section 3.2.2, the increased flexibility of iterative amortization allows it to potentially exploit inaccurate value estimates. We increased the pessimism hyperparameter, $\beta$, to further penalize variance in the value estimate. Experiments with direct amortization use the default $\beta = 1$ in all environments, as we did not find that increasing $\beta$ helped in this setup. For iterative amortization, we use $\beta = 1.5$ on \texttt{Hopper-v2} and $\beta=2.5$ on all other environments. This is only applied during training; while collecting data in the environment, we use $\beta = 1$ to not overly penalize exploration.

\begin{table}[t]
\centering
\caption{\textbf{Training Hyperparameters}.}
\label{table: iapo training hyperparameters}
\begin{tabular}{lr}
\toprule
Hyperparameter & Value \\
\midrule
Discount Factor $(\gamma)$ & $0.99$ \\
$Q$-network Update Rate $(\tau)$ & $5 \cdot 10^{-3}$ \\
Network Optimizer & Adam \\
Learning Rate & $3 \cdot 10^{-4}$ \\
Batch Size & $256$ \\
Initial Random Steps & $5 \cdot 10^3$ \\
Replay Buffer Size & $10^6$ \\
\bottomrule
\end{tabular}
\end{table}

\subsection{Model-Based Value Estimation}
\label{appendix: model-based value estimation}

For model-based experiments, we use a single, deterministic model together with the ensemble of $2$ $Q$-value networks (discussed above).

\paragraph{Model} We use separate networks to estimate the state transition dynamics, $p_\textrm{env} (\mathbf{s}_{t+1} | \mathbf{s}_t , \mathbf{a}_t)$, and reward function, $r (\mathbf{s}_t , \mathbf{a}_t)$. The network architecture is given in Table~\ref{table:iapo dynamics and reward arch}. Each network outputs the mean of a Gaussian distribution; the standard deviation is a separate, learnable parameter. The reward network directly outputs the mean estimate, whereas the state transition network outputs a residual estimate, $\Delta_{\mathbf{s}_t}$, yielding an updated mean estimate through:
\begin{equation}
    \bm{\mu}_{\mathbf{s}_{t+1}} = \mathbf{s}_t + \Delta_{\mathbf{s}_t}. \nonumber
\end{equation}

% \begin{table}[t!]
% \centering
% \caption{\textbf{Model Network Architectures}.}
% \label{table:iapo dynamics and reward arch}
% \begin{tabular}{lr}
% \toprule
% Hyperparameter & Value \\
% \midrule
% Number of Layers & 2 \\
% Number of Units / Layer & 256 \\
% Non-linearity & \texttt{Leaky ReLU} \\
% Layer Normalization & True \\
% \bottomrule
% \end{tabular}
% \end{table}

% \begin{table}[t!]
% \centering
% \caption{\textbf{Model-Based Hyperparameters}.}
% \label{table:iapo mb value est hyperparams}
% \begin{tabular}{lr}
% \toprule
% Hyperparameter & Value \\
% \midrule
% Rollout Horizon, $h$ & 2 \\
% Retrace $\lambda$ & 0.9 \\
% Pre-train Model Updates & $10^3$ \\
% Model-Based Value Targets & True \\
% \bottomrule
% \end{tabular}
% \end{table}

\begin{table}[h!]
\parbox{.45\linewidth}{
\centering
\caption{\textbf{Model Network Architectures}.}
\label{table:iapo dynamics and reward arch}
\begin{tabular}{lr}
\toprule
Hyperparameter & Value \\
\midrule
Number of Layers & 2 \\
Number of Units / Layer & 256 \\
Non-linearity & \texttt{Leaky ReLU} \\
Layer Normalization & True \\
\bottomrule
\end{tabular}
}
\hfill
\parbox{.45\linewidth}{
\centering
\caption{\textbf{Model-Based Hyperparameters}.}
\label{table:iapo mb value est hyperparams}
\begin{tabular}{lr}
\toprule
Hyperparameter & Value \\
\midrule
Rollout Horizon, $h$ & 2 \\
Retrace $\lambda$ & 0.9 \\
Pre-train Model Updates & $10^3$ \\
Model-Based Value Targets & True \\
\bottomrule
\end{tabular}
}
\end{table}

\begin{figure}[t!]
    \centering
    \includegraphics[width=0.48\textwidth]{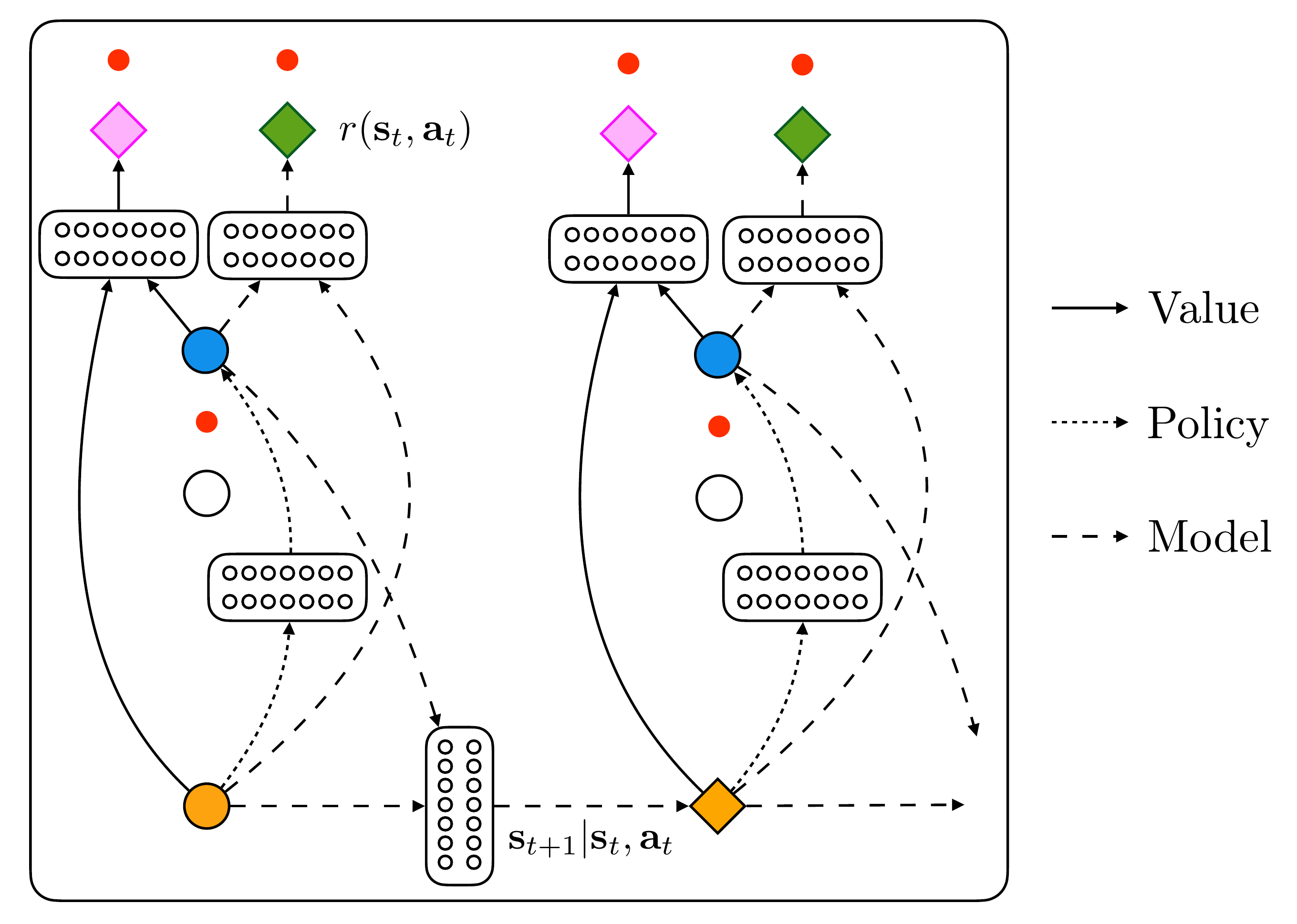}
    \caption{\textbf{Model-Based Value Estimation}. Diagram of model-based value estimation (shown with direct amortization). For clarity, the diagram is shown without the policy prior network, $p_\theta (\mathbf{a}_t | \mathbf{s}_t)$. The model consists of a deterministic reward estimate, $r (\mathbf{s}_t, \mathbf{a}_t)$, (green diamond) and a state estimate, $\mathbf{s}_{t+1} | \mathbf{s}_t, \mathbf{a}_t$, (orange diamond). The model is unrolled over a horizon, $H$, and the $Q$-value is estimated using the Retrace estimator \citep{munos2016safe}, given in Eq.~\ref{eq: iapo retrace}.}
    \label{fig: iapo model diagram}
\end{figure}

\paragraph{Model Training} The state transition and reward networks are both trained using maximum log-likelihood training, using data examples from the replay buffer. Training is performed at the same frequency as policy and $Q$-network training, using the same batch size ($256$) and network optimizer. However, we perform $10^3$ updates at the beginning of training, using the initial random steps, in order to start with a reasonable model estimate.

\paragraph{Value Estimation} To estimate $Q$-values, we combine short model rollouts with the model-free estimates from the $Q$-networks. Specifically, we unroll the model and policy, obtaining state, reward, and policy estimates at current and future time steps. We then apply the $Q$-value networks to these future state-action estimates. Future rewards and value estimates are combined using the Retrace estimator \citep{munos2016safe}. Denoting the estimate from the $Q$-network as $\widehat{Q}_\psi (\mathbf{s} , \mathbf{a})$ and the reward estimate as $\widehat{r} (\mathbf{s}, \mathbf{a})$, we calculate the $Q$-value estimate at the current time step as
\begin{equation}
    \widehat{Q}_\pi (\mathbf{s}_t , \mathbf{a}_t) = \widehat{Q}_\psi (\mathbf{s}_t , \mathbf{a}_t) + \mathbb{E} \left[ \sum_{t^\prime = t}^{t+h} \gamma^{t^\prime - t} \lambda^{t^\prime - t} \delta_{t^\prime} \right],
    \label{eq: iapo retrace}
\end{equation}
where $\delta_{t^\prime}$ is the estimated temporal difference:
\begin{equation}
    \delta_{t^\prime} \equiv \widehat{r} (\mathbf{s}_{t^\prime}, \mathbf{a}_{t^\prime}) + \gamma \widehat{V}_\psi (\mathbf{s}_{t^\prime + 1})  - \widehat{Q}_\psi (\mathbf{s}_{t^\prime} , \mathbf{a}_{t^\prime}),
\end{equation}
$\lambda$ is an exponential weighting factor, $h$ is the rollout horizon, and the expectation is evaluated under the model and policy. In the variational RL setting, the state-value, $V_\pi (\mathbf{s})$, is
\begin{equation}
    V_\pi (\mathbf{s}) = \mathbb{E}_\pi \left[ Q_\pi (\mathbf{s}, \mathbf{a}) - \alpha \log \frac{\pi (\mathbf{a} | \mathbf{s}, \mathcal{O})}{p_\theta (\mathbf{a} | \mathbf{s})} \right].
    \label{eq: iapo state value def}
\end{equation}
In Eq.~\ref{eq: iapo retrace}, we approximate $V_\pi$ using the $Q$-network to approximate $Q_\pi$ in Eq.~\ref{eq: iapo state value def}, yielding $\widehat{V}_\psi (\mathbf{s})$. Finally, to ensure consistency between the model and the $Q$-value networks, we use the model-based estimate from Eq.~\ref{eq: iapo retrace} to provide target values for the $Q$-networks, as in \cite{janner2019trust}.

\paragraph{Future Policy Estimates} Evaluating the expectation in Eq.~\ref{eq: iapo retrace} requires estimates of $\pi$ at future time steps. This is straightforward with direct amortization, which employs a feedforward policy, however, with iterative amortization, this entails recursively applying an iterative optimization procedure. Alternatively, we could use the prior, $p_\theta (\mathbf{a} | \mathbf{s})$, at future time steps, but this does not apply in the max-entropy setting, where the prior is uniform. For computational efficiency, we instead learn a separate direct (amortized) policy for model-based rollouts. That is, with iterative amortization, we create a separate direct network using the same hyperparameters from Table~\ref{table:policy hyperparameters}. This network distills iterative amortization into a direct amortized optimizer, through the KL divergence, $D_\textrm{KL} (\pi_\textrm{it.} || \pi_\textrm{dir.})$. Rollout policy networks are common in model-based RL \citep{silver2016mastering, piche2019probabilistic}.

\begin{figure}[t!]
    \centering
    \begin{subfigure}[t]{0.24\textwidth}
    \centering
    \includegraphics[width=\textwidth]{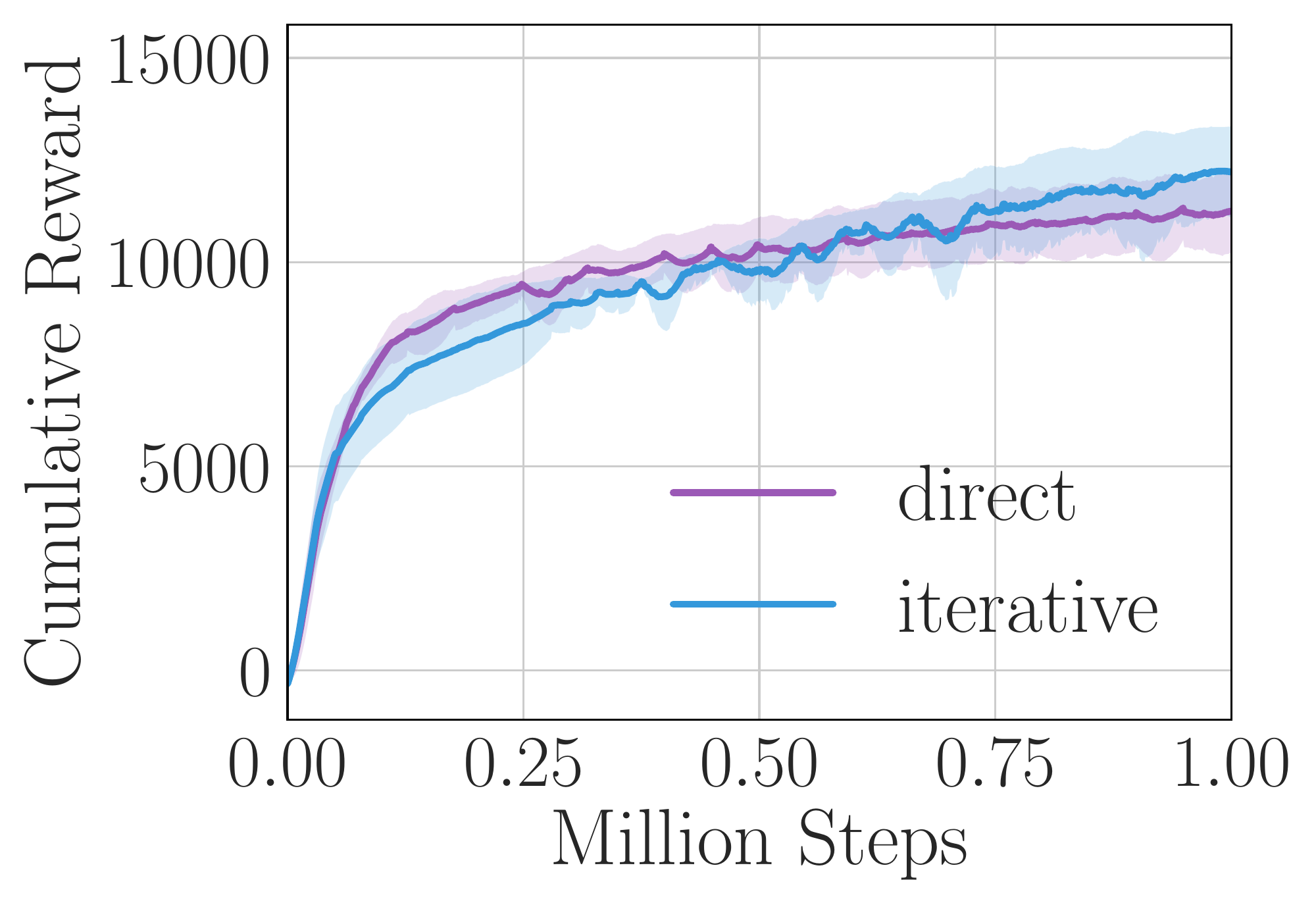}
    \caption{Performance}
    \label{fig: iapo mb perf}
    \end{subfigure}%
    ~
    \begin{subfigure}[t]{0.24\textwidth}
    \centering
    \includegraphics[width=\textwidth]{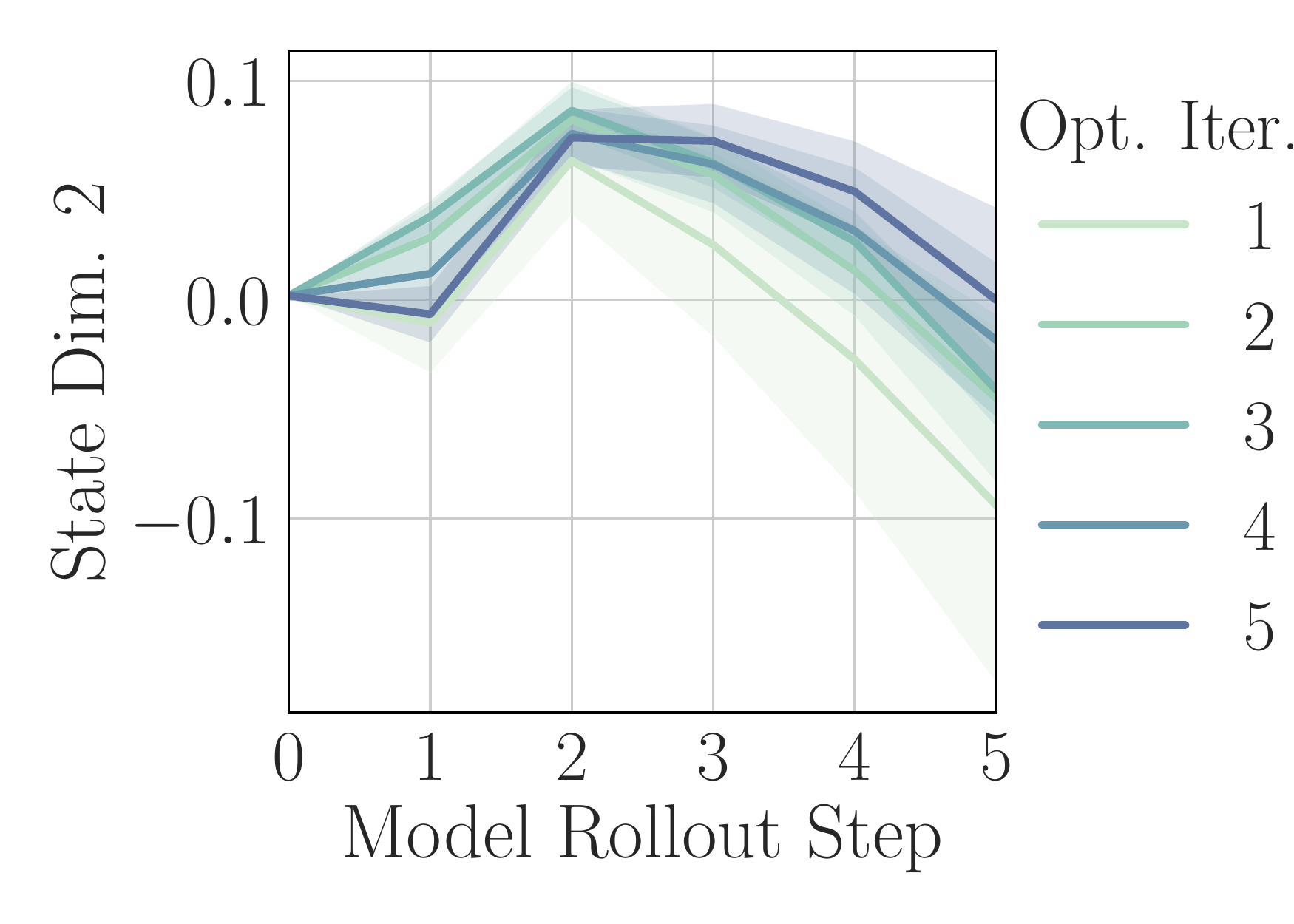}
    \caption{Planning}
    \label{fig: iapo mb planning}
    \end{subfigure}%
    ~
    \begin{subfigure}[t]{0.17\textwidth}
    \centering
    \includegraphics[width=\textwidth]{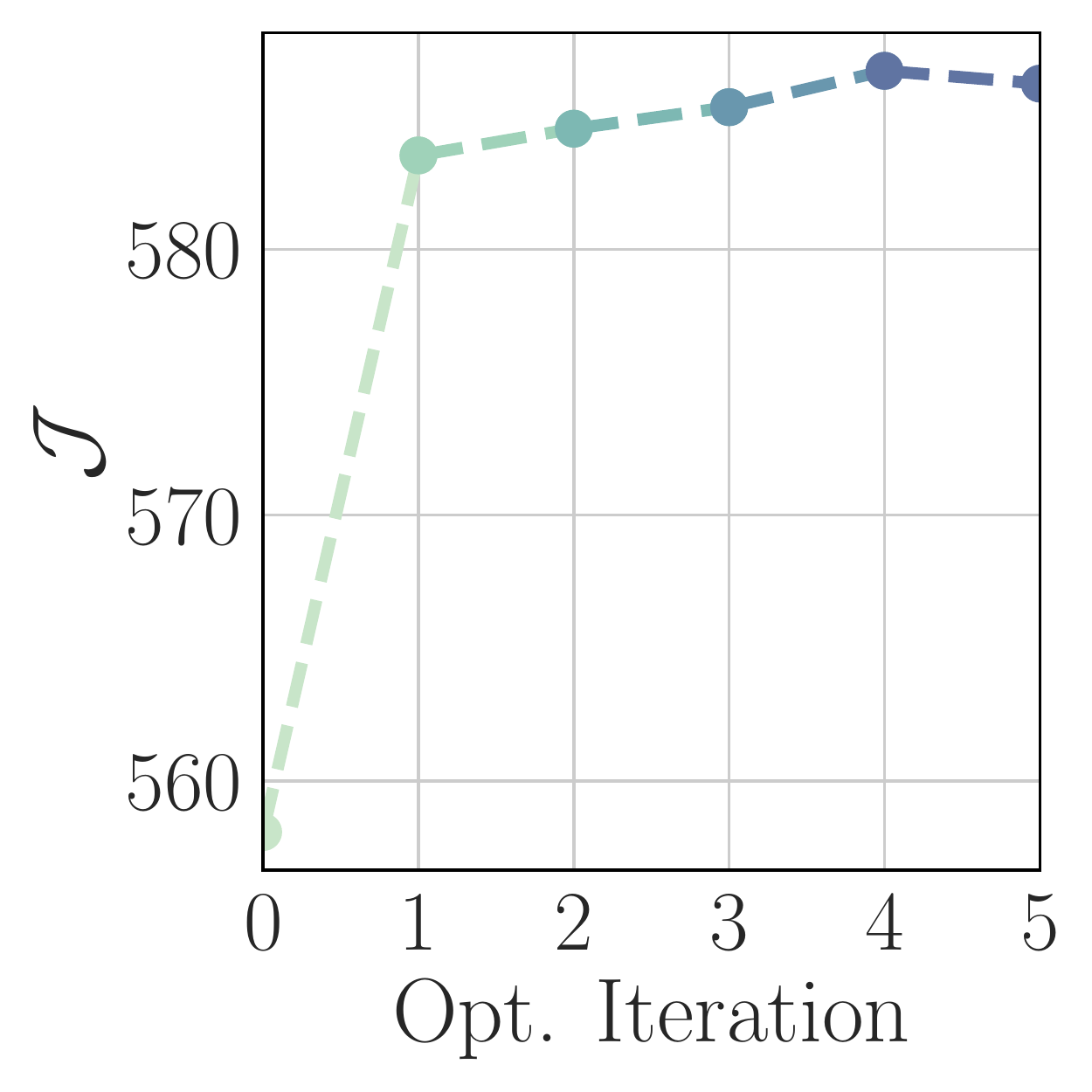}
    \caption{Improvement}
    \label{fig: iapo mb improvement}
    \end{subfigure}
    \caption{\textbf{Model-Based Value Estimates}. \textbf{(a)} Performance comparison of direct and iterative amortization using model-based value estimates. Curves show the mean and $\pm$ std.~dev.~over $4$ random seeds. \textbf{(b)} Planned trajectories over policy optimization iterations. \textbf{(c)} The corresponding estimated objective increases over iterations.}
    \label{fig: iapo app mb value estimation}
\end{figure}

\paragraph{Model-Based Results} In Figure \ref{fig: iapo app mb value estimation}, we show results using iterative amortization with model-based value estimates. Iterative amortization offers a slight improvement over direct amortization in terms of performance at $1$ million steps. As in the model-free case, iterative amortization results in improvement over iterations (Fig.~\ref{fig: iapo mb improvement}), but now as a result of planning trajectories (Fig.~\ref{fig: iapo mb planning}).

\section{Additional Results}
\label{appendix: additional results}

% \begin{wrapfigure}{r}{0.6\textwidth}
%     \begin{center}
%     \begin{subfigure}[t]{0.24\textwidth}
%         \centering
%         \includegraphics[width=\textwidth]{figures/mf_invertedpendulum2.pdf}
%         \caption{\tiny \texttt{InvertedPendulum-v2}}
%     \end{subfigure}%
%     ~ 
%     \begin{subfigure}[t]{0.24\textwidth}
%         \centering
%         \includegraphics[width=\textwidth]{figures/mf_inverteddoublependulum.pdf}
%         \caption{\tiny \texttt{InvertedDoublePendulum-v2}}
%     \end{subfigure}
%   \end{center}
%     \caption{\textbf{Performance Comparison}. Curves show the mean and $\pm$ std.~dev.~over $5$ random seeds.}
%     \label{fig: iapo inverted penulum}
% \end{wrapfigure}

\begin{figure}
\centering
\begin{minipage}{0.65\textwidth}
  \centering
  \begin{subfigure}[t]{0.38\textwidth}
        \centering
        \includegraphics[width=\textwidth]{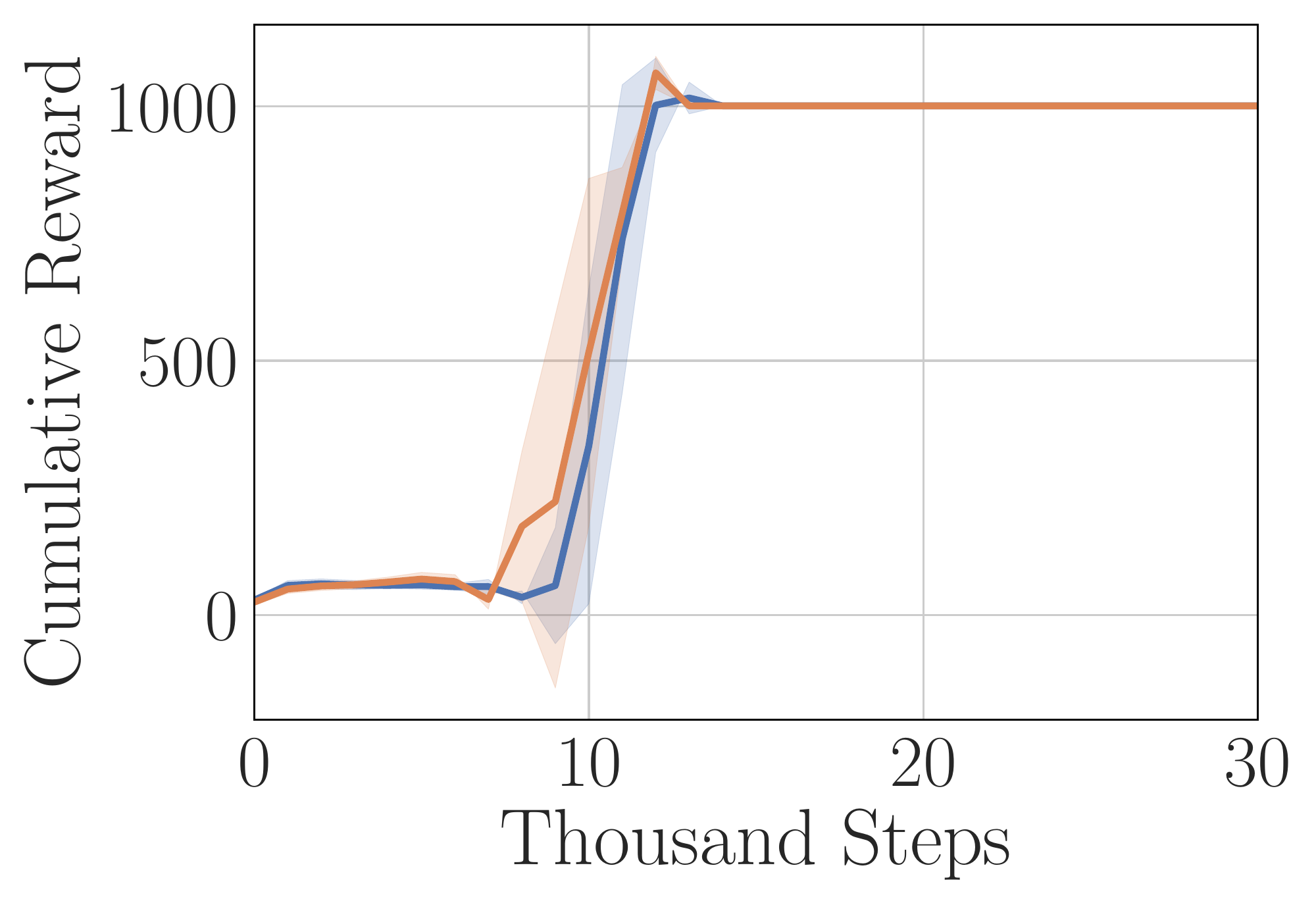}
        \caption{\tiny \texttt{InvertedPendulum-v2}}
    \end{subfigure}%
    ~ 
    \begin{subfigure}[t]{0.38\textwidth}
        \centering
        \includegraphics[width=\textwidth]{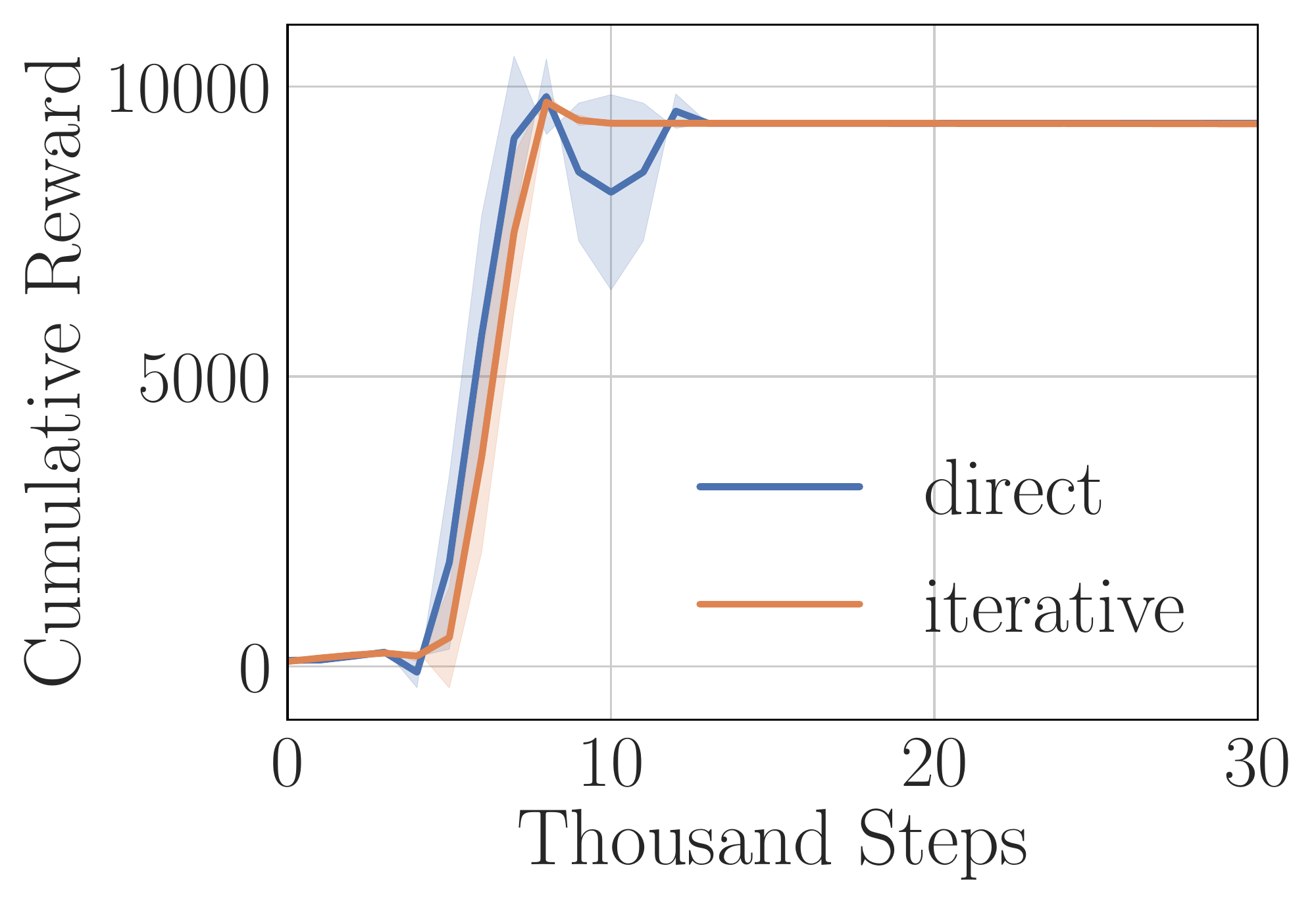}
        \caption{\tiny \texttt{InvertedDoublePendulum-v2}}
    \end{subfigure}
    \caption{\textbf{Pendulum Performance Comparison}. Curves show the mean and $\pm$ std.~dev.~over $5$ random seeds.}
    \label{fig: iapo inverted penulum}
\end{minipage}% 
~ \thickspace \thickspace \thickspace
\begin{minipage}{0.32\textwidth}
  \centering
  \includegraphics[width=0.88\textwidth]{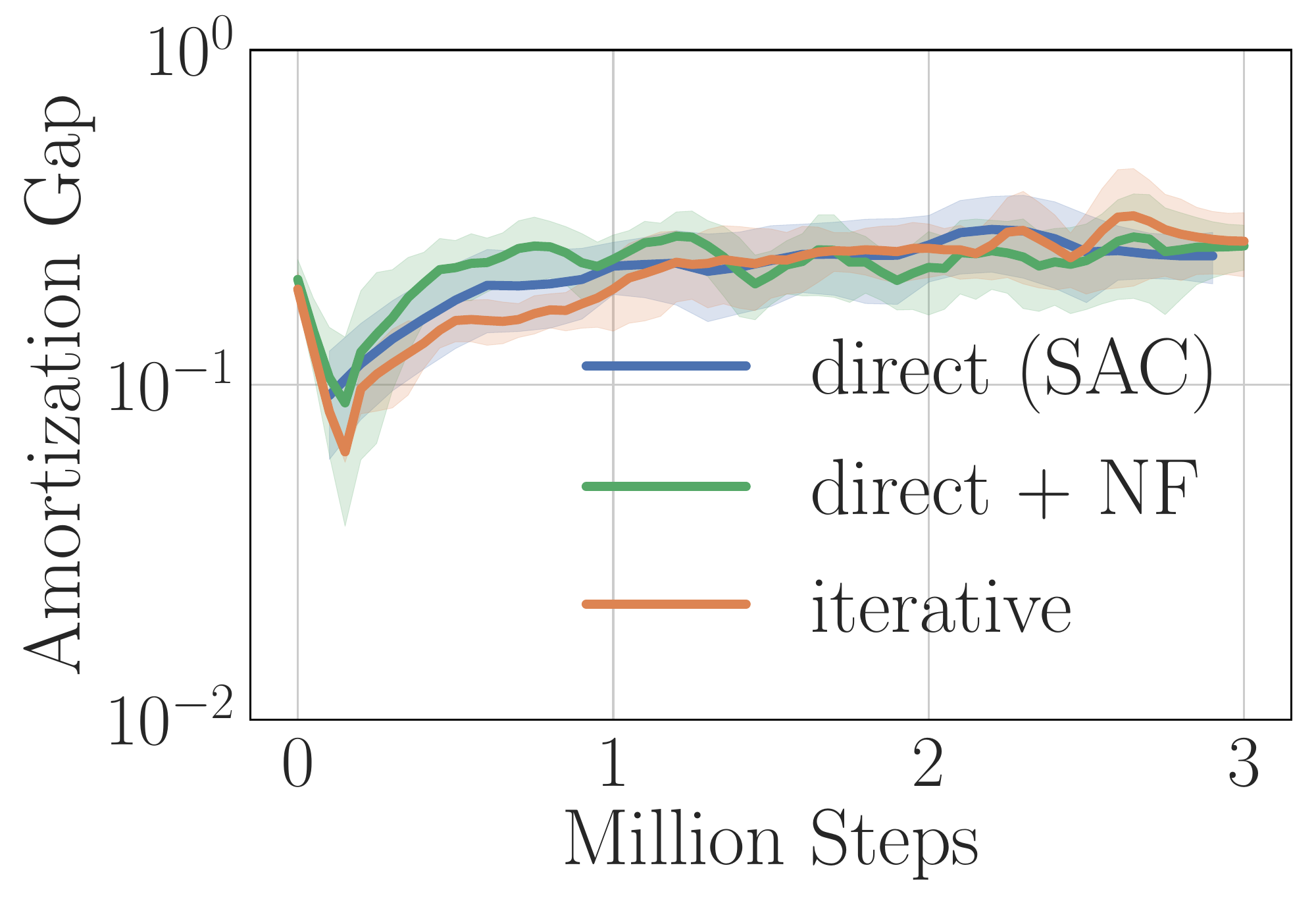}
    \caption{\textbf{Amortization Gap} for direct + NF on \texttt{HalfCheetah-v2}.}
    \label{fig: iapo hc nf am gap}
\end{minipage}
\end{figure}

\subsection{Pendulum Environments}

In Figure \ref{fig: iapo inverted penulum}, we present additional results on the remaining two MuJoCo environments from OpenAI gym: \texttt{InvertedPendulum-v2} and \texttt{InvertedDoublePendulum-v2}. The curves show the mean and standard deviation over $5$ random seeds. Iterative amortization performs comparably with direct amortization on these simpler environments.

\begin{figure}[t!]
    \centering
    \begin{subfigure}[t]{0.24\textwidth}
        \centering
        \includegraphics[width=\textwidth]{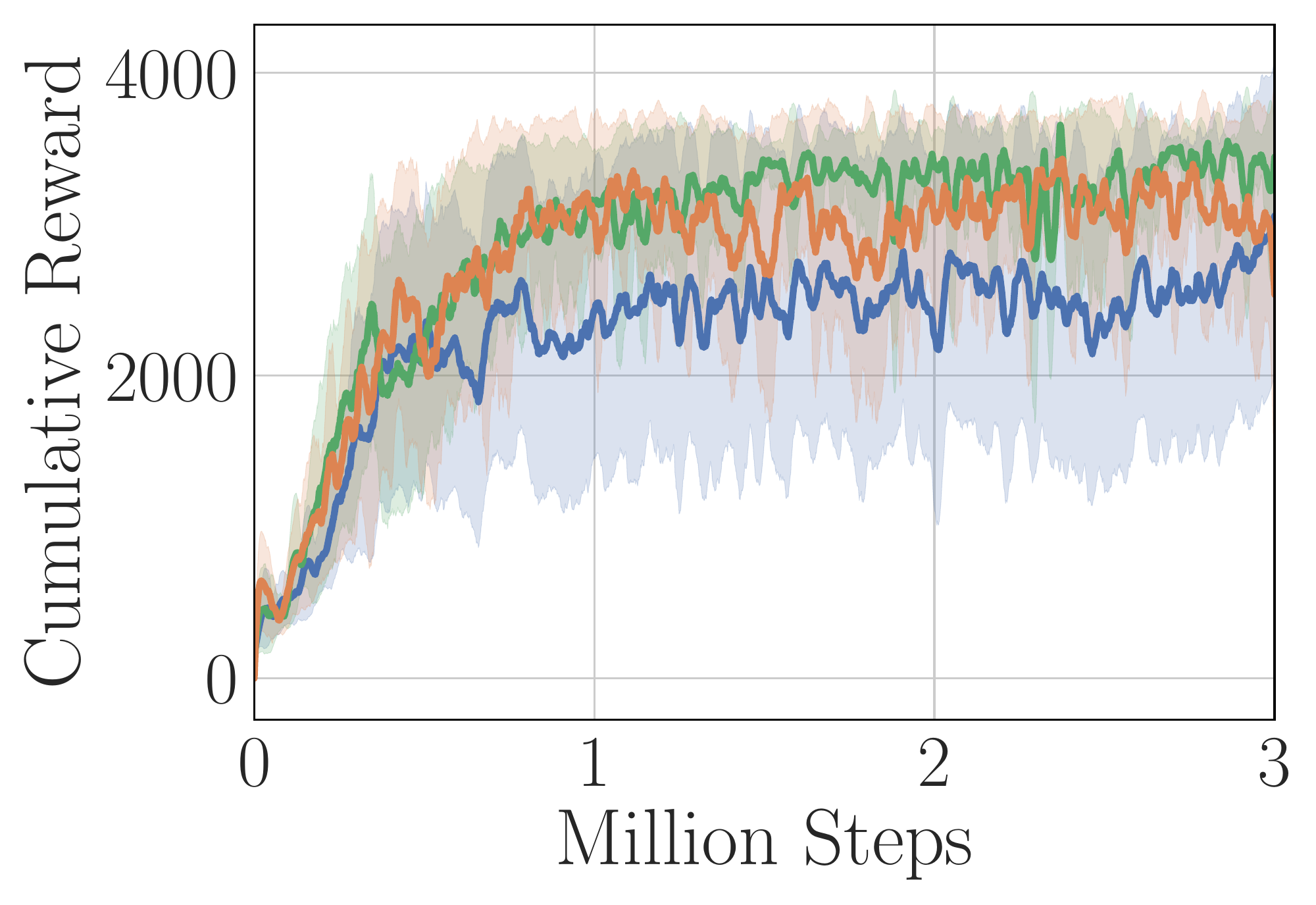}
        \caption{\scriptsize \texttt{Hopper-v2}}
    \end{subfigure}%
    ~
    \begin{subfigure}[t]{0.24\textwidth}
        \centering
        \includegraphics[width=\textwidth]{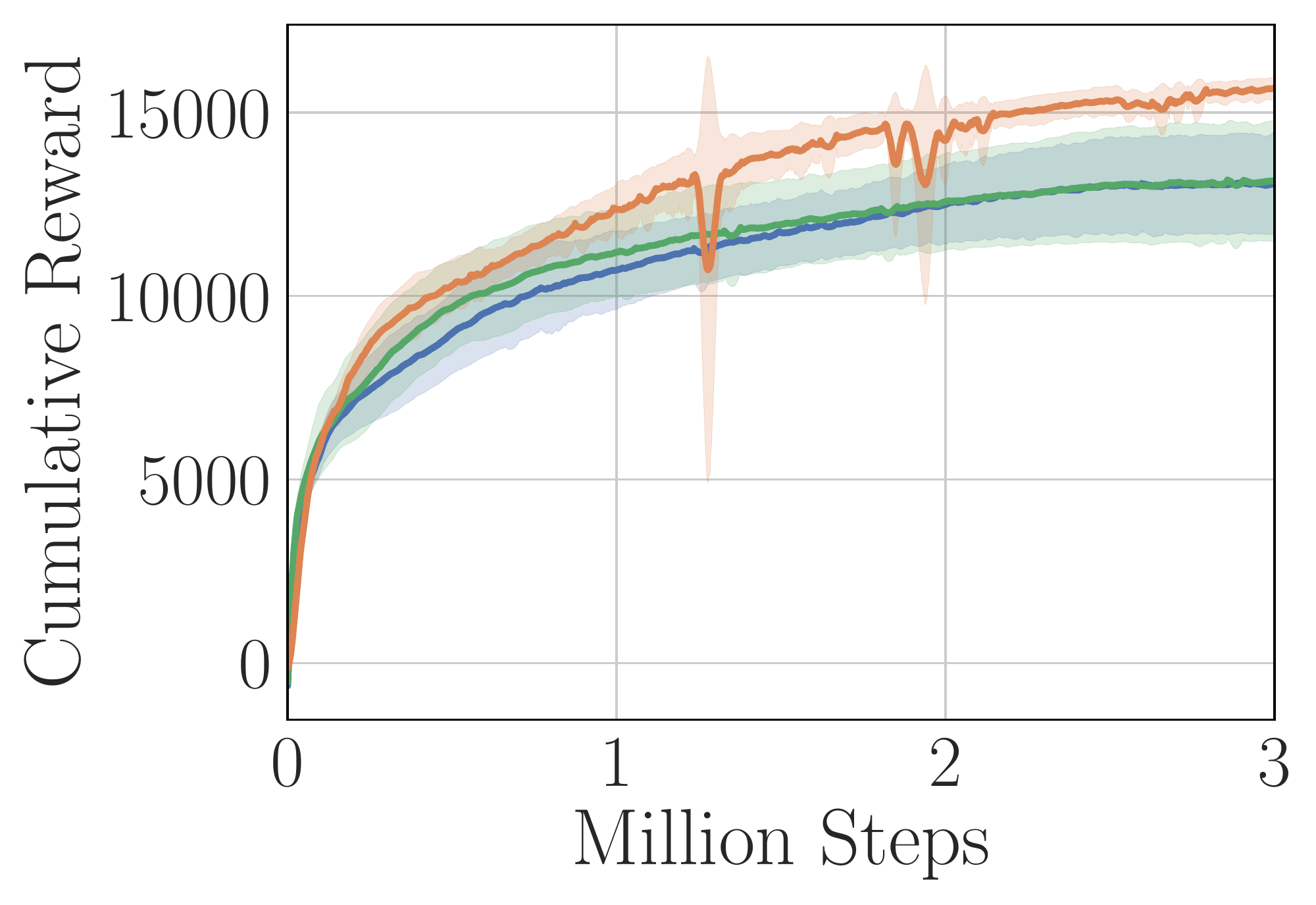}
        \caption{\scriptsize \texttt{HalfCheetah-v2}}
    \end{subfigure}%
    ~ 
    \begin{subfigure}[t]{0.24\textwidth}
        \centering
        \includegraphics[width=\textwidth]{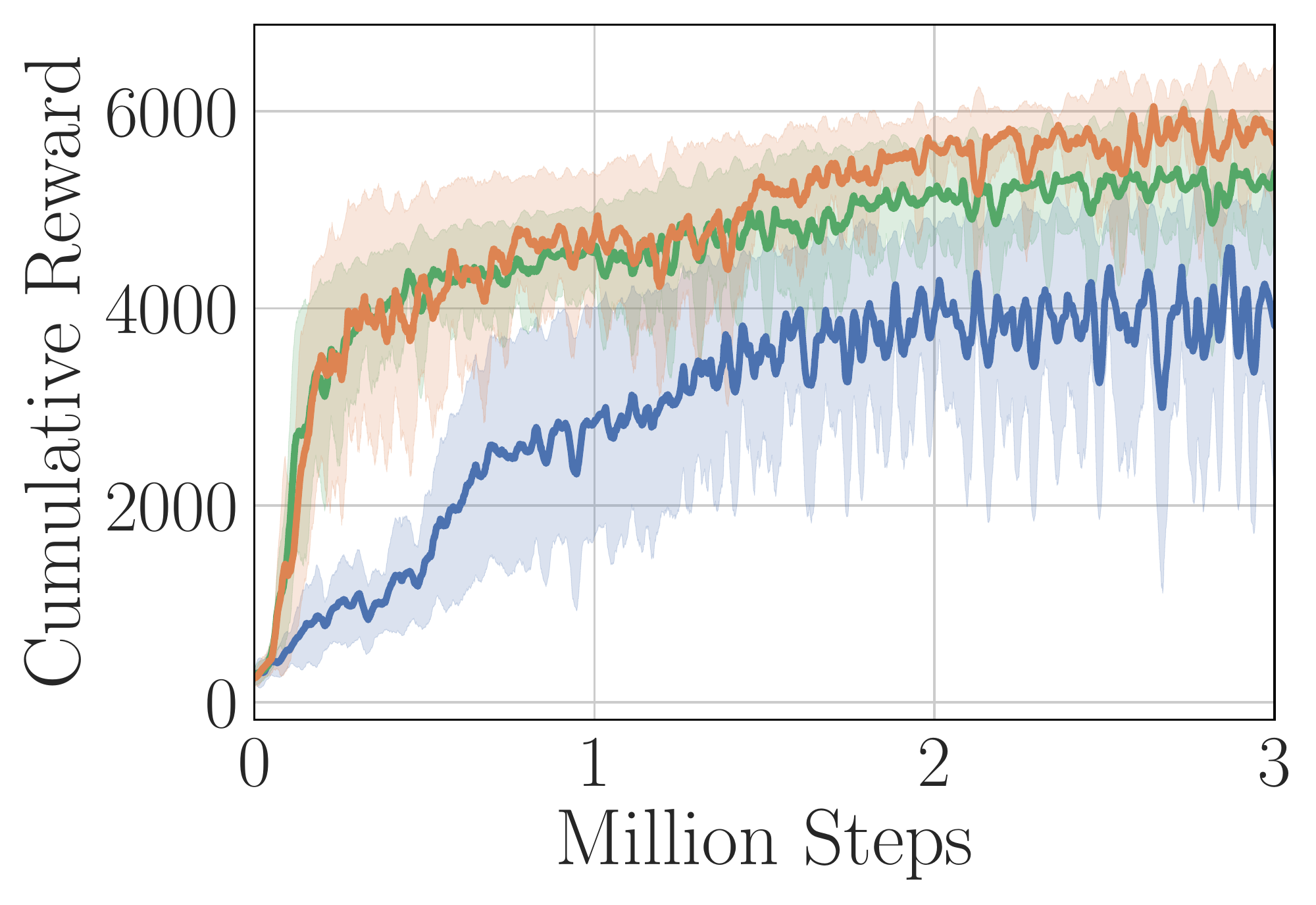}
        \caption{\scriptsize \texttt{Walker2d-v2}}
    \end{subfigure}%
    ~ 
    \begin{subfigure}[t]{0.24\textwidth}
        \centering
        \includegraphics[width=\textwidth]{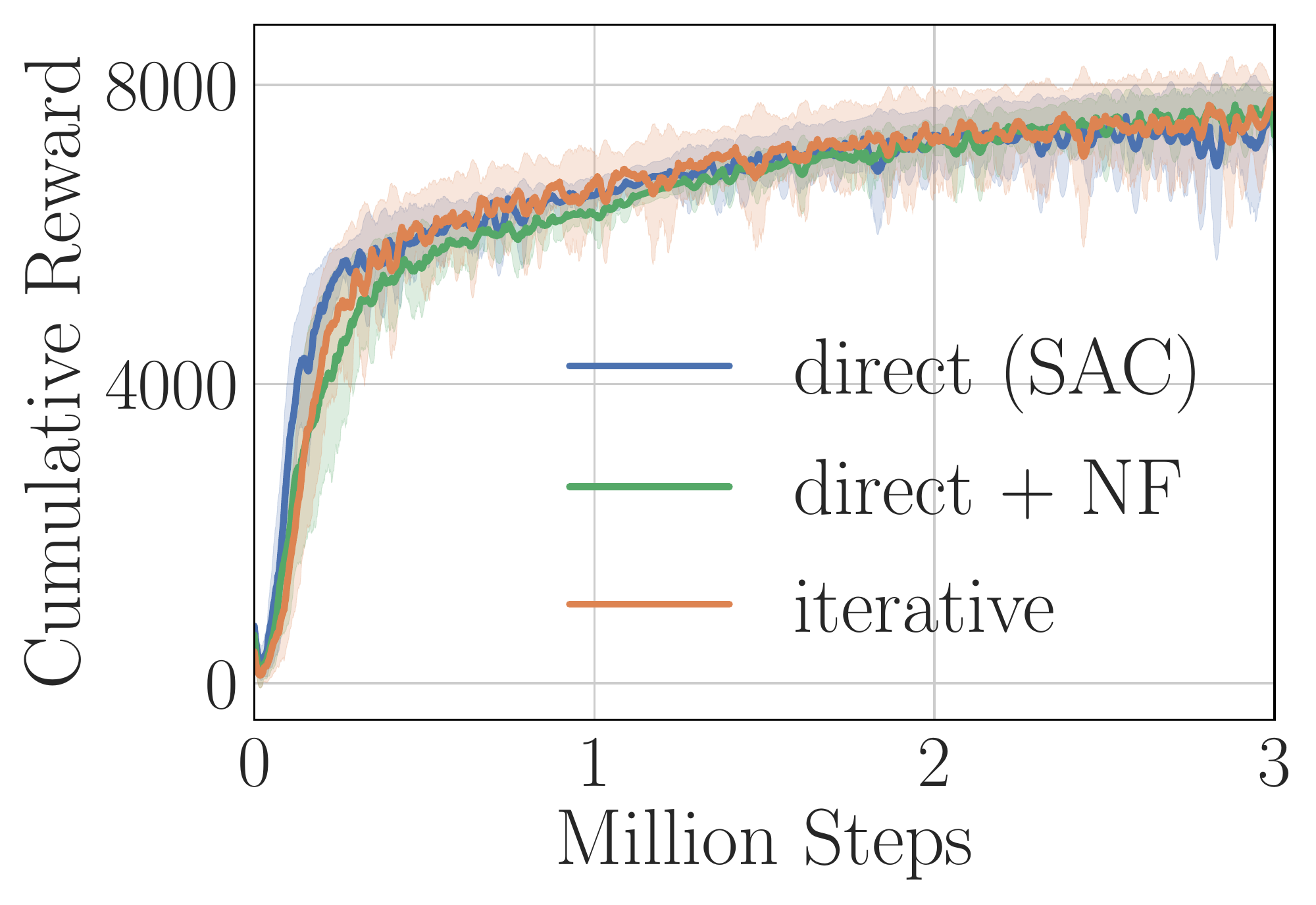}
        \caption{\scriptsize \texttt{Ant-v2}}
    \end{subfigure}
    \caption{\textbf{Performance Comparison with NF Policies}. Curves show the mean $\pm$ std.~dev.~of performance over $5$ random seeds.}
    \label{fig: nf performance}
\end{figure}

\subsection{Comparison with Normalizing Flow-Based Policies}
\label{appendix: comp with nf policies}

Iterative amortization is capable of estimating multiple policy modes, potentially yielding improved exploration. Thus, the benefits of iterative amortization may come purely from this effective improvement in the policy distribution. To test this hypothesis, we compare with direct amortization with normalizing flow-based (NF) policies, formed using two inverse autoregressive flow transforms \citep{kingma2016improved}. Each transform is parameterized by a network with $2$ layers of $256$ units with \texttt{ReLU} activation, and we reverse the action dimension ordering between the transforms to model dependencies in both directions. In Figure \ref{fig: nf performance}, we plot performance on a subset of environments, where we see that direct + NF closes the performance gap on \texttt{Hopper-v2} and \texttt{Walker2d-v2} but is unable to close the gap on \texttt{HalfCheetah-v2}. In Figure \ref{fig: iapo hc nf am gap}, we see that direct + NF is also unable to close the amortization gap early on during training. Thus, improved optimization of iterative amortization, rather than purely improved exploration, does appear to play some role in the performance improvements.

\begin{figure}[h!]
    \centering
    \begin{subfigure}[t]{0.24\textwidth}
        \centering
        \includegraphics[width=\textwidth]{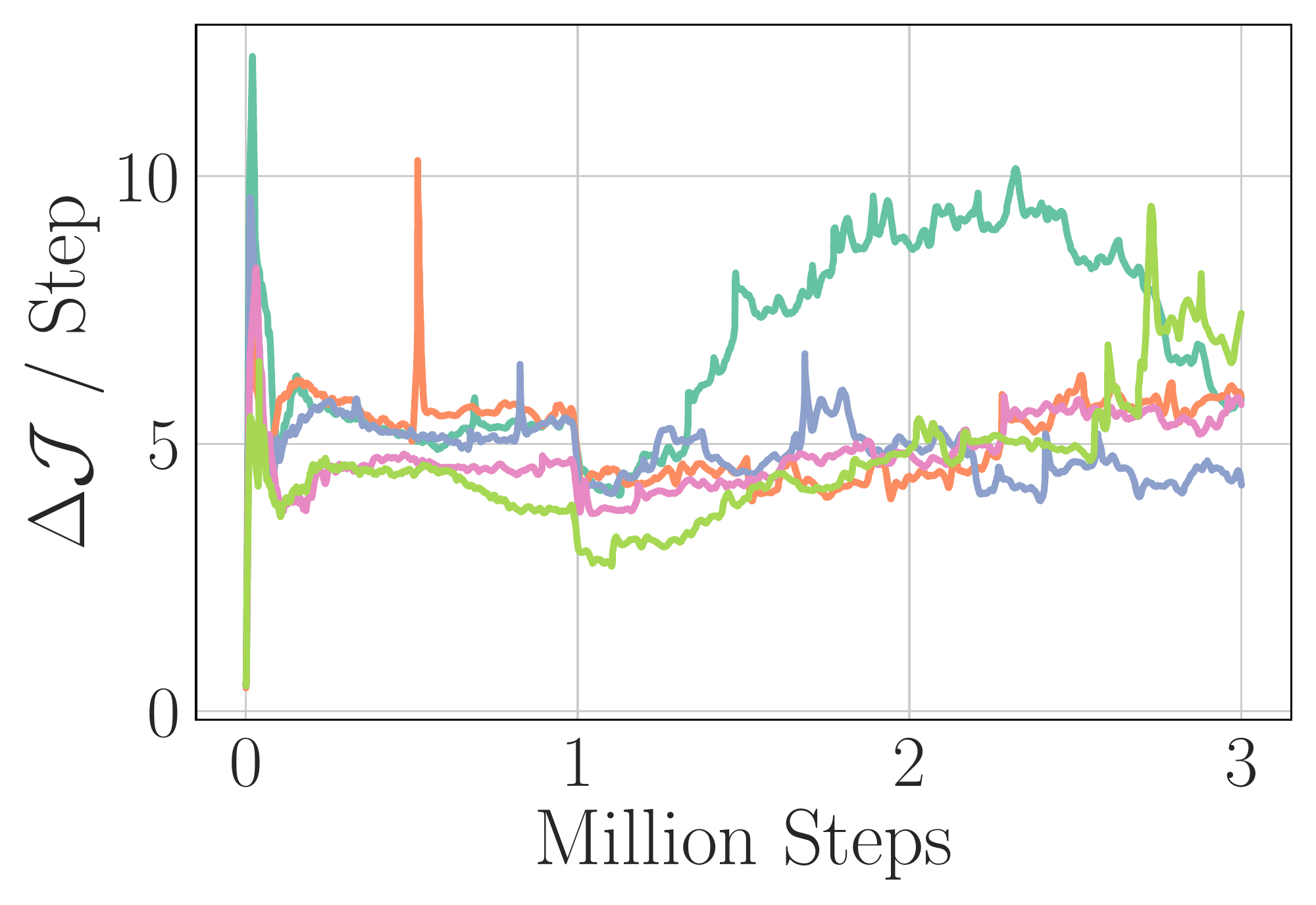}
        \caption{\texttt{Hopper-v2}}
    \end{subfigure}%
    ~ 
    \begin{subfigure}[t]{0.24\textwidth}
        \centering
        \includegraphics[width=\textwidth]{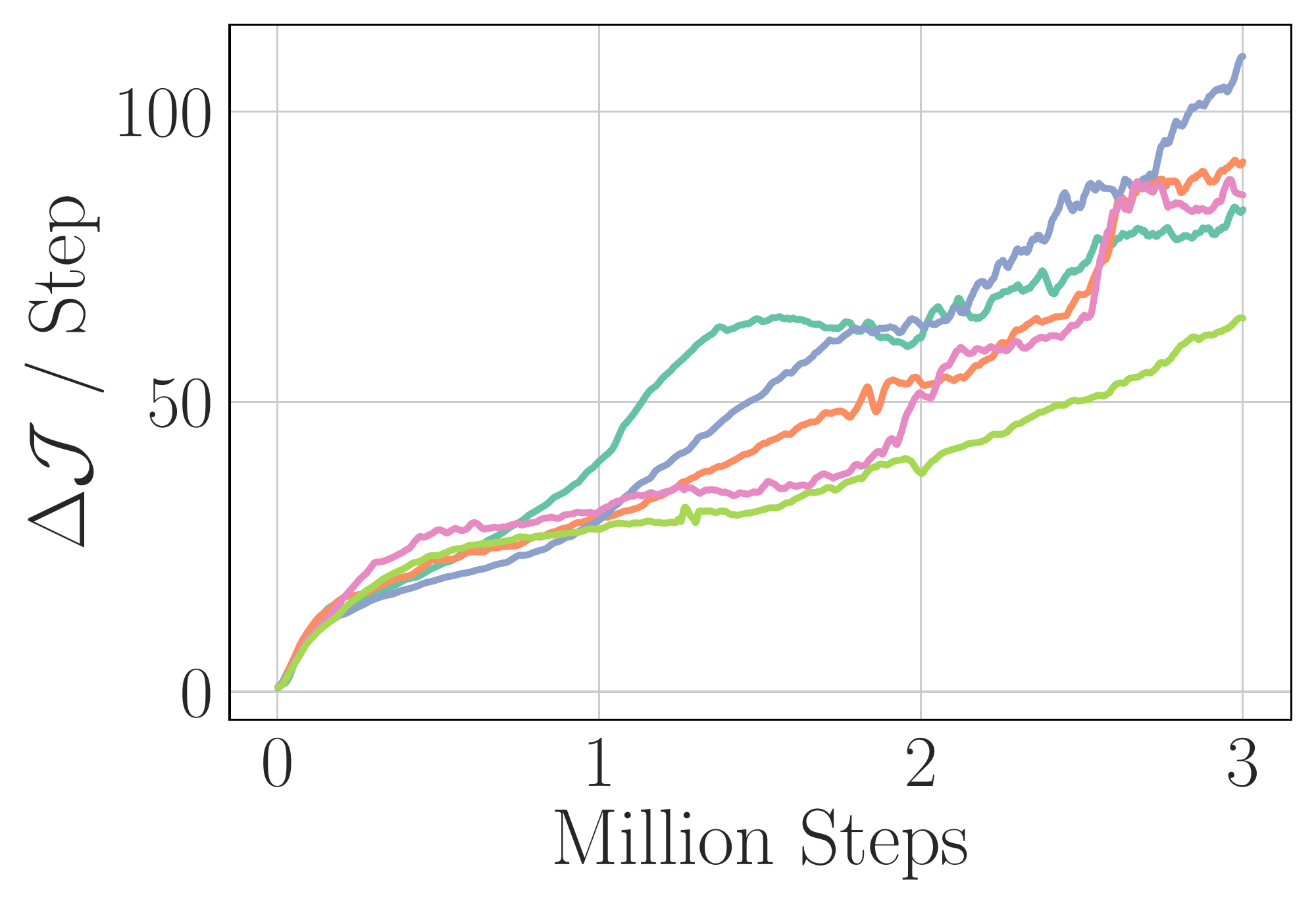}
        \caption{\texttt{HalfCheetah-v2}}
    \end{subfigure}%
    ~ 
    \begin{subfigure}[t]{0.24\textwidth}
        \centering
        \includegraphics[width=\textwidth]{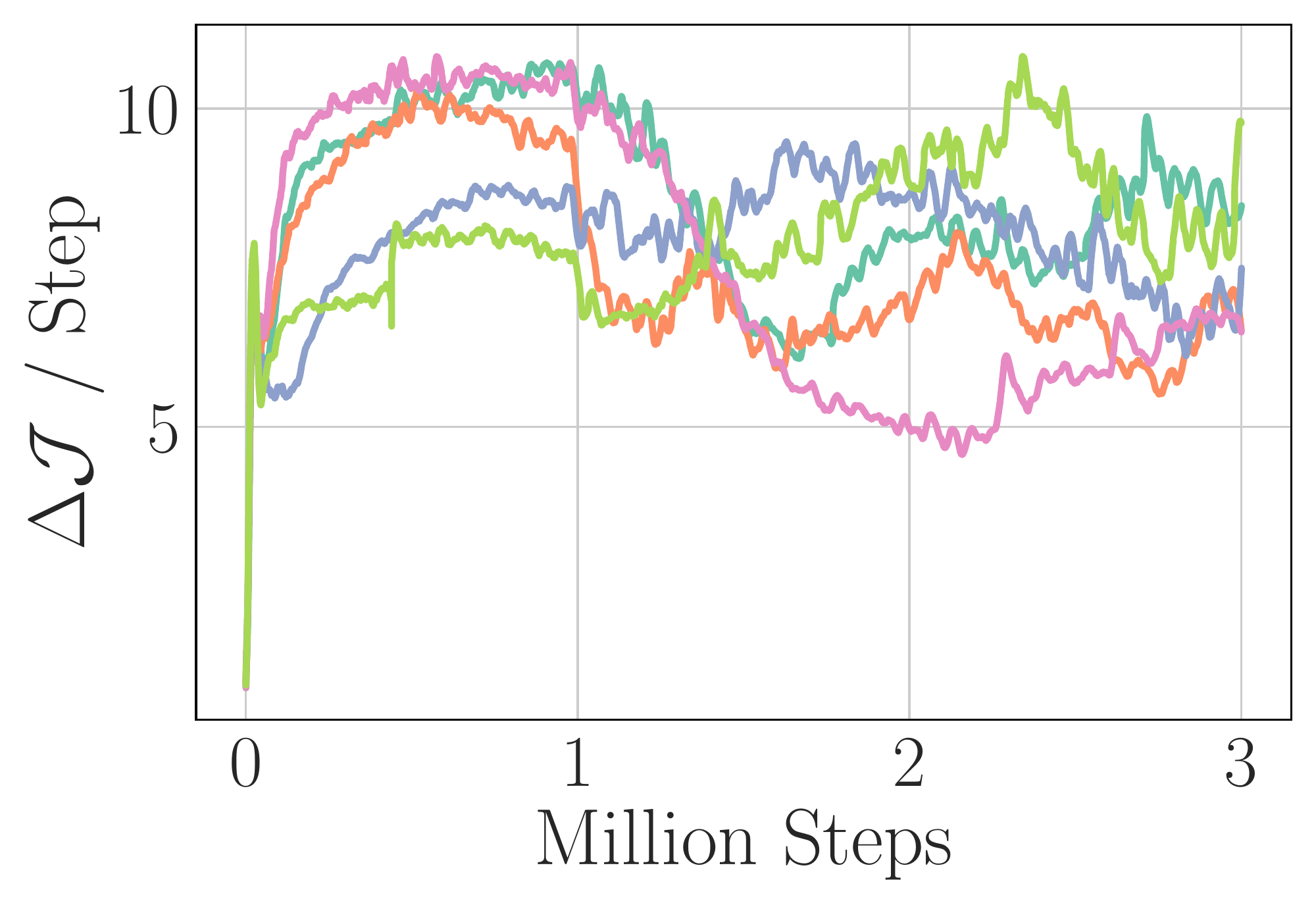}
        \caption{\texttt{Walker2d-v2}}
    \end{subfigure}%
    ~ 
    \begin{subfigure}[t]{0.24\textwidth}
        \centering
        \includegraphics[width=\textwidth]{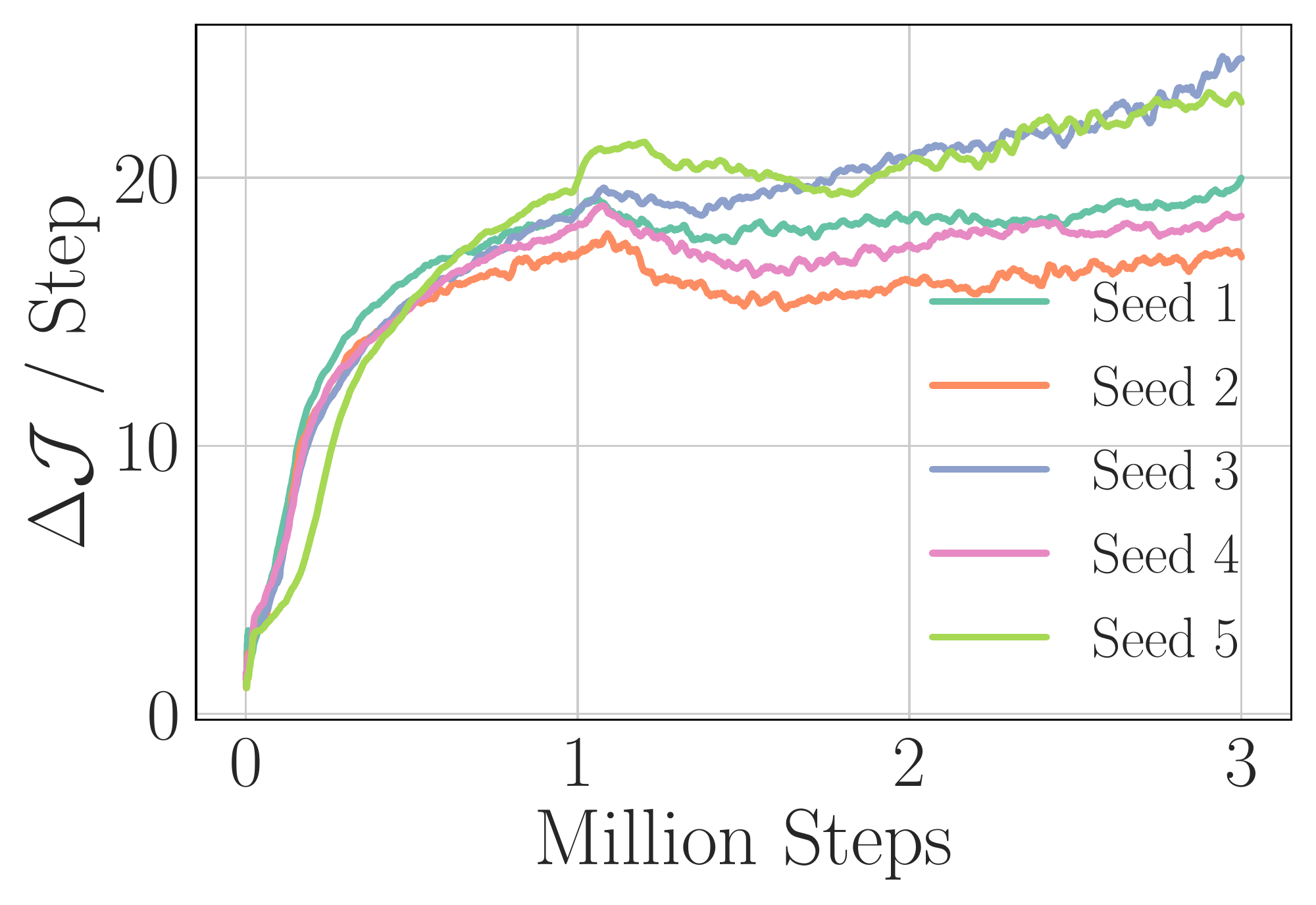}
        \caption{\texttt{Ant-v2}}
    \end{subfigure}
    \caption{\textbf{Per-Step Improvement}. Each plot shows the per-step improvement in the estimated variational RL objective, $\mathcal{J}$, throughout training resulting from iterative amortized policy optimization. Each curve denotes a different random seed.}
    \label{fig: per step inf imp}
\end{figure}

\subsection{Improvement per Step}

In Figure~\ref{fig: per step inf imp}, we plot the average improvement in the variational objective per step throughout training, with each curve showing a different random seed. That is, each plot shows the average change in the variational objective after running $5$ iterations of iterative amortized policy optimization. With the exception of \texttt{HalfCheetah-v2}, the improvement remains relatively constant throughout training and consistent across seeds.

\begin{figure}[t!]
    \centering
    \begin{subfigure}[t]{0.24\textwidth}
        \centering
        \includegraphics[width=\textwidth]{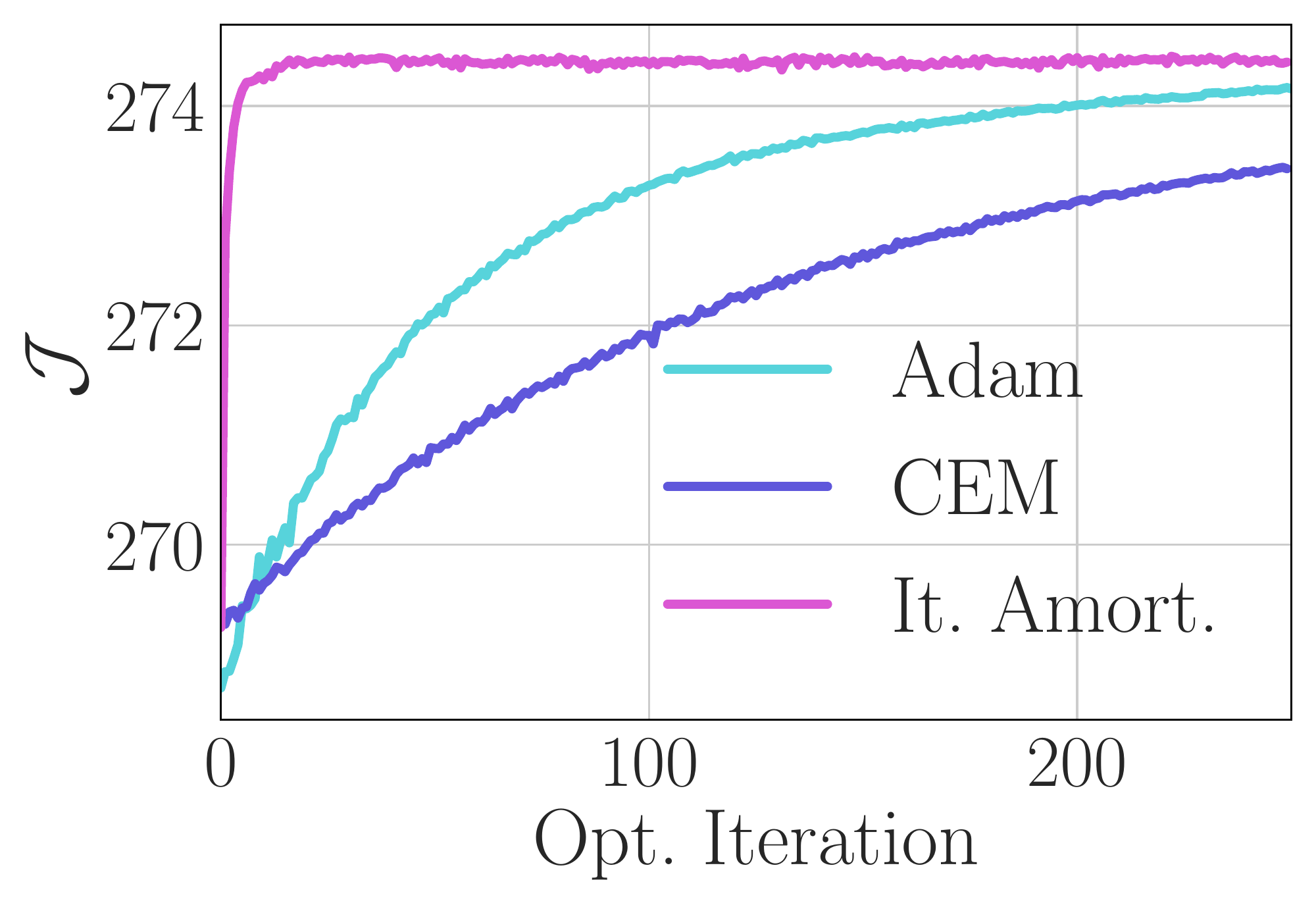}
        \caption{\tiny \texttt{Hopper-v2}}
    \end{subfigure}%
    ~ 
    \begin{subfigure}[t]{0.24\textwidth}
        \centering
        \includegraphics[width=\textwidth]{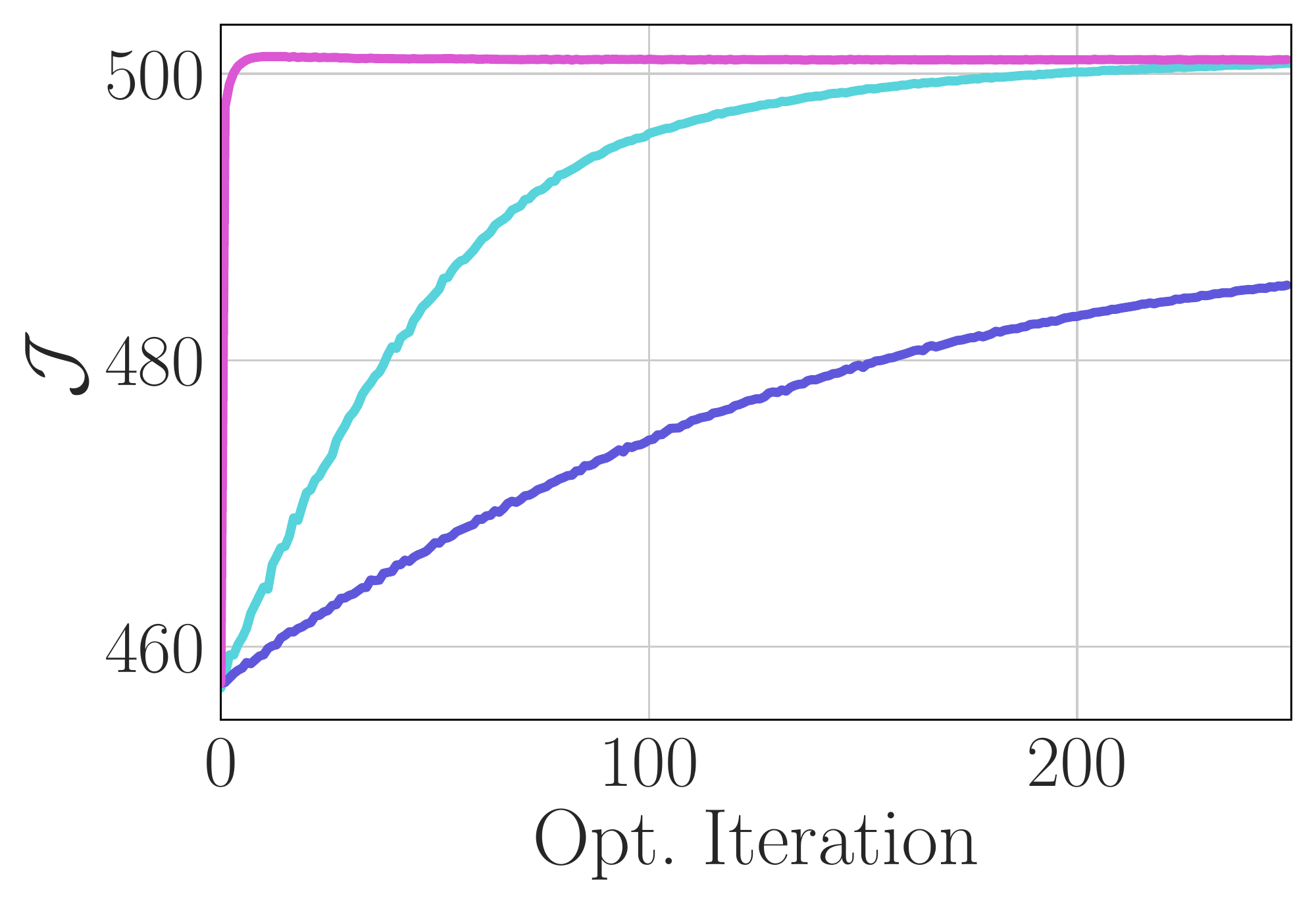}
        \caption{\tiny \texttt{HalfCheetah-v2}}
    \end{subfigure}%
    ~ 
    \begin{subfigure}[t]{0.24\textwidth}
        \centering
        \includegraphics[width=\textwidth]{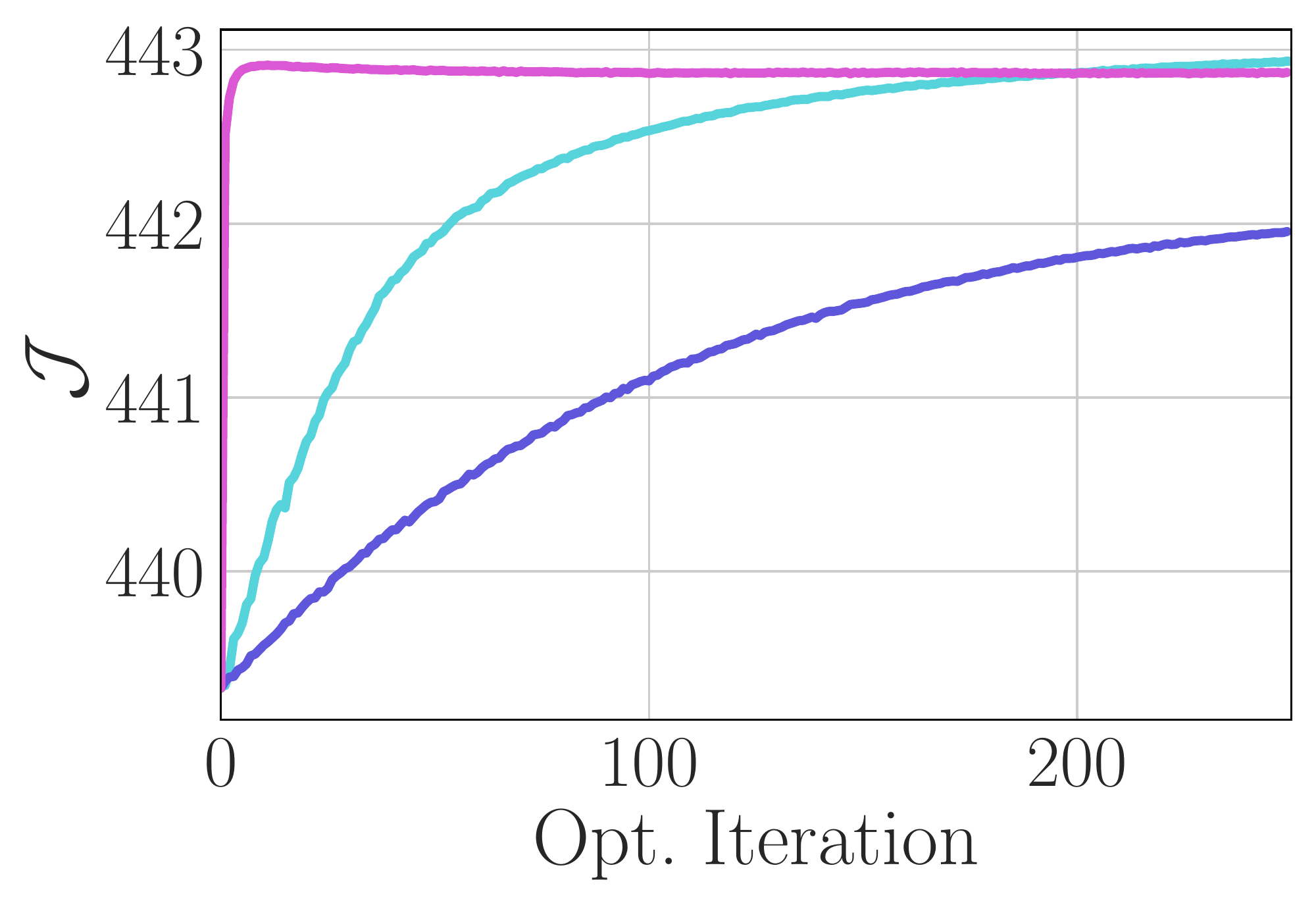}
        \caption{\tiny \texttt{Walker2d-v2}}
    \end{subfigure}%
    ~ 
    \begin{subfigure}[t]{0.24\textwidth}
        \centering
        \includegraphics[width=\textwidth]{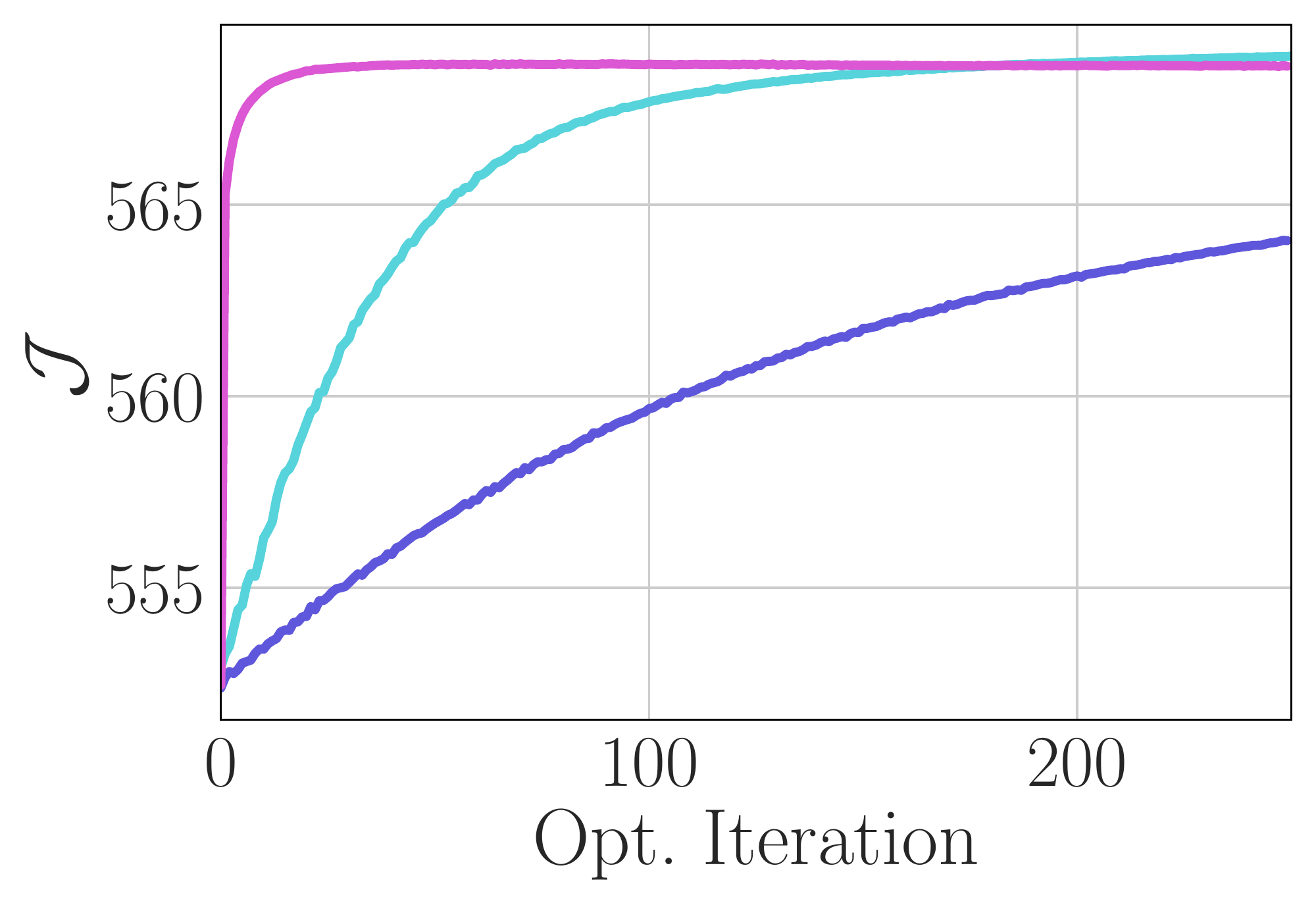}
        \caption{\tiny \texttt{Ant-v2}}
    \end{subfigure}
    
    \begin{subfigure}[t]{0.24\textwidth}
        \centering
        \includegraphics[width=\textwidth]{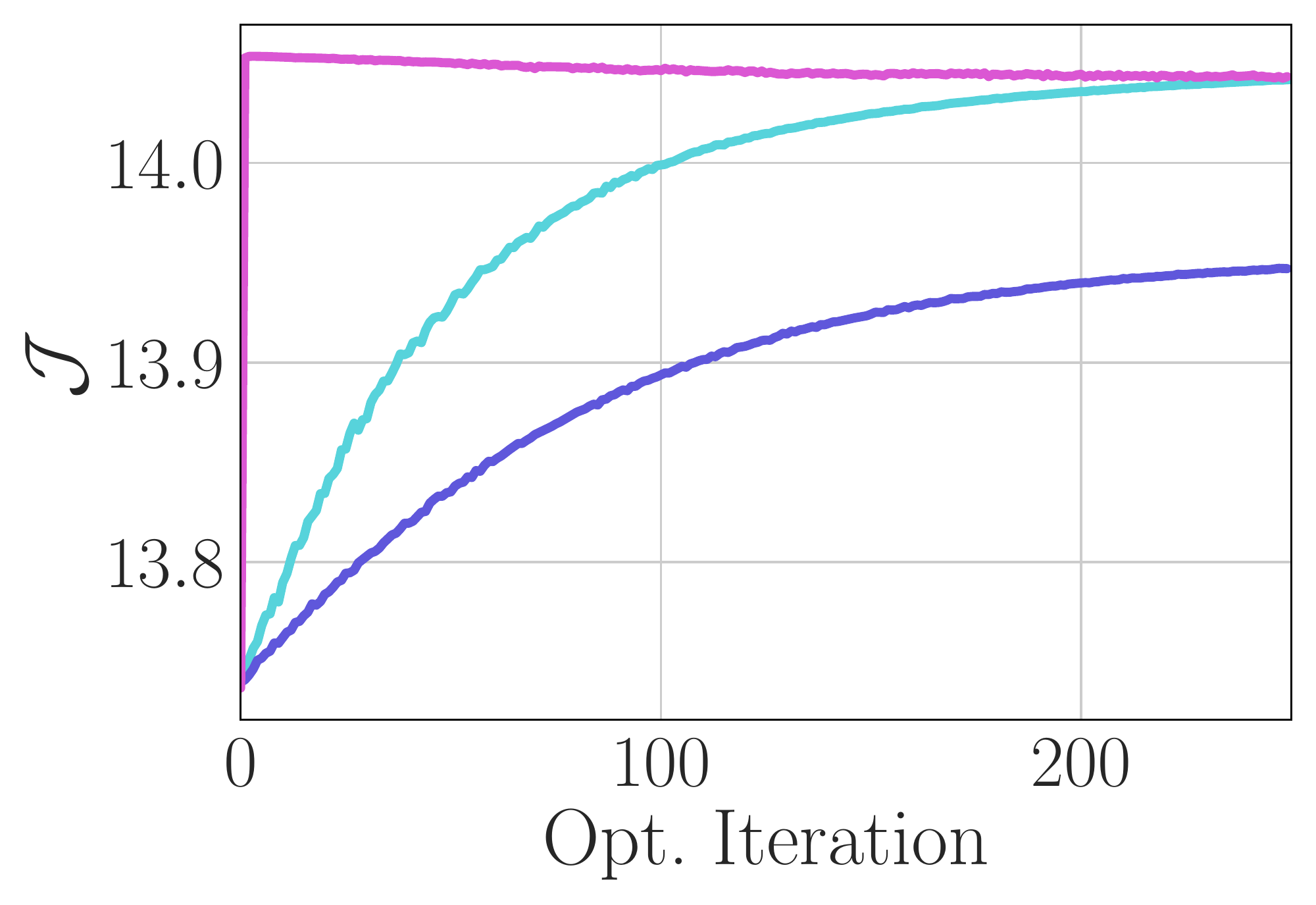}
        \caption{\tiny \texttt{Swimmer-v2}}
    \end{subfigure}%
    ~ 
    \begin{subfigure}[t]{0.24\textwidth}
        \centering
        \includegraphics[width=\textwidth]{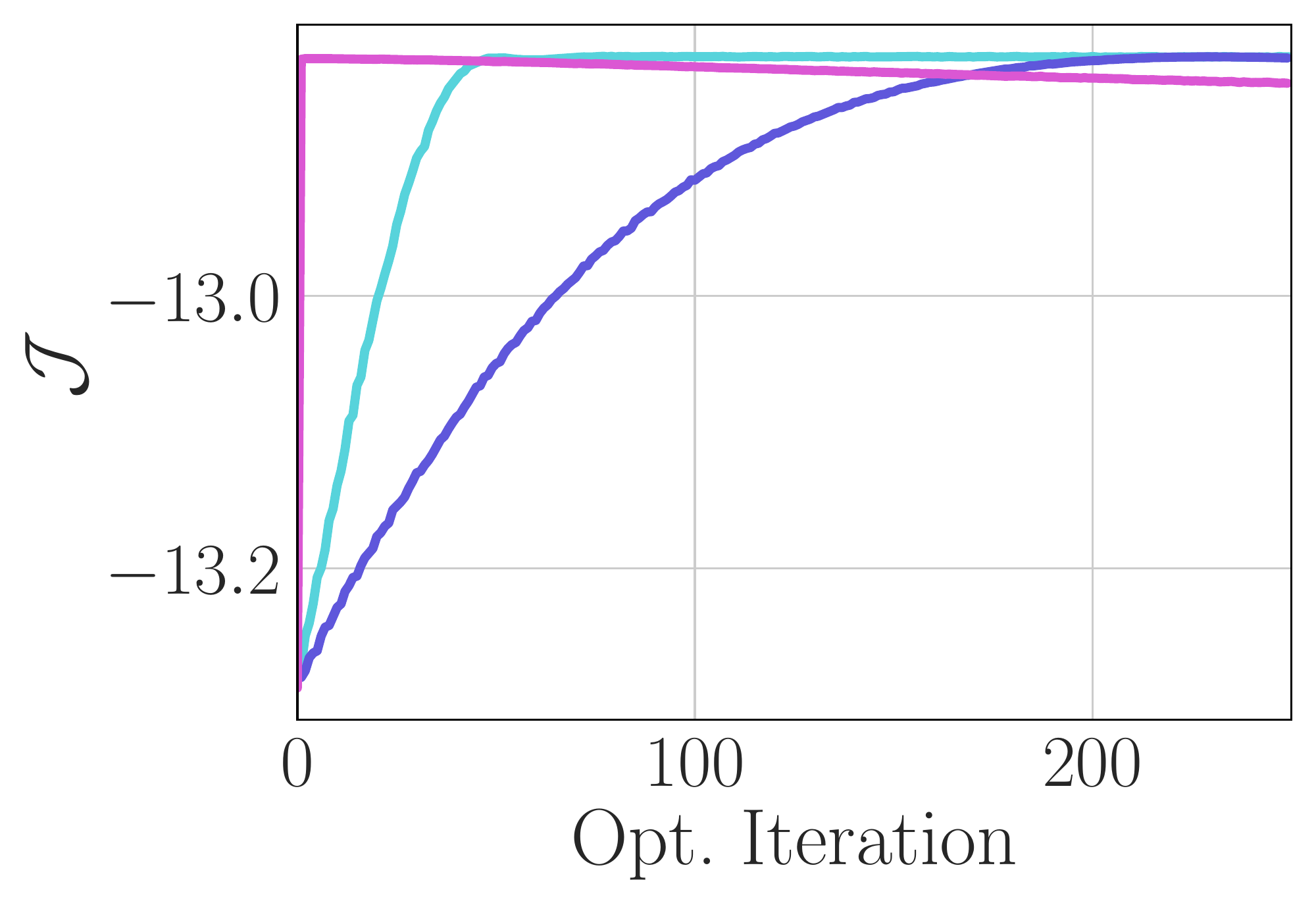}
        \caption{\tiny \texttt{Reacher-v2}}
    \end{subfigure}%
    ~ 
    \begin{subfigure}[t]{0.24\textwidth}
        \centering
        \includegraphics[width=\textwidth]{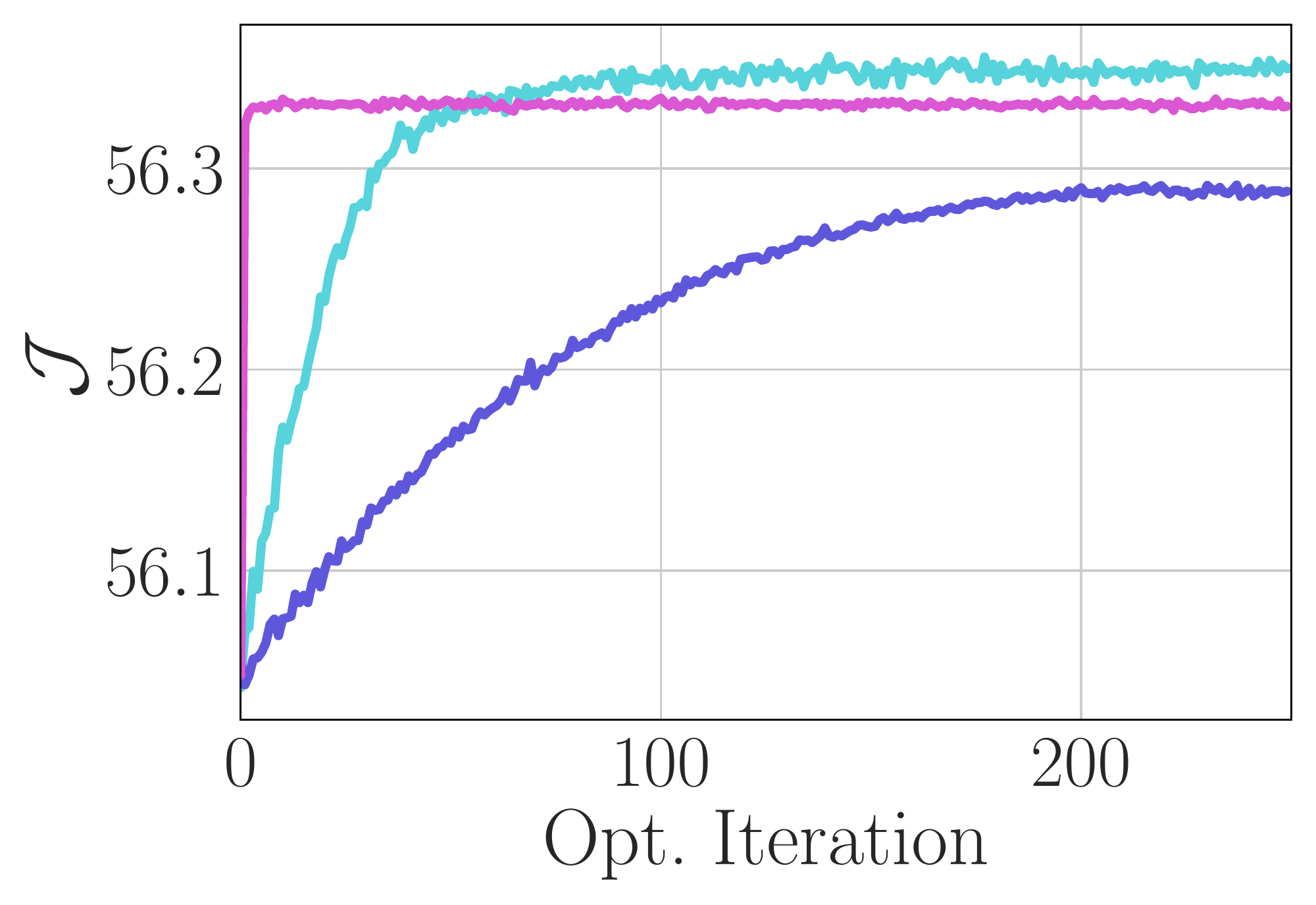}
        \caption{\tiny \texttt{InvertedPendulum-v2}}
    \end{subfigure}%
    ~ 
    \begin{subfigure}[t]{0.24\textwidth}
        \centering
        \includegraphics[width=\textwidth]{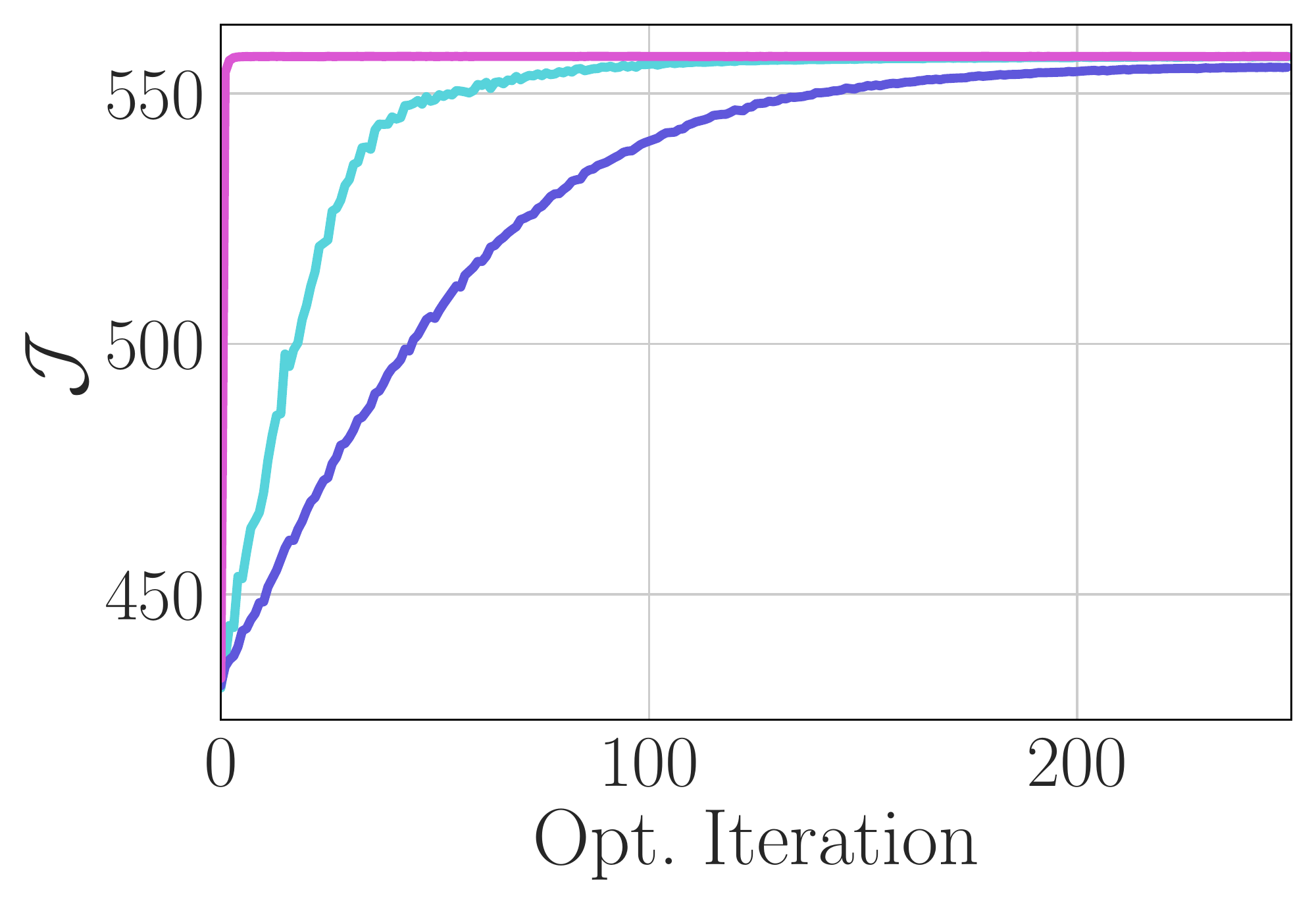}
        \caption{\tiny \texttt{InvertedDoublePendulum-v2}}
    \end{subfigure}
    \caption{\textbf{Comparison with Iterative Optimizers}. Average estimated objective over policy optimization iterations, comparing with Adam \citep{kingma2014adam} and CEM \citep{rubinstein2013cross}. These iterative optimizers require over an order of magnitude more iterations to reach comparable performance with iterative amortization, making them impractical in many applications.}
    \label{fig: iapo comp grad}
\end{figure}

\begin{figure*}[t!]
    \centering
    \begin{subfigure}[t]{0.32\textwidth}
        \centering
        \includegraphics[width=\textwidth]{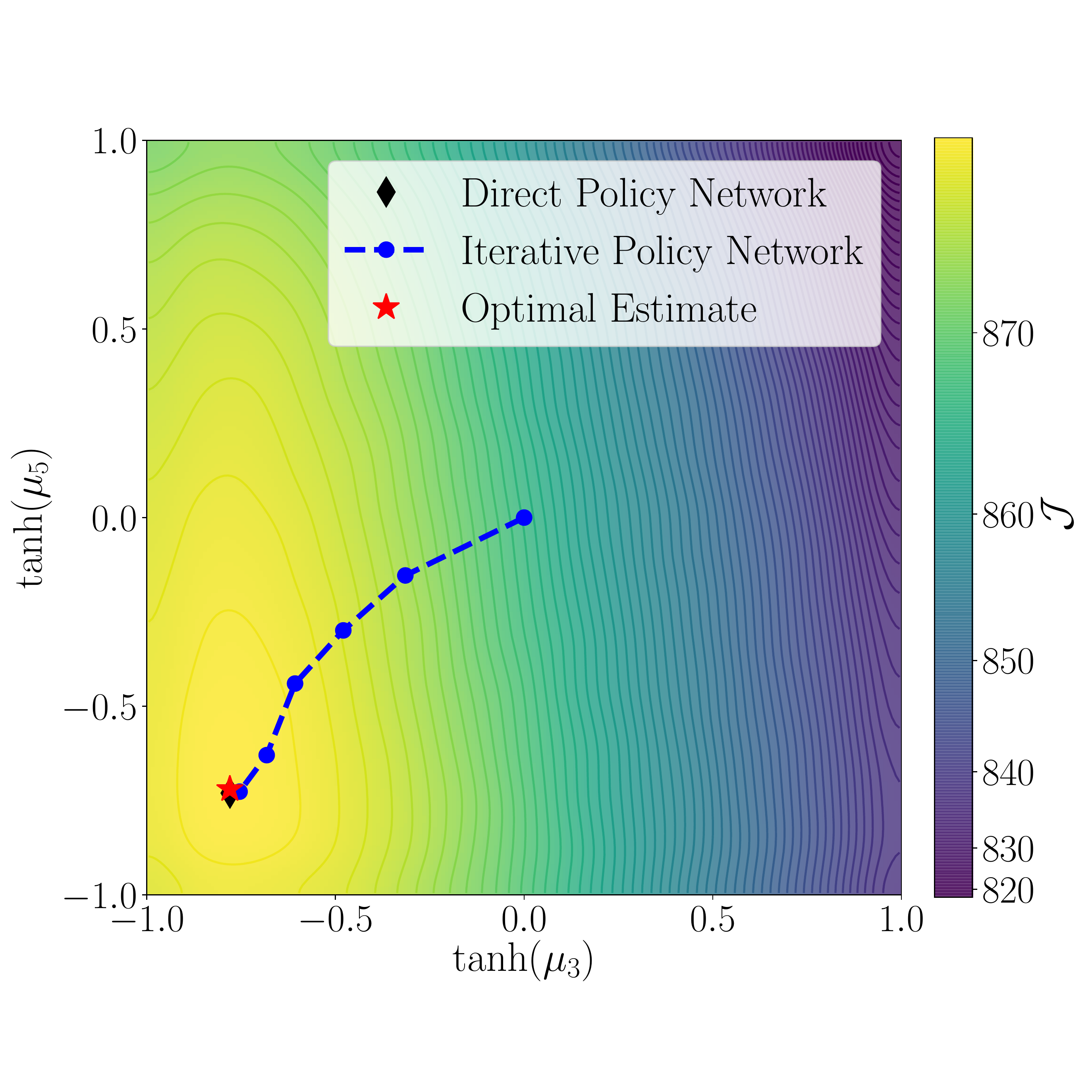}
        \caption{\texttt{HalfCheetah-v2}}
    \end{subfigure}%
    ~ 
    \begin{subfigure}[t]{0.32\textwidth}
        \centering
        \includegraphics[width=\textwidth]{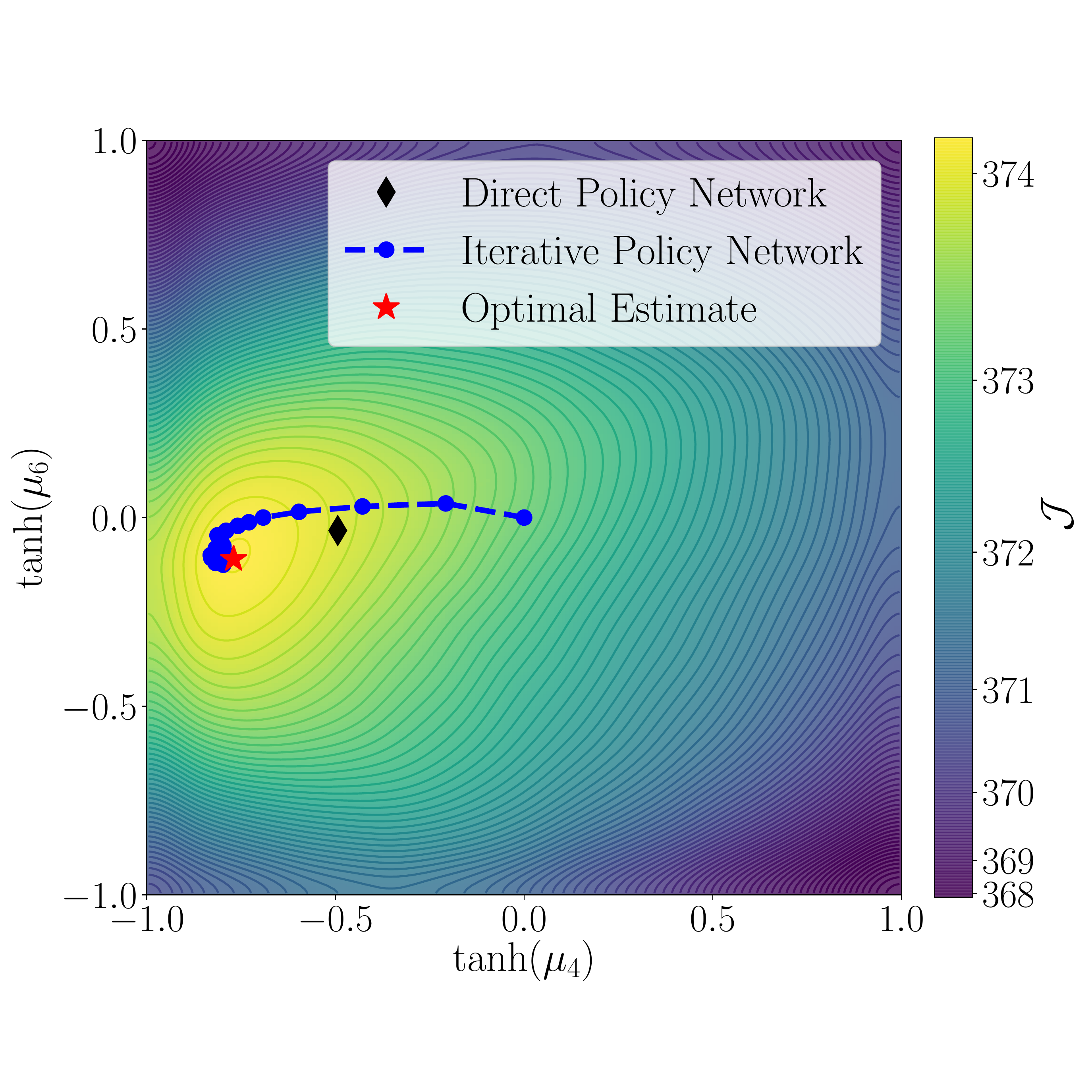}
        \caption{\texttt{Walker2d-v2}}
    \end{subfigure}%
    ~ 
    \begin{subfigure}[t]{0.32\textwidth}
        \centering
        \includegraphics[width=\textwidth]{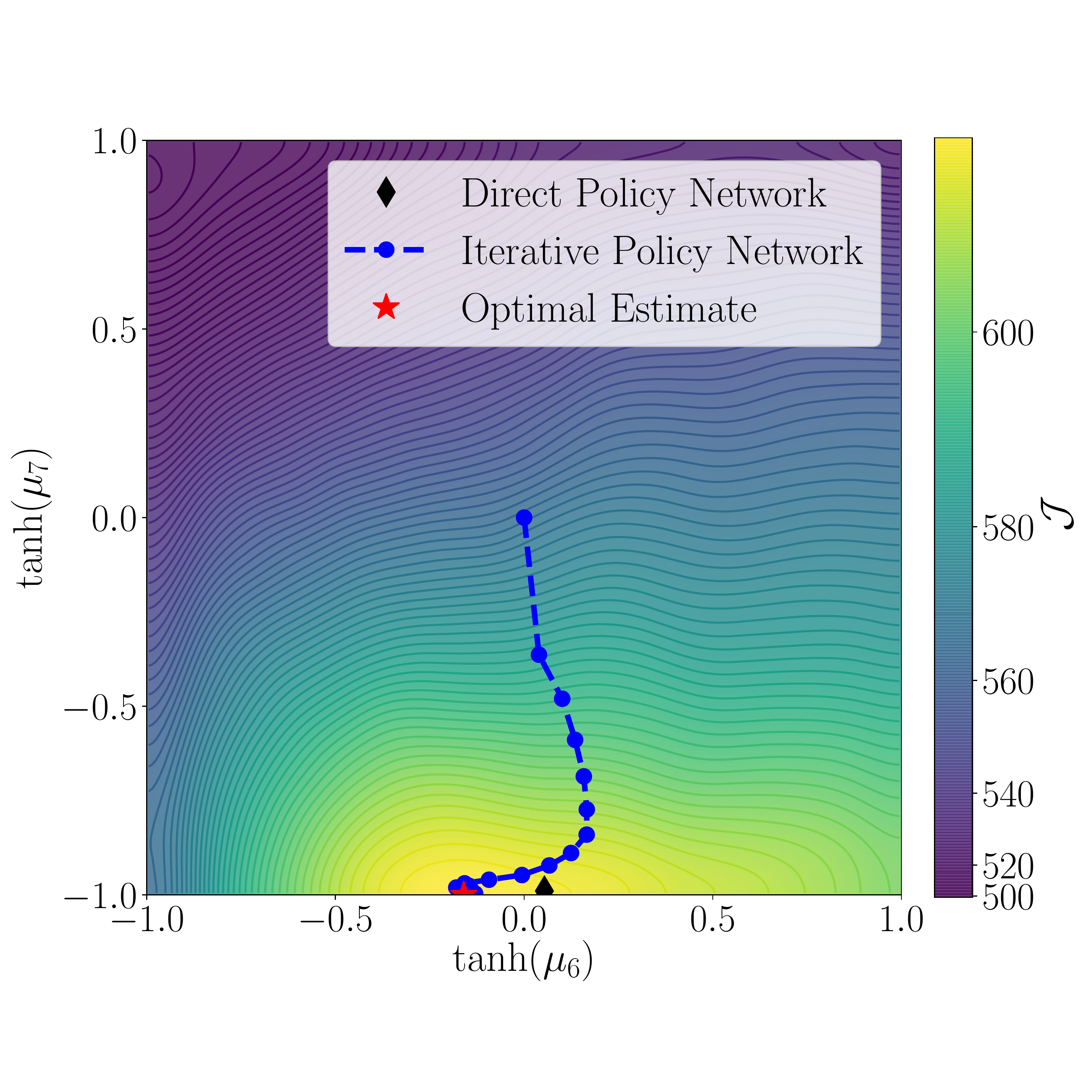}
        \caption{\texttt{Ant-v2}}
    \end{subfigure}
    \caption{\textbf{2D Optimization Plots}. Each plot shows the optimization objective over two dimensions of the policy mean, $\bm{\mu}$. This optimization surface contains the value function trained using a direct amortized policy. The black diamond, denoting the estimate of this direct policy, is generally near-optimal, but does not match the optimal estimate (red star). Iterative amortized optimizers are capable of generalizing to these surfaces in each case, reaching optimal policy estimates.}
    \label{fig: iapo additional 2d plots}
\end{figure*}

\subsection{Comparison with Iterative Optimizers}
\label{sec: iapo comparison with iterative optimizers}

% Iterative amortized policy optimization obtains the \textit{accuracy} benefits of iterative optimization while retaining the \textit{efficiency} benefits of amortization. In Section~\ref{sec: iapo experiments}, we compared the accuracy of iterative and direct amortization, seeing that iterative amortization yields reduced amortization gaps (Figure~\ref{fig: iapo amortization gap}) and improved performance (Figure~\ref{fig: model-free performance}). In this section, we compare iterative amortization with two popular iterative optimizers: Adam \citep{kingma2014adam}, a gradient-based optimizer, and cross-entropy method (CEM) \citep{rubinstein2013cross}, a gradient-free optimizer.

Iterative amortized policy optimization obtains the \textit{accuracy} benefits of iterative optimization while retaining the \textit{efficiency} benefits of amortization. In Section 4, we compared the accuracy of iterative and direct amortization, seeing that iterative amortization yields reduced amortization gaps (Figure 6) and improved performance (Figure 5). In this section, we compare iterative amortization with two popular iterative optimizers: Adam \citep{kingma2014adam}, a gradient-based optimizer, and cross-entropy method (CEM) \citep{rubinstein2013cross}, a gradient-free optimizer.

% To compare the accuracy and efficiency of the optimizers, we collect $100$ states for each seed in each environment from the model-free experiments in Section~\ref{sec: iapo mf perf comp}. For each optimizer, we optimize the variational objective, $\mathcal{J}$, starting from the same initialization. Tuning the step size, we found that $0.01$ yielded the steepest improvement without diverging for both Adam and CEM. Gradients are evaluated with $10$ action samples. For CEM, we sample $100$ actions and fit a Gaussian mean and variance to the top $10$ samples. This is comparable with QT-Opt \citep{kalashnikov2018qt}, which draws $64$ samples and retains the top $6$ samples.

To compare the accuracy and efficiency of the optimizers, we collect $100$ states for each seed in each environment from the model-free experiments in Section 4.2.2. For each optimizer, we optimize the variational objective, $\mathcal{J}$, starting from the same initialization. Tuning the step size, we found that $0.01$ yielded the steepest improvement without diverging for both Adam and CEM. Gradients are evaluated with $10$ action samples. For CEM, we sample $100$ actions and fit a Gaussian mean and variance to the top $10$ samples. This is comparable with QT-Opt \citep{kalashnikov2018qt}, which draws $64$ samples and retains the top $6$ samples.

The results, averaged across states and random seeds, are shown in Figure~\ref{fig: iapo comp grad}. CEM (gradient-free) is less efficient than Adam (gradient-based), which is unsurprising, especially considering that Adam effectively approximates higher-order curvature through momentum terms. However, Adam and CEM both require over \textit{an order of magnitude} more iterations to reach comparable performance with iterative amortization. While iterative amortized policy optimization does not always obtain asymptotically optimal estimates, we note that these networks were trained with only $5$ iterations, yet continue to improve and remain stable far beyond this limit. Finally, comparing wall clock time for each optimizer, iterative amortization is only roughly $1.25 \times$ slower than CEM and $1.15 \times$ slower than Adam, making iterative amortization still substantially more efficient.

\subsection{Additional 2D Optimization Plots}
\label{sec: iapo additional 2d opt plots}

In Figure 1, we provided an example of suboptimal optimization resulting from direct amortization on the \texttt{Hopper-v2} environment. We also demonstrated that iterative amortization is capable of automatically generalizing to this optimization surface, outperforming the direct amortized policy. To show that this is a general phenomenon, in Figure~\ref{fig: iapo additional 2d plots}, we present examples of corresponding 2D plots for each of the other environments considered in this paper. As before, we see that direct amortization is near-optimal, but, in some cases, does not match the optimal estimate. In contrast, iterative amortization is able to find the optimal estimate, again, generalizing to the unseen optimization surfaces.

\subsection{Additional Optimization \& the Amortization Gap}
\label{sec: iapo additional iters}

In Section 4, we compared the performance of direct and iterative amortization, as well as their estimated amortization gaps. In this section, we provide additional results analyzing the relationship between policy optimization and the performance in the actual environment. As emphasized in the main paper, this relationship is complex, as optimizing an inaccurate $Q$-value estimate does not improve task performance. Likewise, optimization improvement accrues over the course of training, facilitating collecting high-value state-action pairs in the environment.

\begin{figure}[t!]
    \centering
    \begin{subfigure}[t]{0.24\textwidth}
        \centering
        \includegraphics[width=\textwidth]{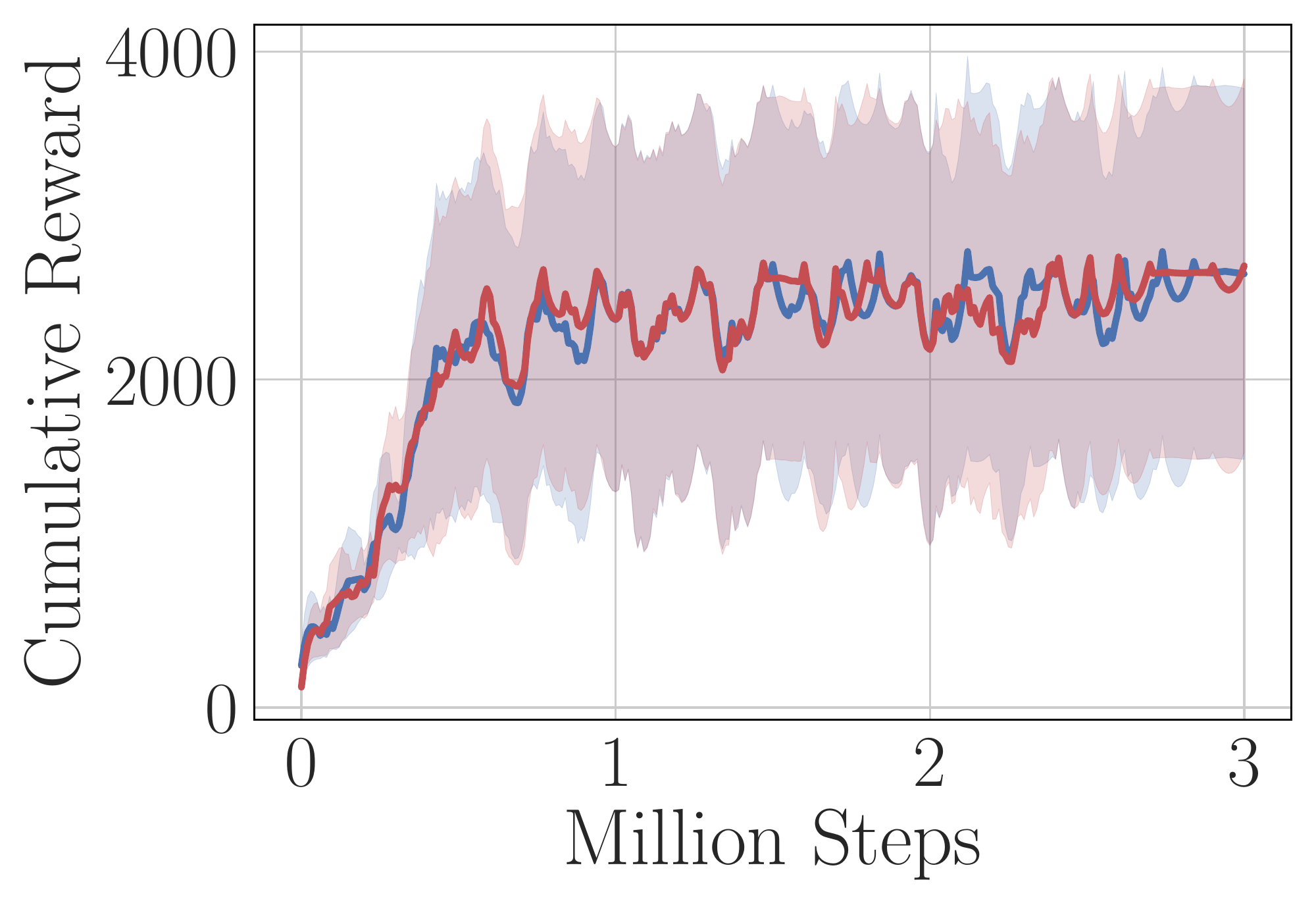}
        \caption{\texttt{Hopper-v2}}
    \end{subfigure}%
    ~ 
    \begin{subfigure}[t]{0.24\textwidth}
        \centering
        \includegraphics[width=\textwidth]{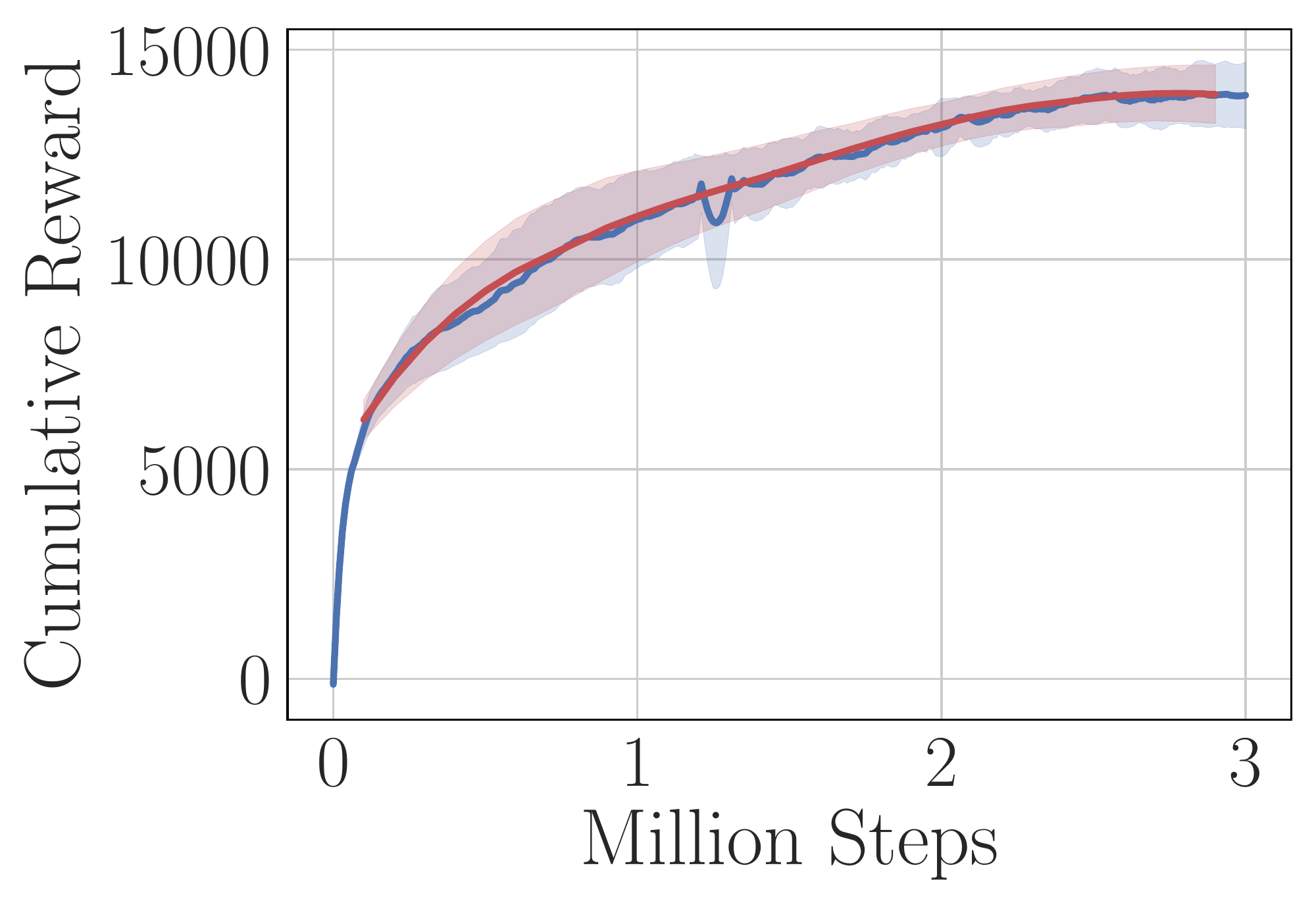}
        \caption{\texttt{HalfCheetah-v2}}
    \end{subfigure}%
    ~ 
    \begin{subfigure}[t]{0.24\textwidth}
        \centering
        \includegraphics[width=\textwidth]{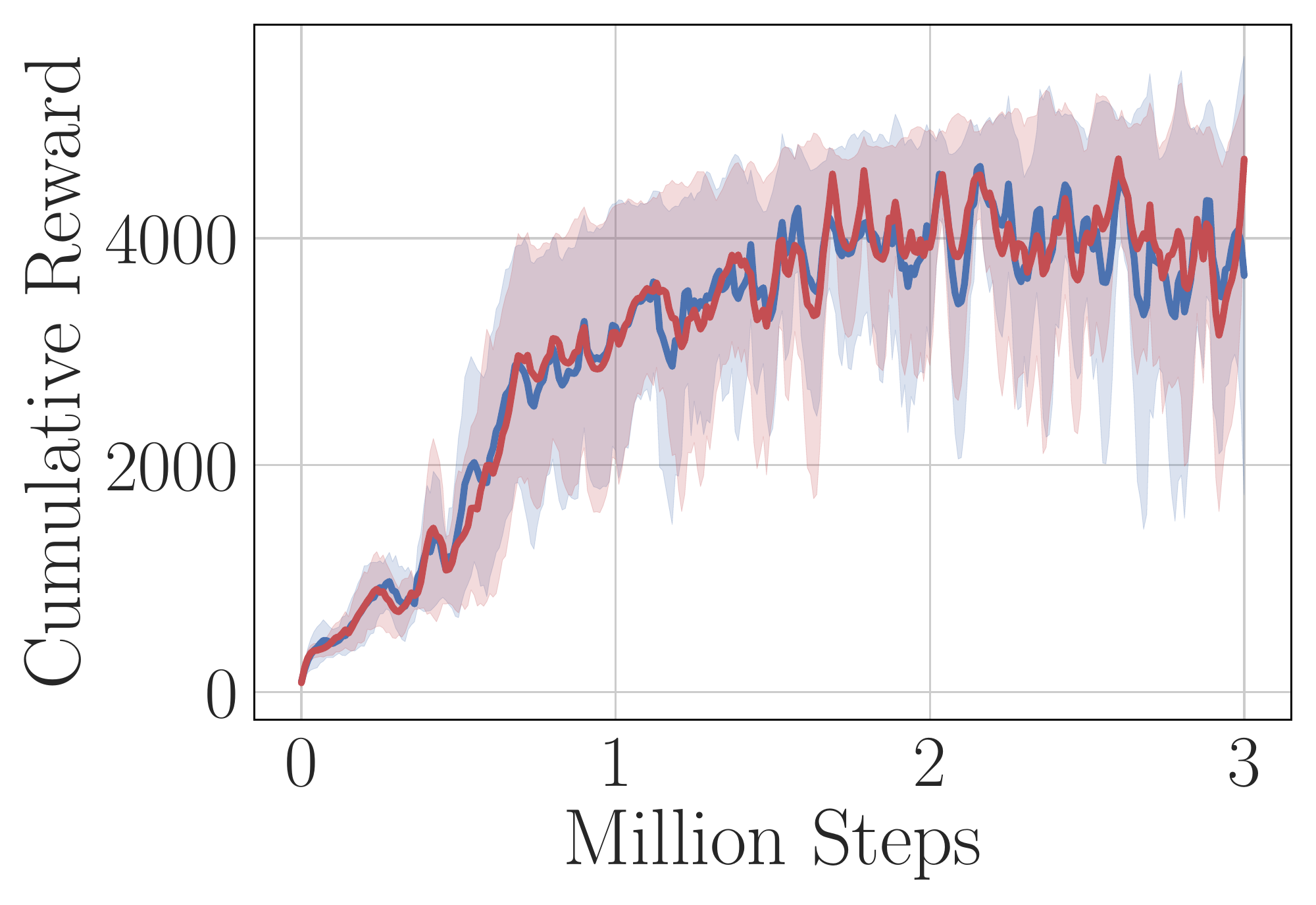}
        \caption{\texttt{Walker2d-v2}}
    \end{subfigure}%
    ~ 
    \begin{subfigure}[t]{0.24\textwidth}
        \centering
        \includegraphics[width=\textwidth]{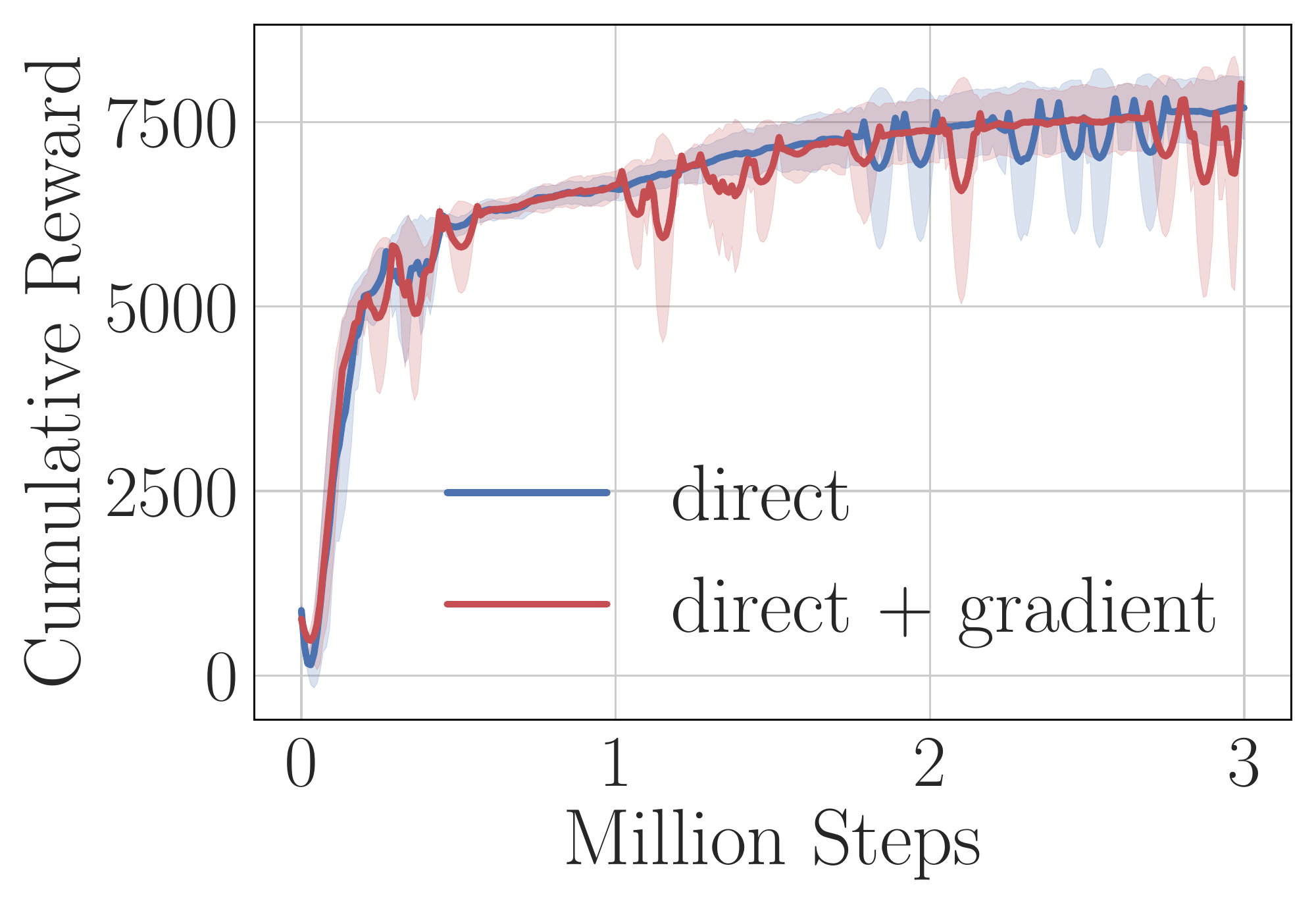}
        \caption{\texttt{Ant-v2}}
    \end{subfigure}
    \caption{\textbf{Test-Time Gradient-Based Optimization}. Each plot compares the performance of direct amortization vs.~direct amortization with $50$ additional gradient-based policy optimization iterations. Note that this additional optimization is only performed at test time.}
    \label{fig: iapo add grad iter test time}
\end{figure}

The amortization gap quantifies the suboptimality in the objective, $\mathcal{J}$, of the policy estimate. As described in Section~\ref{appendix: am gap calc}, we estimate the optimized policy by performing additional gradient-based optimization on the policy distribution parameters (mean and variance). However, as noted, this gap is relatively small compared to the objective itself. Thus, when we deploy this optimized policy for evaluation in the actual environment, as shown for direct amortization in Figure~\ref{fig: iapo add grad iter test time}, we do not observe a noticeable difference in performance.

\begin{figure}[t!]
    \centering
    \begin{subfigure}[t]{0.24\textwidth}
        \centering
        \includegraphics[width=\textwidth]{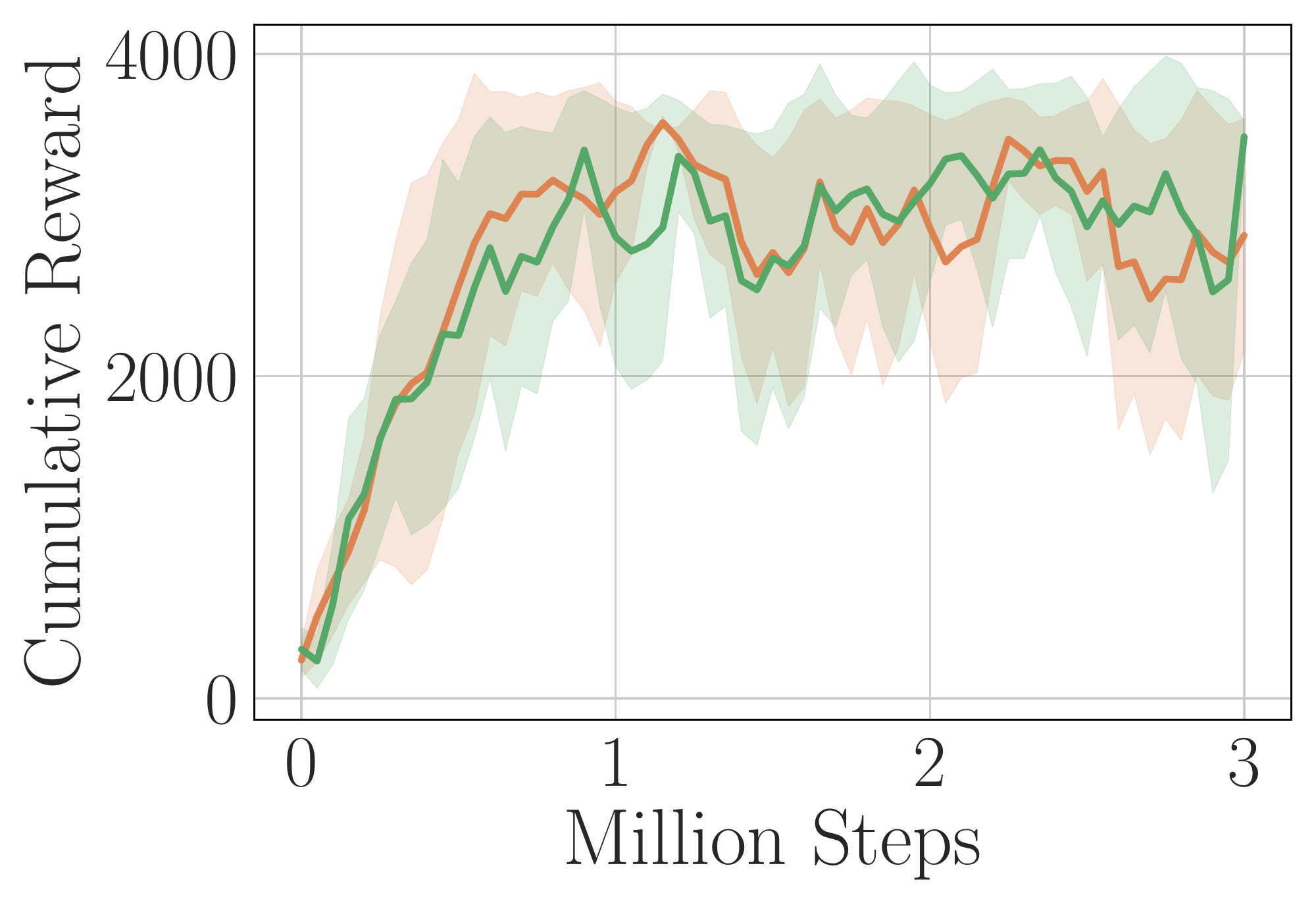}
        \caption{\texttt{Hopper-v2}}
    \end{subfigure}%
    ~ 
    \begin{subfigure}[t]{0.24\textwidth}
        \centering
        \includegraphics[width=\textwidth]{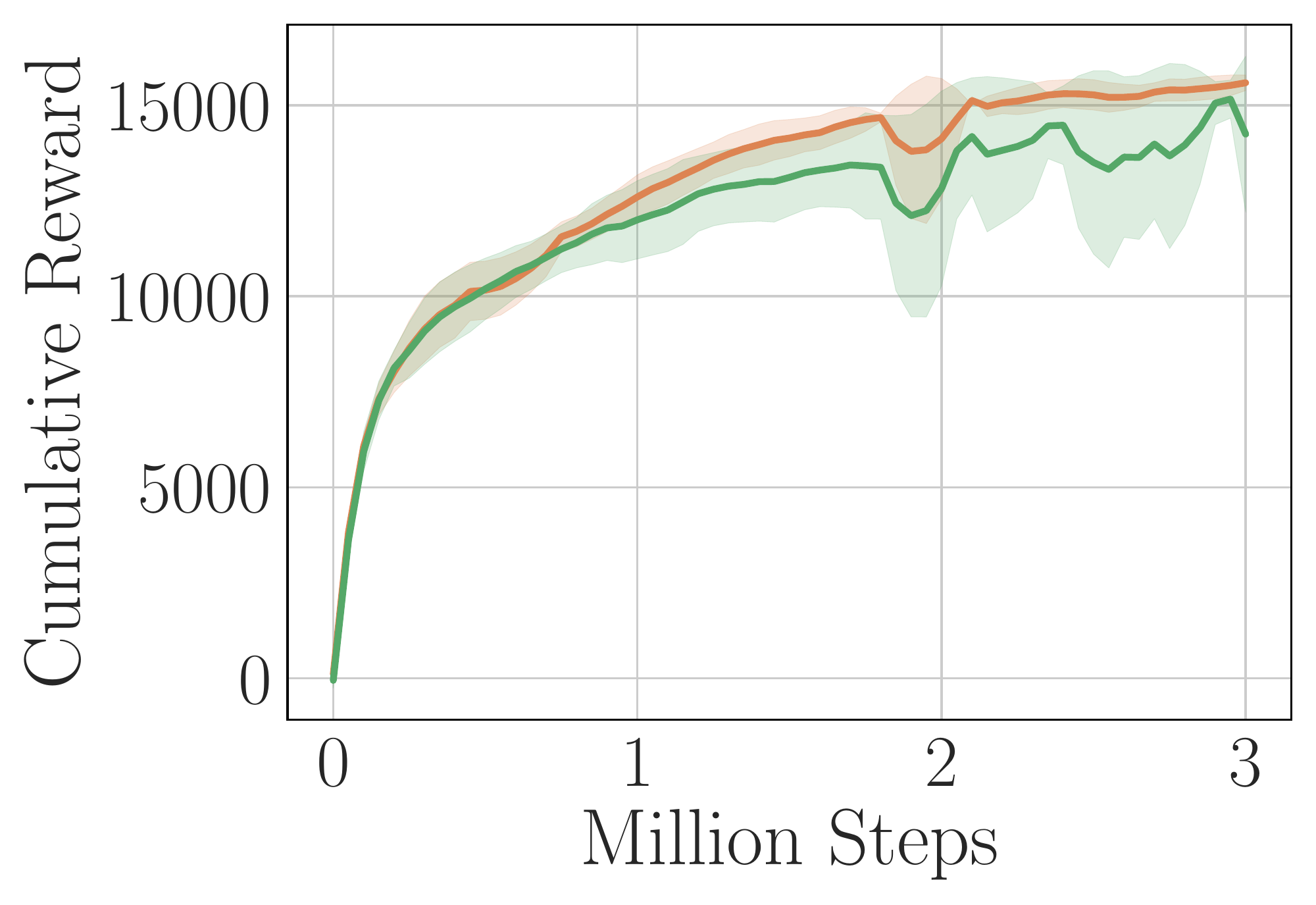}
        \caption{\texttt{HalfCheetah-v2}}
    \end{subfigure}%
    ~ 
    \begin{subfigure}[t]{0.24\textwidth}
        \centering
        \includegraphics[width=\textwidth]{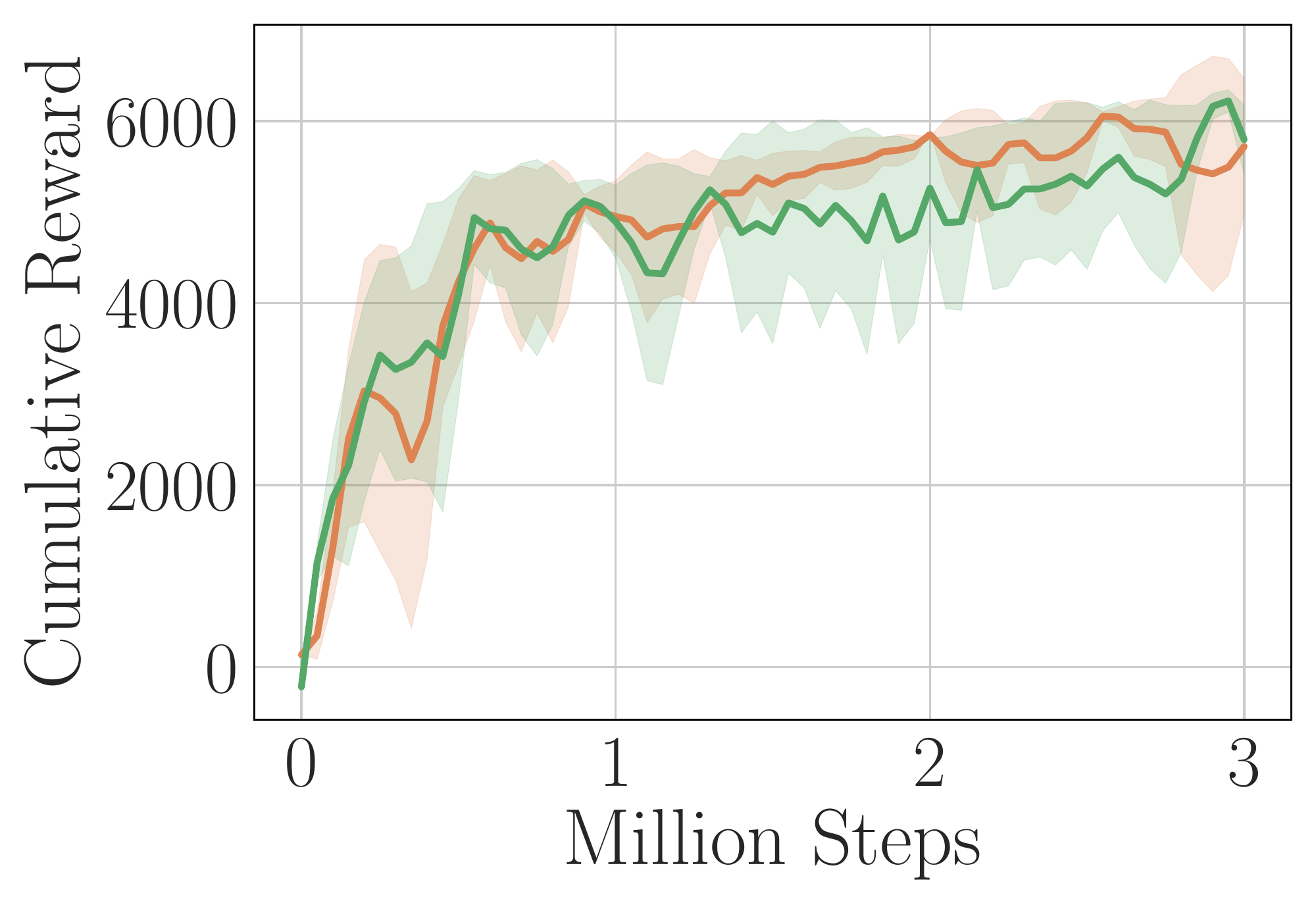}
        \caption{\texttt{Walker2d-v2}}
    \end{subfigure}%
    ~ 
    \begin{subfigure}[t]{0.24\textwidth}
        \centering
        \includegraphics[width=\textwidth]{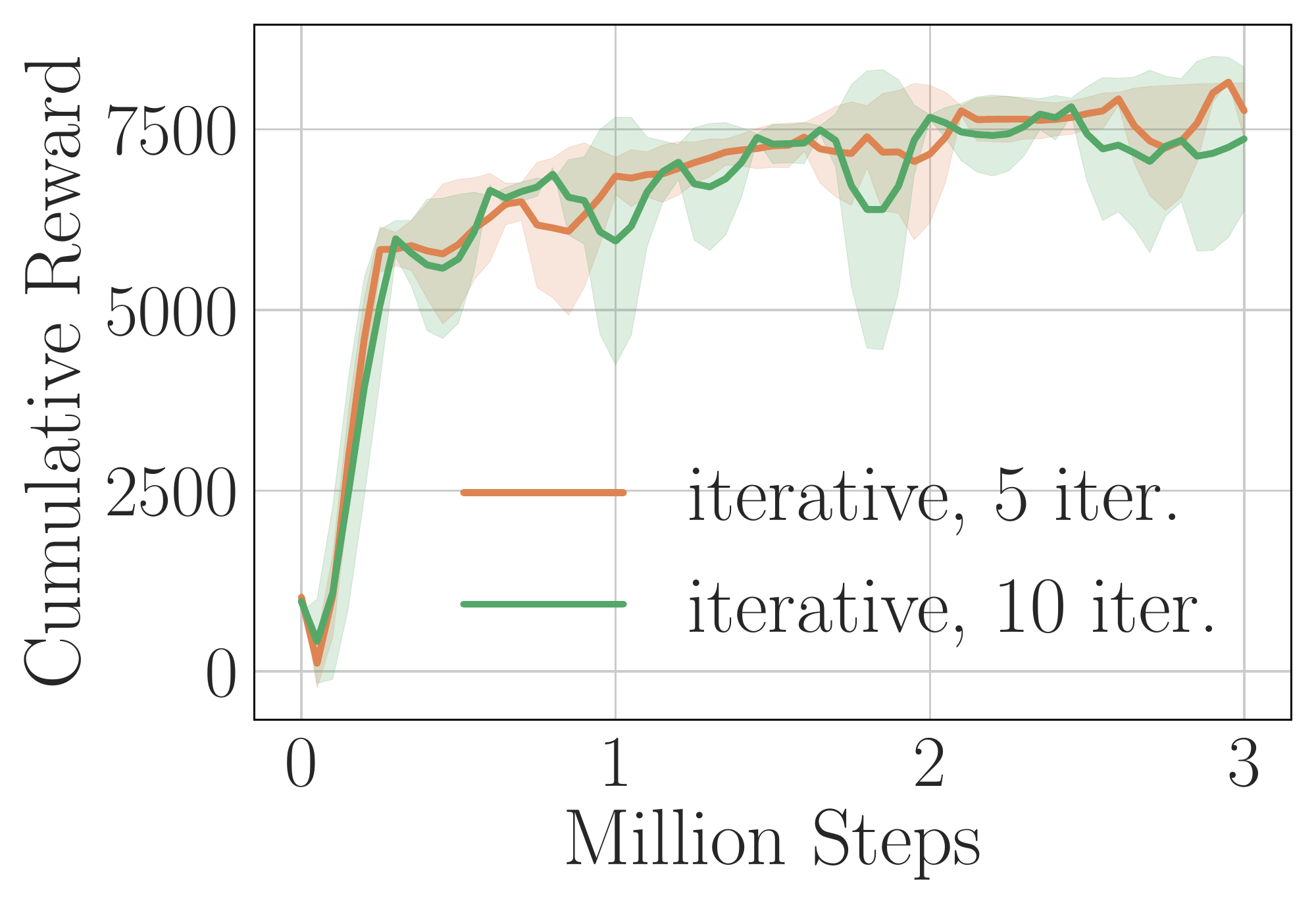}
        \caption{\texttt{Ant-v2}}
    \end{subfigure}
    \caption{\textbf{Additional Amortized Test-Time Iterations}. Each plot compares the performance of iterative amortization (trained with $5$ iterations) vs.~the same agent with an additional $5$ iterations at evaluation. Performance remains similar or slightly worse.}
    \label{fig: iapo add iter test time}
\end{figure}

Likewise, in Section 4.2.2, we observed that using additional amortized iterations during evaluation further decreased the amortization gap for iterative amortization. Yet, when we deploy this more fully optimized policy in the environment, as shown in Figure~\ref{fig: iapo add iter test time}, we do not generally observe a corresponding performance improvement. In fact, on \texttt{HalfCheetah-v2} and \texttt{Walker2d-v2}, we observe a slight \textit{decrease} in performance. This further highlights the fact that additional policy optimization may exploit inaccurate $Q$-value estimates.

However, importantly, in Figures~\ref{fig: iapo add grad iter test time} and \ref{fig: iapo add iter test time}, the additional policy optimization is only performed for evaluation. That is, the data collected with the more fully optimized policy is not used for training and therefore cannot be used to correct the inaccurate value estimates. Thus, while more accurate policy optimization, as quantified by the amortization gap, may not substantially affect \textit{evaluation} performance, it can play a role in improving \textit{training}.

\begin{figure}[t!]
    \centering
    \begin{subfigure}[t]{0.24\textwidth}
        \centering
        \includegraphics[width=\textwidth]{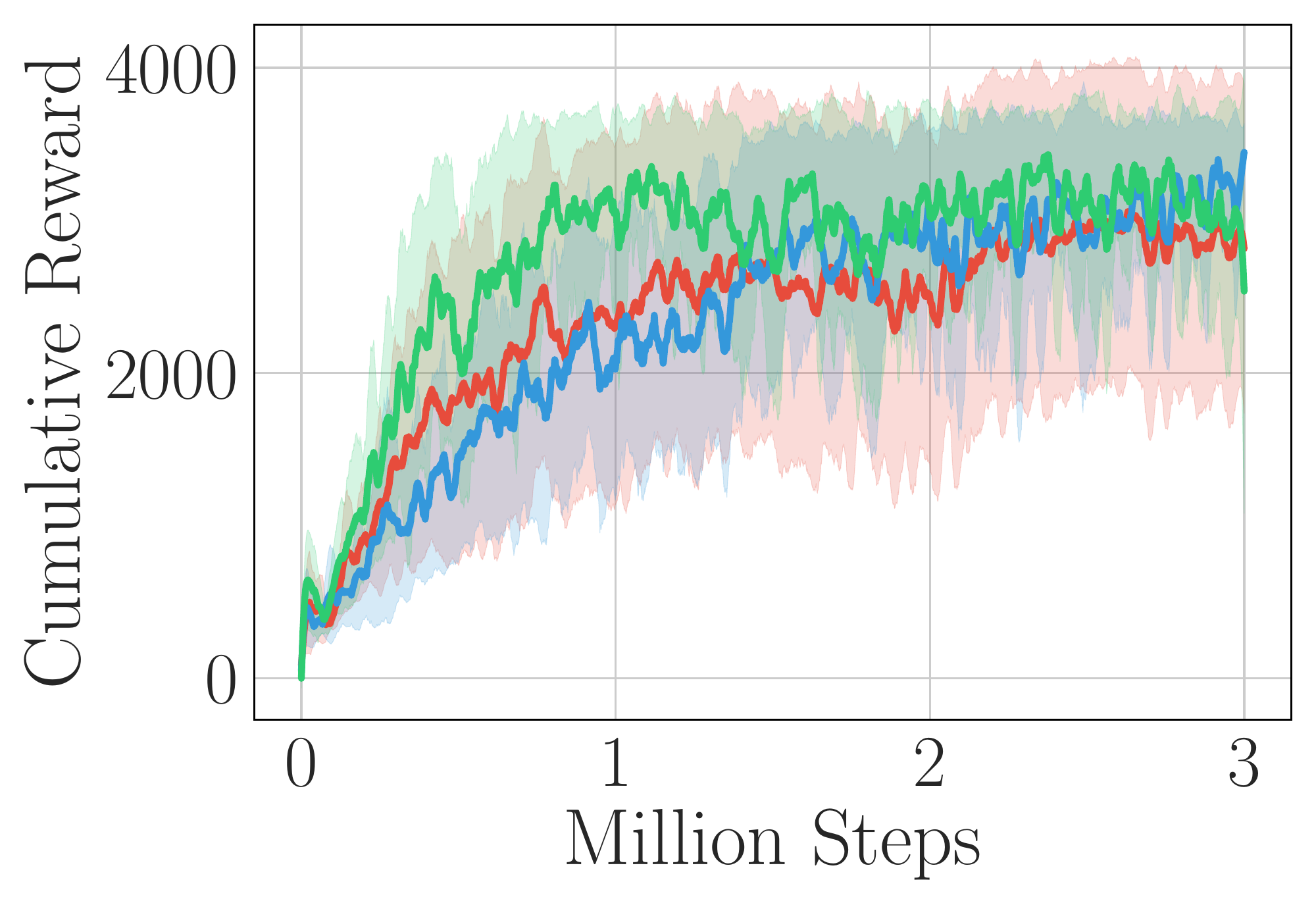}
        \caption{\texttt{Hopper-v2}}
    \end{subfigure}%
    ~ 
    \begin{subfigure}[t]{0.24\textwidth}
        \centering
        \includegraphics[width=\textwidth]{figures/it_halfcheetah4.pdf}
        \caption{\texttt{HalfCheetah-v2}}
    \end{subfigure}%
    ~ 
    \begin{subfigure}[t]{0.24\textwidth}
        \centering
        \includegraphics[width=\textwidth]{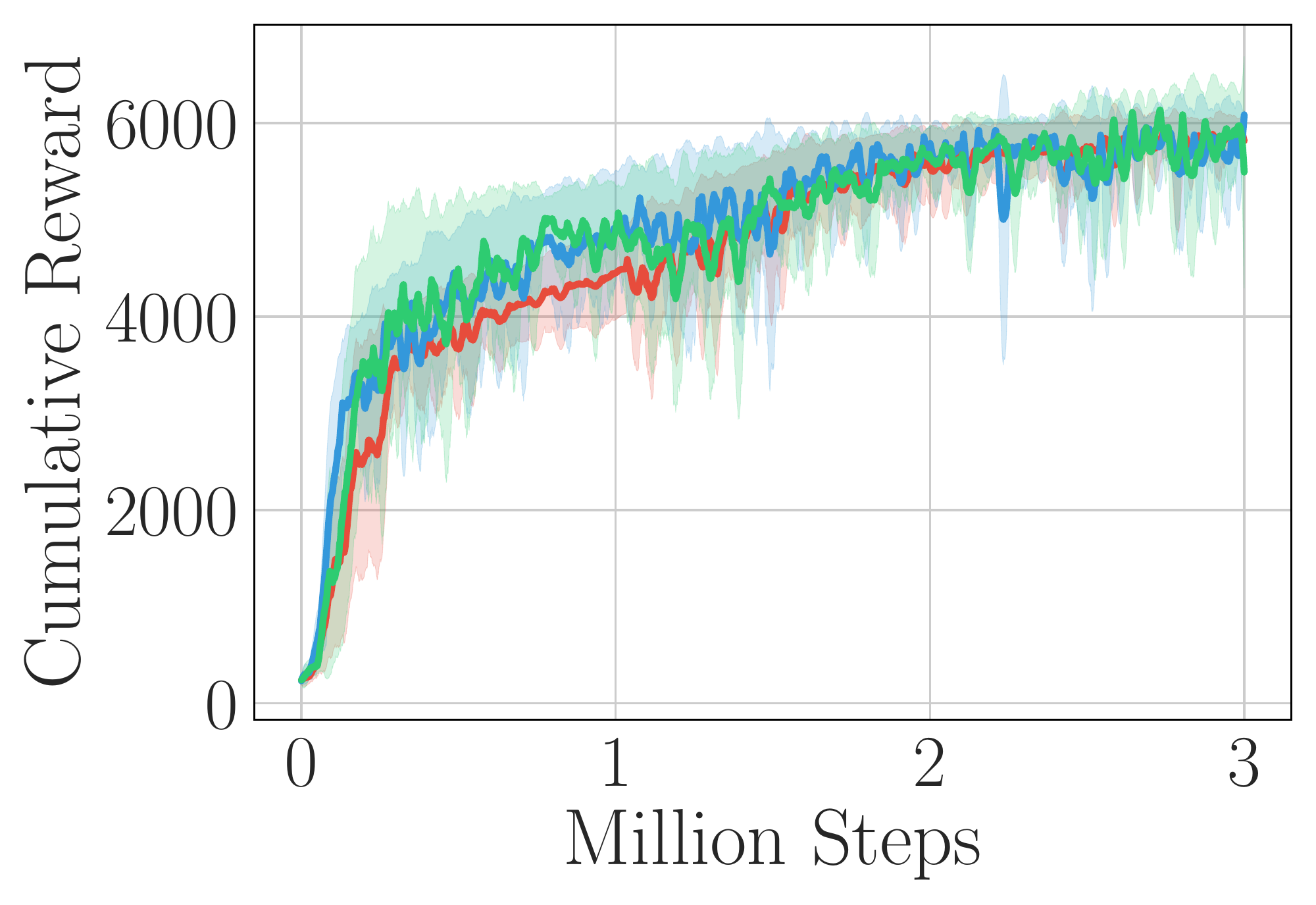}
        \caption{\texttt{Walker2d-v2}}
    \end{subfigure}%
    ~ 
    \begin{subfigure}[t]{0.24\textwidth}
        \centering
        \includegraphics[width=\textwidth]{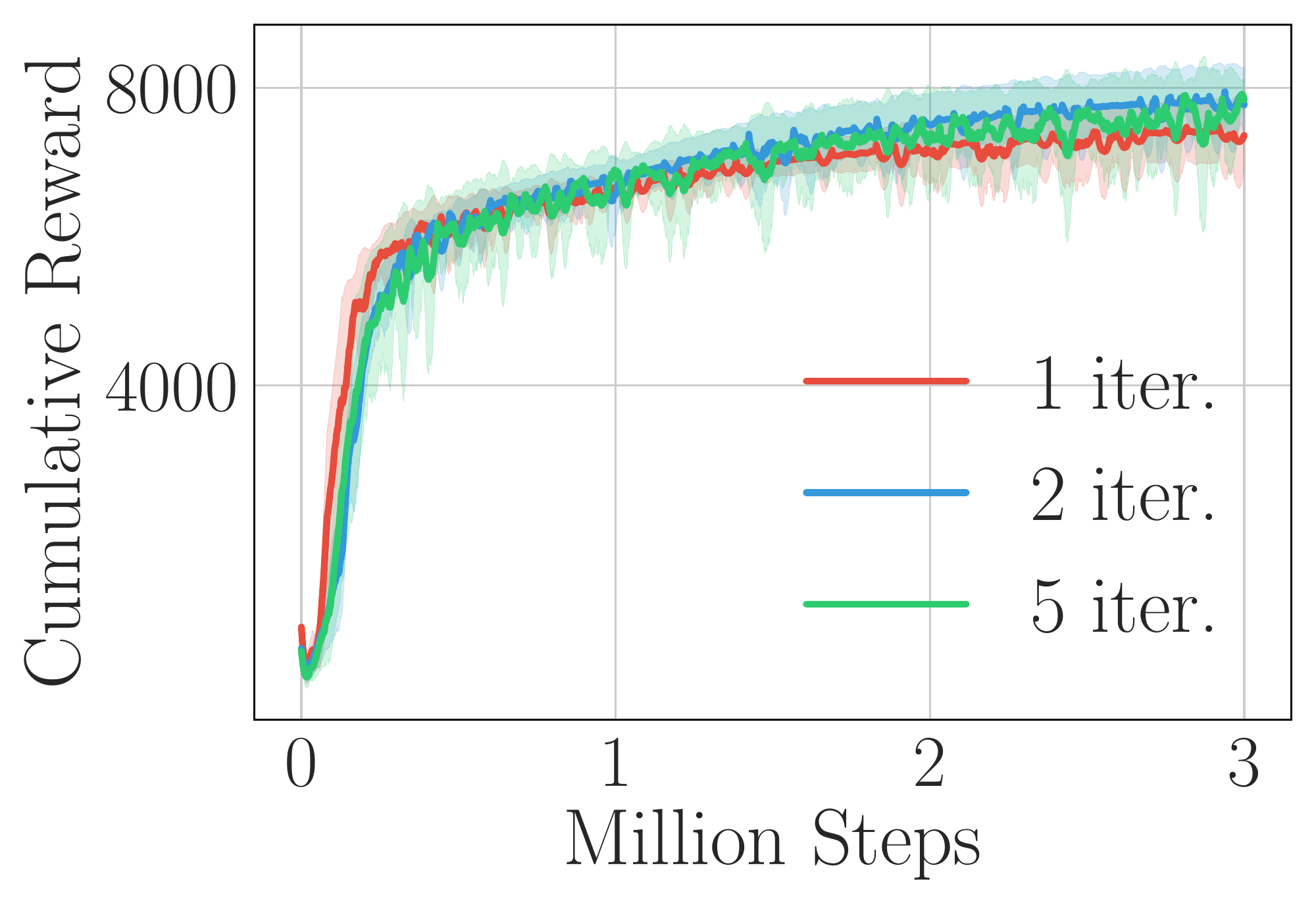}
        \caption{\texttt{Ant-v2}}
    \end{subfigure}
    \caption{\textbf{Performance of Varying Iterations During Training}. }
    \label{fig: iapo train iters perf am gap}
\end{figure}

\begin{figure}[t!]
    \centering
    \begin{subfigure}[t]{0.24\textwidth}
        \centering
        \includegraphics[width=\textwidth]{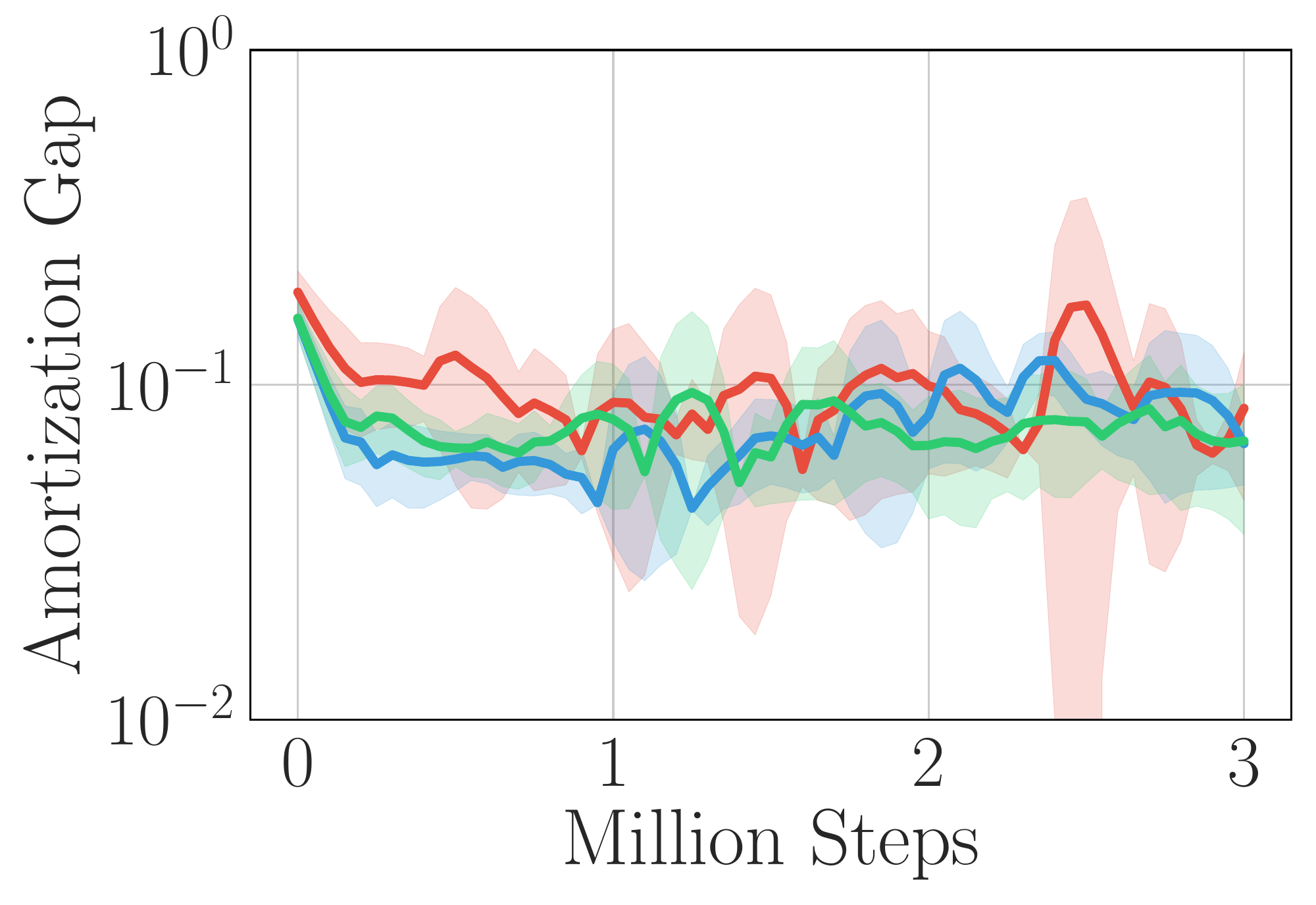}
        \caption{\texttt{Hopper-v2}}
    \end{subfigure}%
    ~ 
    \begin{subfigure}[t]{0.24\textwidth}
        \centering
        \includegraphics[width=\textwidth]{figures/am_gap_it_halfcheetah3.pdf}
        \caption{\texttt{HalfCheetah-v2}}
    \end{subfigure}%
    ~ 
    \begin{subfigure}[t]{0.24\textwidth}
        \centering
        \includegraphics[width=\textwidth]{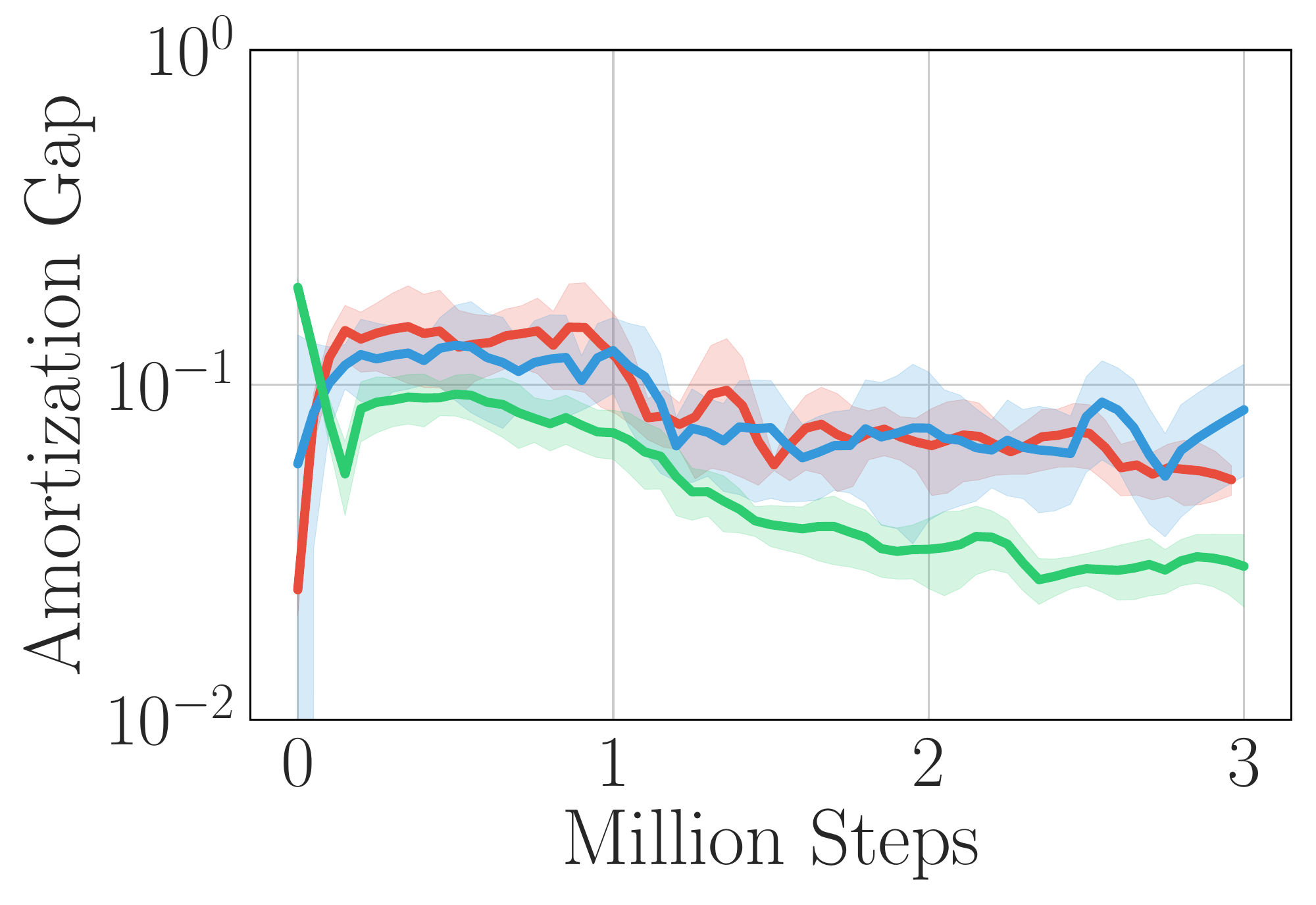}
        \caption{\texttt{Walker2d-v2}}
    \end{subfigure}%
    ~ 
    \begin{subfigure}[t]{0.24\textwidth}
        \centering
        \includegraphics[width=\textwidth]{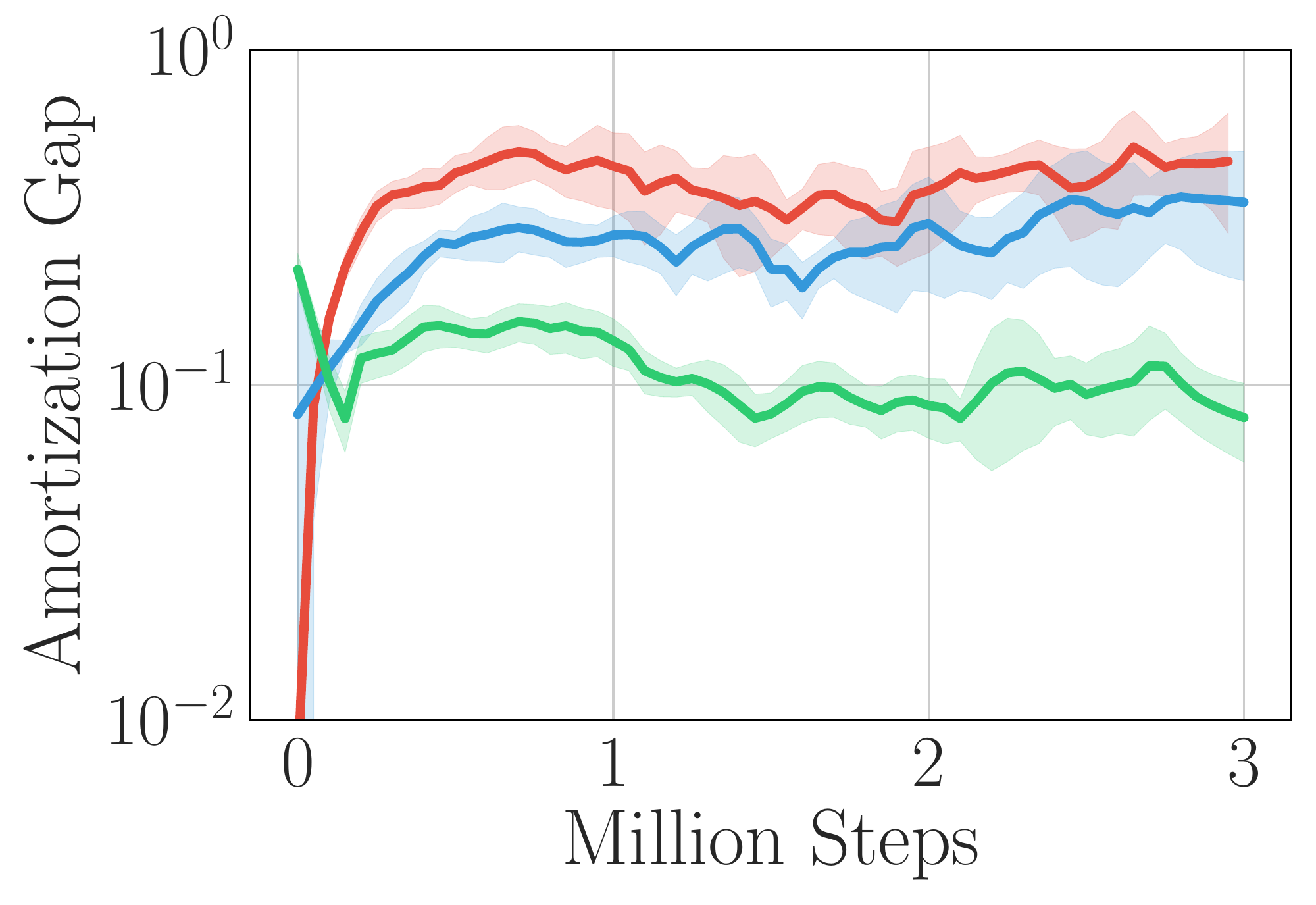}
        \caption{\texttt{Ant-v2}}
    \end{subfigure}
    \caption{\textbf{Amortization Gap of Varying Iterations During Training}. }
    \label{fig: iapo train iters perf am gap2}
\end{figure}

This aspect is explored in Figures~\ref{fig: iapo train iters perf am gap} \& \ref{fig: iapo train iters perf am gap2}, where we plot the performance and amortization gap of iterative amortization with varying numbers of iterations ($1$, $2$ and $5$) throughout training. While the trend is not exact in all cases, we generally observe that increasing iterations yields improved performance and more accurate optimization. \texttt{Walker2d-v2} provides an interesting example. Even with a single iteration, we see that iterative amortization outperforms direct amortization, suggesting that multi-modality is the dominant factor for improved performance here. Yet, $1$ iteration is slightly worse compared with $2$ and $5$ iterations early in training, both in terms of performance and optimization. As the amortization gap decreases later in training, we see that the performance gap ultimately decreases. Further work could help to analyze this process in even more detail.

% \begin{figure}[t!]
%     \centering
%     \begin{subfigure}[t]{0.24\textwidth}
%         \centering
%         \includegraphics[width=\textwidth]{figures/it_halfcheetah.pdf}
%         \caption{}
%     \end{subfigure}%
%     ~ 
%     \begin{subfigure}[t]{0.24\textwidth}
%         \centering
%         \includegraphics[width=\textwidth]{figures/am_gap_it_halfcheetah.pdf}
%         \caption{}
%     \end{subfigure}
%     \caption{\textbf{Iterations During Training}. \textbf{(a)} Performance and \textbf{(b)} estimated amortization gap for varying numbers of policy optimization iterations per step during training on \texttt{HalfCheetah-v2}. Increasing the iterations improves performance and decreases the estimated amortization gap.}
%     \label{fig: iapo train iters perf am gap}
% \end{figure}

\subsection{Multiple Policy Estimates}
\label{appendix: multiple policy estimates}

As discussed in Section~3.2, iterative amortization has the added benefit of potentially obtaining multiple policy distribution estimates, due to stochasticity in the optimization procedure as well as initialization. In contrast, unless latent variables are incorporated into the policy, direct amortization is limited to a single policy estimate. In Section 4.3.2, we analyzed multi-modality by comparing the distance between difference optimization runs of iterative amortization on \texttt{Walker2d-v2}. Here, we present the same analysis on each of the other three environments considered in this paper.

\begin{figure}[t!]
    \centering
    \begin{subfigure}[t]{0.25\textwidth}
    \centering
    \includegraphics[width=\textwidth]{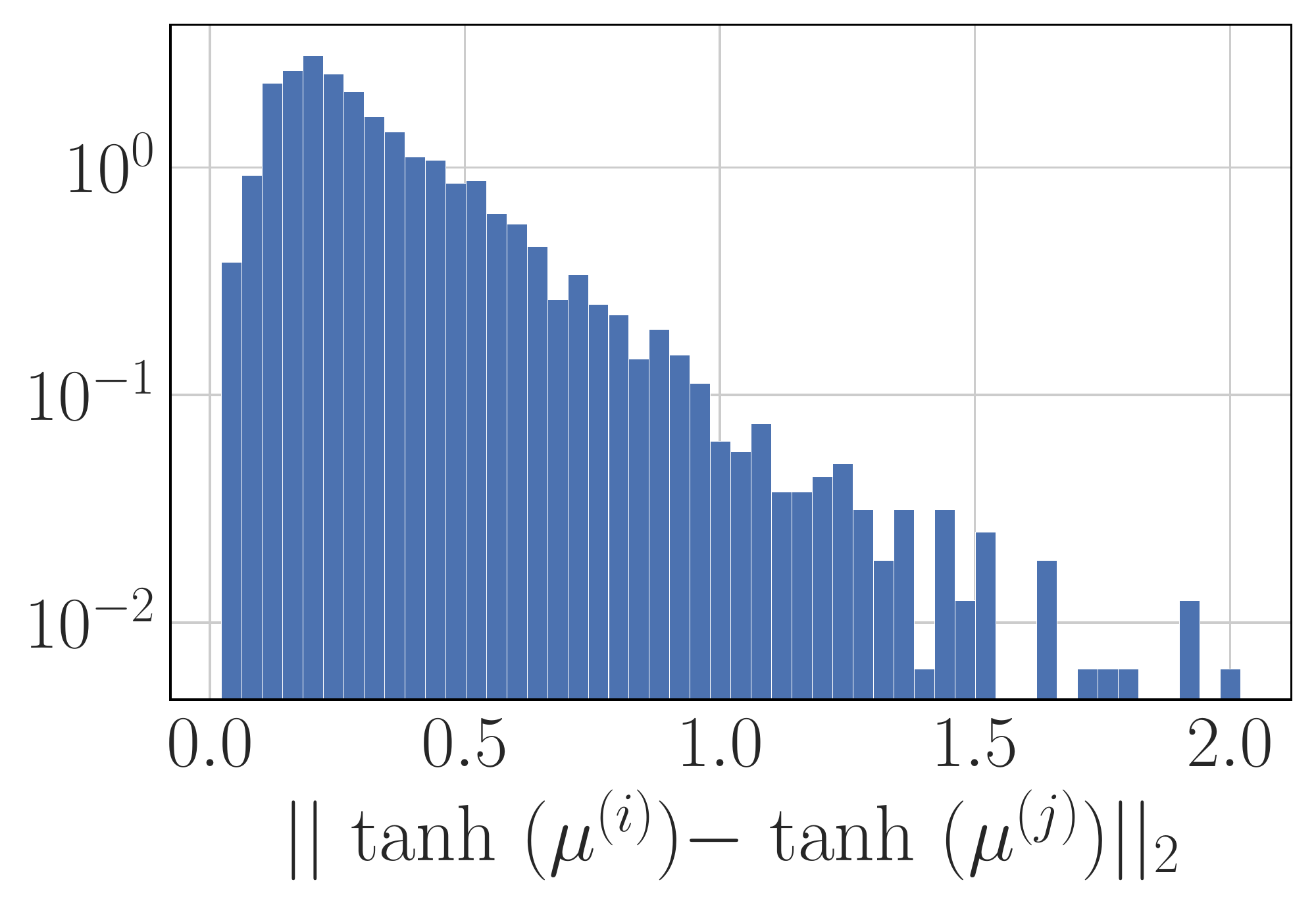}
    \caption{}
    \end{subfigure} 
    \begin{subfigure}[t]{0.22\textwidth}
    \centering
    \includegraphics[width=\textwidth]{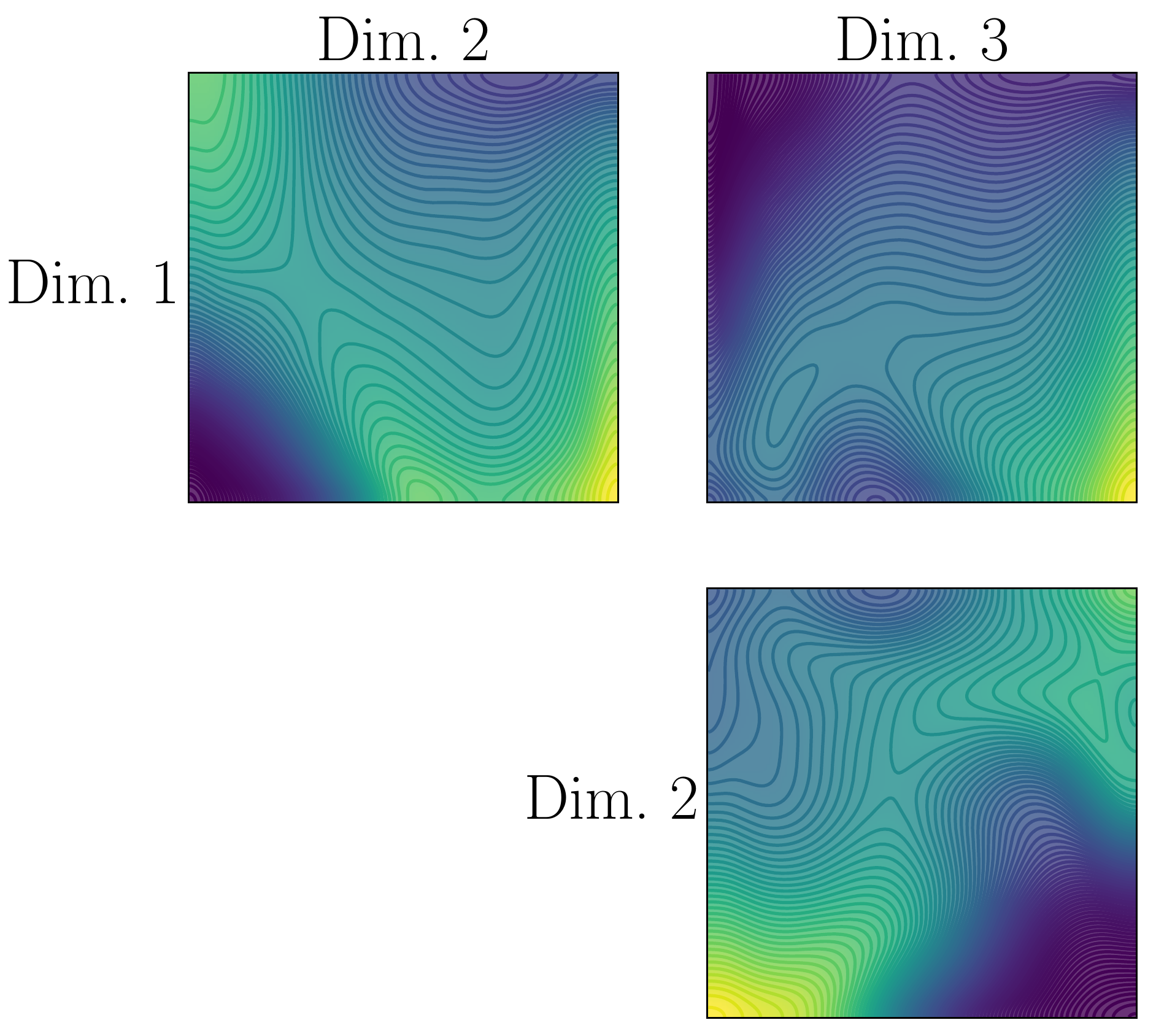}
    \caption{}
    \end{subfigure}
    \begin{subfigure}[t]{0.47\textwidth}
    \centering
    \includegraphics[width=\textwidth]{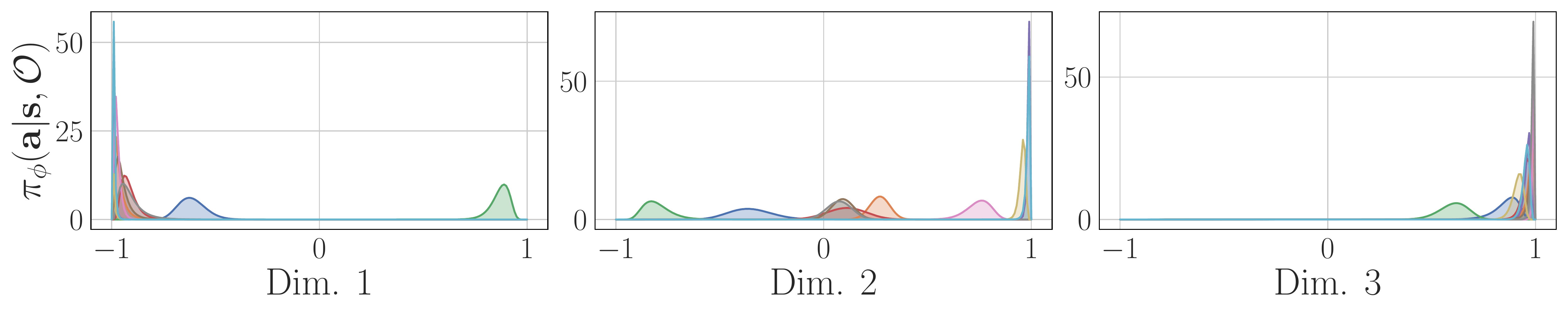}
    \caption{}
    \end{subfigure}
    \caption{\textbf{Multiple Policy Modes on \texttt{Hopper-v2}}.}
    \label{fig: iapo multiples modes experiment hopper}
\end{figure}

\begin{figure}[t!]
    \centering
    \begin{subfigure}[t]{0.25\textwidth}
    \centering
    \includegraphics[width=\textwidth]{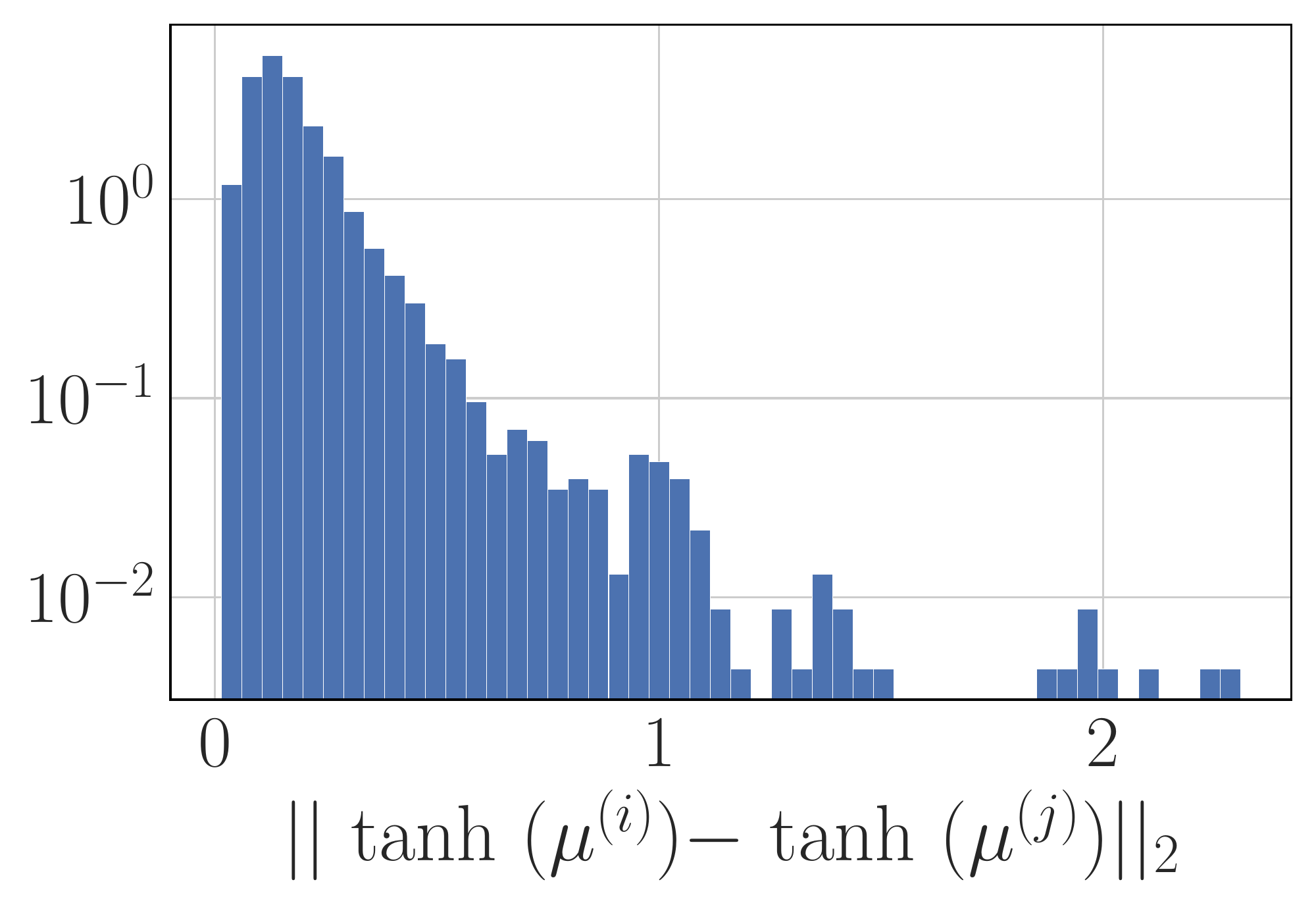}
    \caption{}
    \label{fig: iapo mean distance}
    \end{subfigure} 
    \begin{subfigure}[t]{0.22\textwidth}
    \centering
    \includegraphics[width=\textwidth]{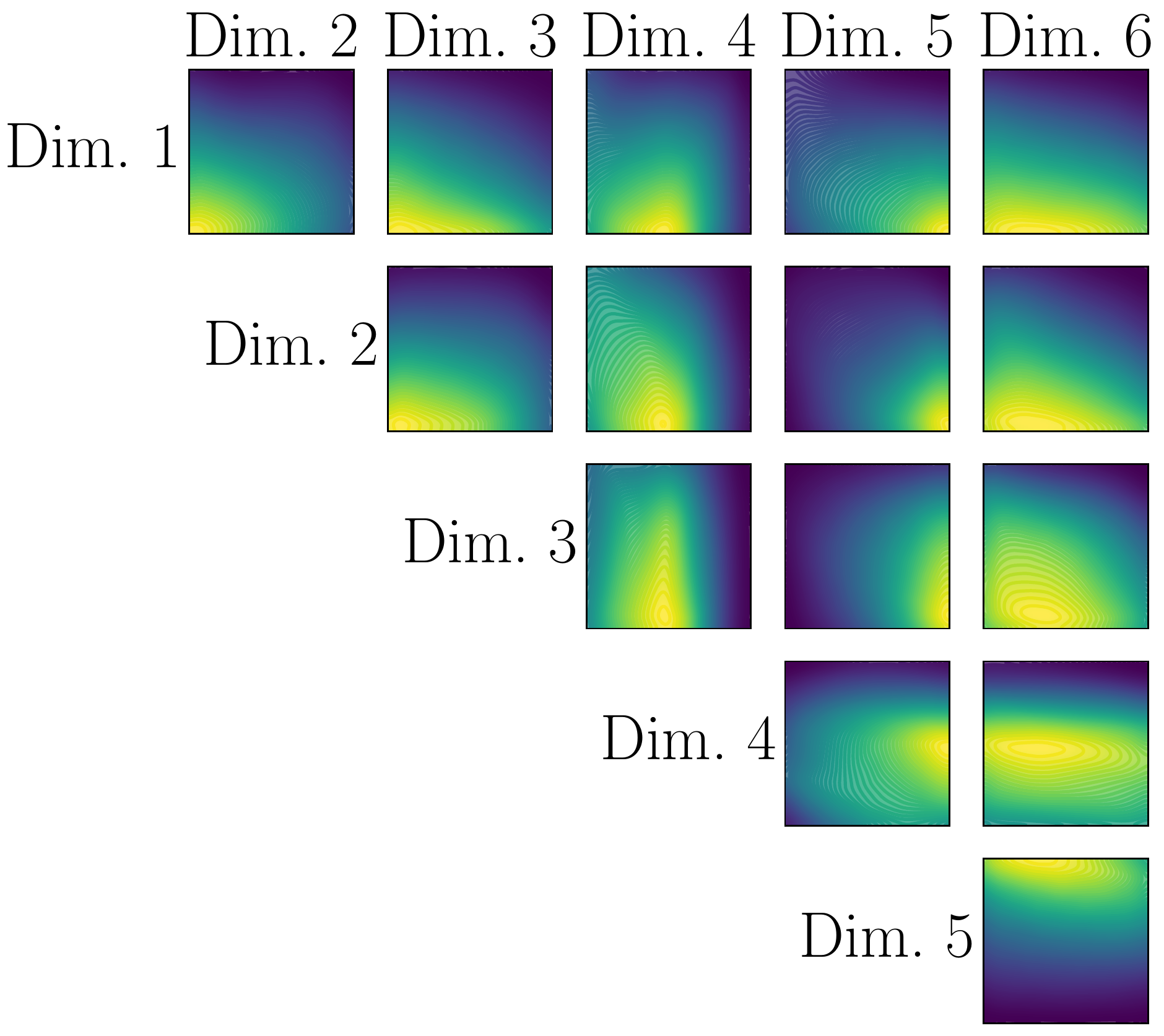}
    \caption{}
    \label{fig: iapo multi dim opt}
    \end{subfigure}
    \begin{subfigure}[t]{0.47\textwidth}
    \centering
    \includegraphics[width=\textwidth]{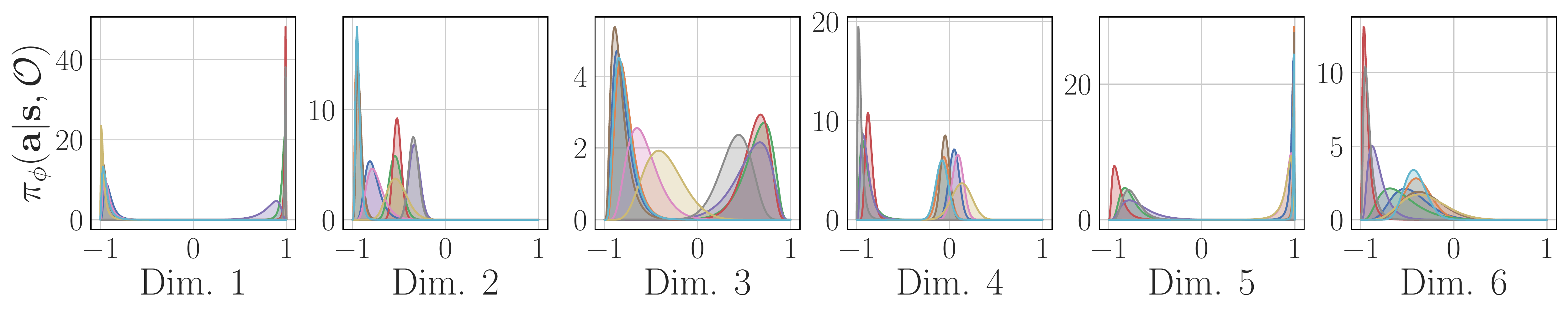}
    \caption{}
    \label{fig: iapo multi dim density}
    \end{subfigure}
    \caption{\textbf{Multiple Policy Modes on \texttt{HalfCheetah-v2}}.}
    \label{fig: iapo multiples modes experiment halfcheetah}
\end{figure}

\begin{figure}[t!]
    \centering
    \begin{subfigure}[t]{0.25\textwidth}
    \centering
    \includegraphics[width=\textwidth]{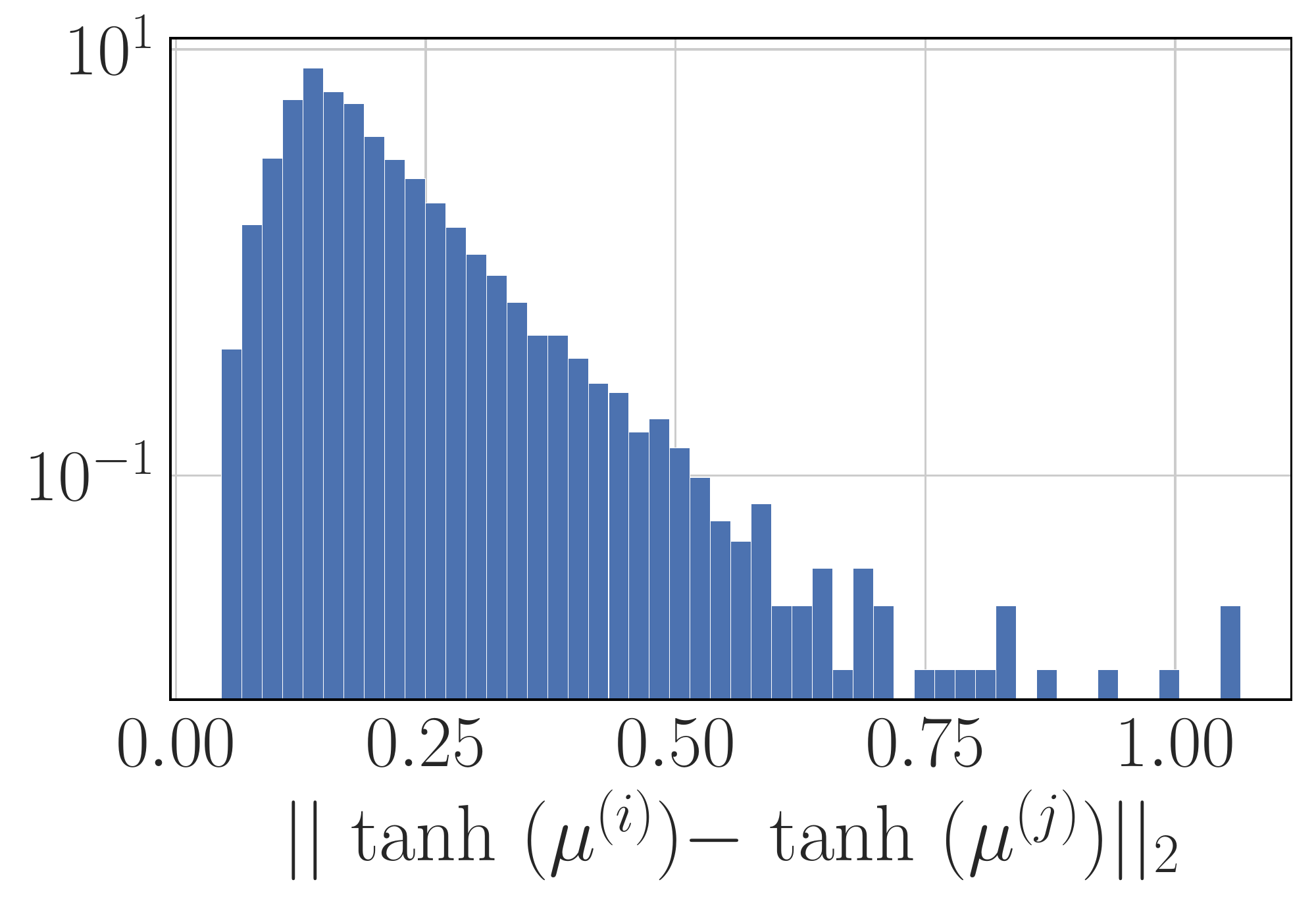}
    \caption{}
    \end{subfigure} 
    \begin{subfigure}[t]{0.22\textwidth}
    \centering
    \includegraphics[width=\textwidth]{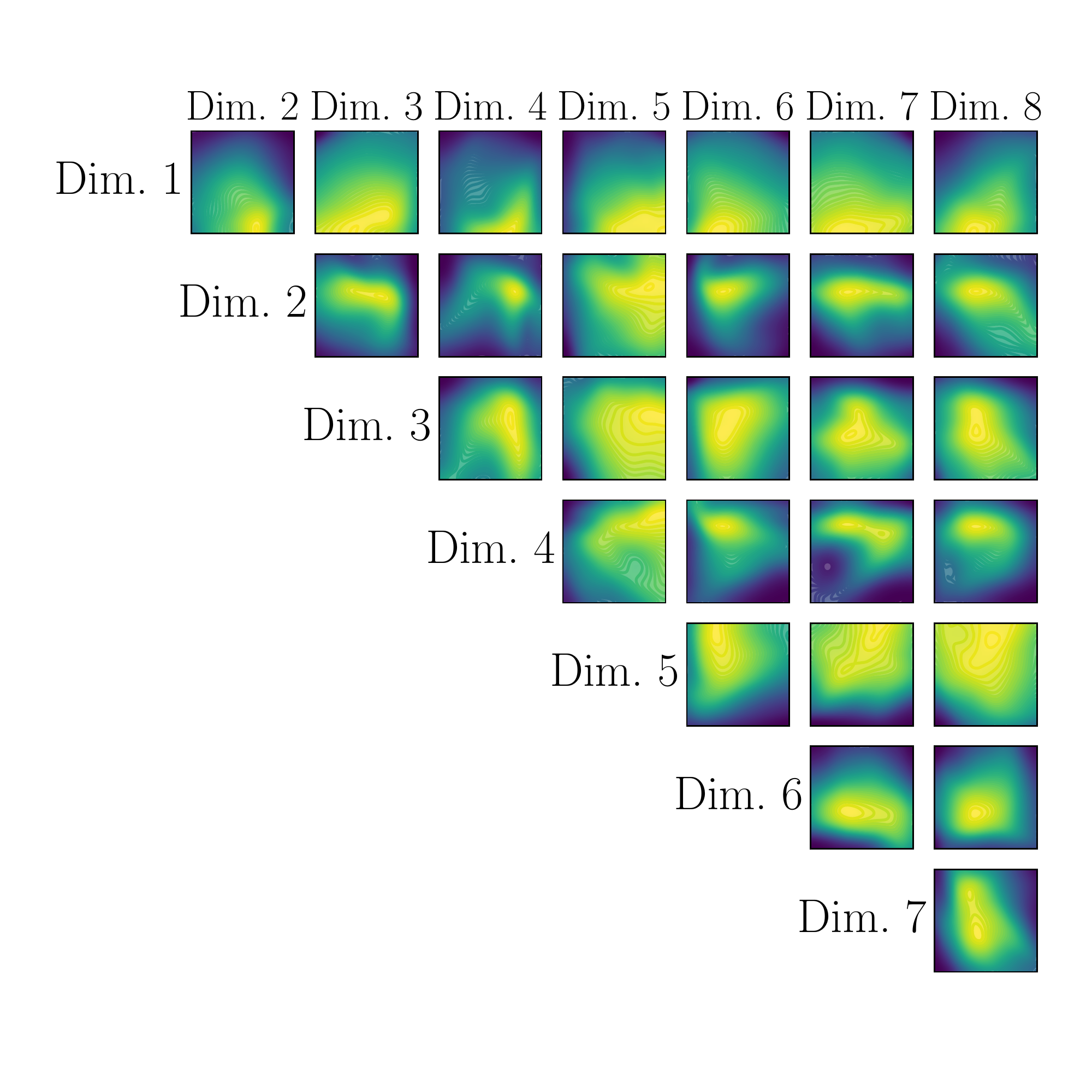}
    \caption{}
    \end{subfigure}
    \begin{subfigure}[t]{0.47\textwidth}
    \centering
    \includegraphics[width=\textwidth]{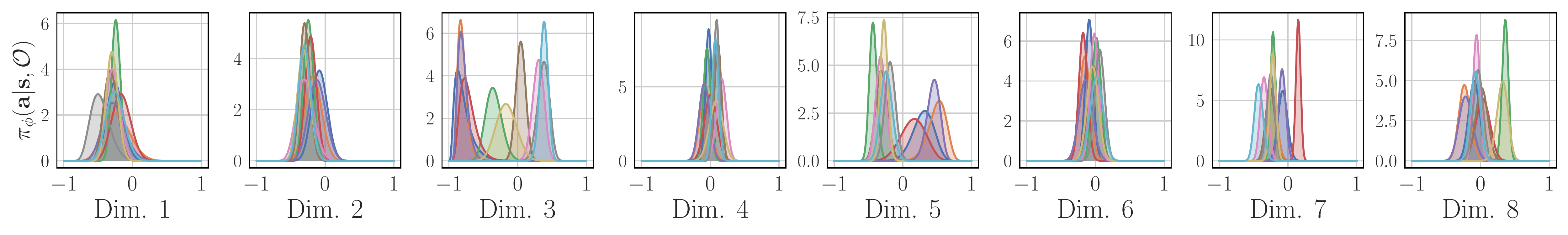}
    \caption{}
    \end{subfigure}
    \caption{\textbf{Multiple Policy Modes on \texttt{Ant-v2}}.}
    \label{fig: iapo multiples modes experiment ant}
\end{figure}

Again, we perform $10$ separate runs of policy optimization per state and evaluate the L2 distance between the means of these policy estimates after applying the \texttt{tanh} transform. Note that in MuJoCo action spaces, which are bounded to $[-1, 1]$, the maximum distance is $2 \sqrt{|\mathcal{A}|}$, where $|\mathcal{A}|$ is the size of the action space. We evaluate the policy mean distance over $100$ states and all $5$ experiment seeds. Results are shown in Figures~\ref{fig: iapo multiples modes experiment halfcheetah}, \ref{fig: iapo multiples modes experiment hopper}, and \ref{fig: iapo multiples modes experiment ant}, where we plot \textbf{(a)} the histogram of distances between policy means ($\bm{\mu}$) across optimization runs ($i$ and $j$) over seeds and states at $3$ million environment steps, \textbf{(b)} the projected optimization surface on each pair of action dimensions for the state with the largest distance, and  \textbf{(c)} the policy density for $10$ optimization runs. As before, we see that some subset of states retain multiple policy modes.

% \section{Appendix}

% Optionally include extra information (complete proofs, additional experiments and plots) in the appendix.
% This section will often be part of the supplemental material.

\end{document}